\documentclass[letterpage,journal]{IEEEtran}
\usepackage[english]{babel}
\usepackage[T1]{fontenc}
\usepackage[utf8]{inputenc}

\usepackage[numbers,sort&compress]{natbib}

\usepackage[pdftex]{graphicx}
\usepackage{epstopdf}
\usepackage{pdflscape} 
\usepackage{xspace}
\usepackage[cmex10]{amsmath}
\usepackage{amssymb} 
\usepackage{amstext}
\usepackage{amsfonts} 

\usepackage[noend]{algorithmic}
\usepackage{algorithm}

\usepackage{array}

\usepackage[colorlinks]{hyperref}
\usepackage{url}
\usepackage{tabularx}
\usepackage{enumitem}
\usepackage{multirow}
\usepackage{longtable}
\usepackage{booktabs}
\usepackage{caption}
\usepackage[newcommands]{ragged2e}
\newcommand{\PreserveBackslash}[1]{\let\temp=\\#1\let\\=\temp}
\newcolumntype{L}[1]{>{\PreserveBackslash\raggedright}p{#1}}
\newcolumntype{R}[1]{>{\PreserveBackslash\raggedleft}m{#1}}
\newcolumntype{C}[1]{>{\PreserveBackslash\centering}m{#1}}

\newcommand{\pmin}{\ensuremath{p_\text{min}}}
\newcommand{\pmax}{\ensuremath{p_\text{max}}}
\newcommand{\maxgen}{\ensuremath{\max_\text{gen}}}
\newcommand{\Operators}{\ensuremath{\textit{Op}}}
\newcommand{\Metric}{\ensuremath{\textit{OM}}\xspace}
\newcommand{\fixappl}{\ensuremath{fix_\text{appl}}}
\newcommand{\Window}{\ensuremath{\mathcal{W}}}
\newcommand{\oopfixappl}{\ensuremath{o_{op}^{\fixappl}}}
\newcommand{\nsuccopt}{\ensuremath{n_{succ, op}^t}}
\newcommand{\nfailopt}{\ensuremath{n_{fail, op}^t}}
\newcommand{\irace}{\ensuremath{\textsc{irace}}\xspace}
\newcommand{\qmin}{\ensuremath{q_\text{min}}}
\newcommand{\bbob}{\textsc{bbob}\xspace}
\newcommand{\op}{\ensuremath{\textit{op}}\xspace}
\newcommand{\nsuccopg}{\ensuremath{n_{succ, op}^g}\xspace}
\newcommand{\nfailopg}{\ensuremath{n_{fail, op}^g}\xspace}
\newcommand{\OM}{\textit{OM}\xspace}
\newcommand{\JaDE}{\emph{JaDE}\xspace}
\newcommand{\UAOSFW}{\emph{U-AOS-FW}\xspace}
\newcommand{\RecPMaos}{\emph{RecPM-AOS}\xspace}
\newcommand{\PMAdapSS}{\emph{PM-AdapSS}\xspace}
\newcommand{\PM}{\emph{PM}\xspace}
\newcommand{\RecPM}{\emph{RecPM}\xspace}
\newcommand{\RSHADE}{\emph{R-SHADE}\xspace}
\newcommand{\FAUCMAB}{\emph{F-AUC-MAB}\xspace}
\newcommand{\Compass}{\emph{Compass}\xspace}

\newcommand{\fopt}{\ensuremath{f_\mathrm{opt}}}
\newcommand{\aRT}{\ensuremath{\mathrm{aRT}}\xspace}
\newcommand{\RecPMi}{\emph{RecPM1}\xspace}
\newcommand{\DERecPMaos}{\emph{DE-RecPM-AOS}\xspace}

\newcommand{\tableequation}[1]{%
\parbox{20em}{\begin{equation}#1\end{equation}}}

\begin{document}

\title{Unified Framework for the Adaptive \\Operator Selection of Discrete Parameters}

\author{
    \IEEEauthorblockN{Mudita Sharma\IEEEauthorrefmark{1}, Manuel L\'opez-Ib\'a\~nez\IEEEauthorrefmark{2}, Dimitar Kazakov\IEEEauthorrefmark{3}}
    \\\IEEEauthorblockA{\IEEEauthorrefmark{1}University of Essex
    \{ms19173\}@essex.ac.uk}
    \\\IEEEauthorblockA{\IEEEauthorrefmark{2}University of Manchester
    \{manuel.lopez-ibanez\}@manchester.ac.uk}
    \\\IEEEauthorblockA{\IEEEauthorrefmark{3}University of York
    \{dimitar.kazakov\}@york.ac.uk}
}

\maketitle

\begin{abstract}
We conduct an exhaustive survey of adaptive selection of operators (AOS) in Evolutionary Algorithms (EAs). We simplified the AOS structure by adding more components to the framework to built upon the existing categorisation of AOS methods. In addition to simplifying, we looked at the commonality among AOS methods from literature to generalise them. Each component is presented with a number of alternative choices, each represented with a formula. We make three sets of comparisons. First, the methods from literature are tested on the \bbob test bed with their default hyper parameters. Second, the hyper parameters of these methods are tuned using an offline configurator known as \irace. Third, for a given set of problems, we use \irace to select the best combination of components and tune their hyper parameters.
\end{abstract}


\IEEEpeerreviewmaketitle

\section{Introduction}\label{sec:introduction}

\IEEEPARstart{E}{volutionary} algorithms (EAs) are derivative free algorithms that have proven to be useful as optimisation techniques~\cite{Goldberg89} for a wide range of problems in different application areas~\cite{fredriksson2005optimizing, karabouga2011novel, IyeSax2004, Gibbs05:cal, FueDoeHarIor09, GarOliAlb2013tec}. They are known to guarantee near optimum solution for a given problem. The nature of the optimisation problem considered is usually black-box where the algorithm is unaware of the properties of the problem such as continuous, convex, unimodal, multimodal, separable, quadratic, high dimensional, gaussian noise, and so on.

Their performance highly depend on the choice of parameter values. Thus, it is important to employ a parameter selection method that selects the near-optimal parameter values of the algorithm. The parameters with finite or infinite number of choices are known as discrete or continuous parameters respectively. The type of crossover operator such as one-point and multi-point in Genetic Algorithm is a discrete parameter whereas crossover rate in the range $[0.0, 1.0]$ is a continuous parameter. We do not know in advance the best setting of the EA given the problem and have limited knowledge about effect of parameters on EA performance. Thus, to decide the values of parameters either tuning or controlling can be used. Tuning uses offline configurators that refines the optimal configurations after a few runs (Sequential Parameter Optimisation~\cite{BarLasPre2005cec}) or discards the configurations if they are worse than others based on the statistical evidence (\irace~\cite{LopDubPerStuBir2016irace}). The controlling methods use deterministic, self-adaptation or adaptation methods. In case of parameter tuning (offline method), the parameters are tuned and trained on a parameter set. Once the parameters are tuned the values do not change during the run of the algorithm. On the contrary, parameter control (online method) has no training phase and parameter values are adapted and learned during the run of the algorithm. Both these methods have limitations.
On the one hand, offline configurators need extra budget in terms of the function evaluations and do not generalise well on variety of problems, online tuning on the other hand slows down the learning process due to slow exploration of the parameter search space. A survey on parameter tuning and control can be found in~\cite{eiben2011evolutionary, KarHooEib2015:tec}. In this paper we have used the properties of both offline tuning and online control to learn the optimal parameter values during the run of an algorithm. 

We focus on the adaptive selection of discrete parameters (or operators) during the run, known as Adaptive Operator Selection (AOS). The concept of AOS was introduced around 1990s~\cite{davis1989adapting}. Since then, there have been various AOS methods proposed in literature that vary broadly in various aspects such as amount of information to use from the past performance of the algorithm and whether including previous quality in the learning process is an effective approach. It has been classified into two components, Credit Assignment (CA) and Operator Selection (OS)~\cite{Fialho2010PhD}. CA involves a definition based on the fitness achievement over a solution. The most commonly used CA technique is fitness improvement from parent to offspring. OS takes the information captured by CA and estimates the quality of each operator followed by calculating its probability. In the end, a selection technique is employed to select an operator for evolving a parent based on probability assigned to each operator. The same selection technique is used to evolve all parents in a generation. As this process progresses, algorithm learns more and more about the landscape and after a number of generations, this process moves the solutions in a particular search direction. 
\cite{maturana2011adaptive} presents AOS component coupled with Adaptive Operator Management (AOM). Authors explored the hyper-parameter choices for two well-known AOS method: Compass~\cite{maturana2008compass} and Ex-DMAB~\cite{fialho2008extreme}. The exploration of choices is limited to type of window and hyper-parameters of these two AOS methods. They introduced the concept of AOM where operators are unborn, alive or dead based on their performance criteria. It decided whether an operator is important for current stage or not. 
\cite{di2015experimental} makes choices for an AOS method~\cite{maturana2008compass} in order to balance exploration and exploitation concluding that adaptive methods are better than static methods. A survey on adaptive methods can be found in~\cite{AleMos2016slr}. 


Many novel AOS methods can be designed by combining different components of existing AOS methods. There can be many combinations possible from the existing methods that are not tested as an AOS method. To test the efficiency of these methods as an AOS method, this paper presents a unified AOS framework that builds upon the existing classification of an AOS method~\cite{fialho2009analysis}. This is done by analysing multiple AOS methods from the literature to design a simplified framework. The AOS methods used to build the framework are originally proposed in the literature to tune the parameters of various EAs such as genetic algorithms and differential evolution. The framework consist of multiple choices for each AOS component. Some of the choices are inspired from Reinforcement Learning utilised to adapt parameters of different EAs. The framework can be utilised to explore different combinations of the AOS components' choices. As an estimate, we can generate more than $5000$ novel AOS methods from the framework. It can also be used to replicate various known AOS methods from the literature. An AOS method is build from the framework by setting the component choices and fixing the values of their parameters. In order to make the framework widely applicable, choices with diverse properties are included such as immediate progress to far-sighted progress, focusing on the clustered achievement to the outliers, etc. In addition to this, some novel choices are added to the components and each choice is generalised. In the process of generalisation, a number of hyper-parameters are introduced within the choices. The resultant framework is applicable for the online adaptation of discrete parameters of an evolutionary algorithm. 

As the framework consists of various AOS methods with their hyper-parameters, an offline configurator is employed to select an optimal AOS method and tune its hyper-parameters. Thus, we present a combination of the online adaptive methods with an offline configurator to improve the search performance of differential evolution (DE). Along with selecting an AOS method with its parameters, the tuner also decides the parameters of the DE algorithm. We have utilised the framework to find a suitable tuned AOS method to adapt nine commonly used mutation strategies in DE on the \bbob benchmark set. The resultant framework is flexible enough to replace DE with any EA to tune its discrete set of parameters. 
It is build upon the existing classification of AOS method~\cite{fialho2009analysis}. The following three tasks are performed to achieve a unified framework of AOS methods combined with an offline configurator: 
\begin{itemize}
    \item We build upon the existing classification~\cite{di2015experimental} by identifying the new components to simplify the structure of AOS. AOS methods are known to have two major components, credit assignment (assigning reward to an operator) and operator selection (assigning probability to each operator based on quality). We have used this existing classification and further classified these components. The classification is shown in figure~\ref{classification}
    \begin{figure}
    \caption{Adaptive Operator Selection classification}
    \label{classification}
    \centering
    \includegraphics[scale=0.5]{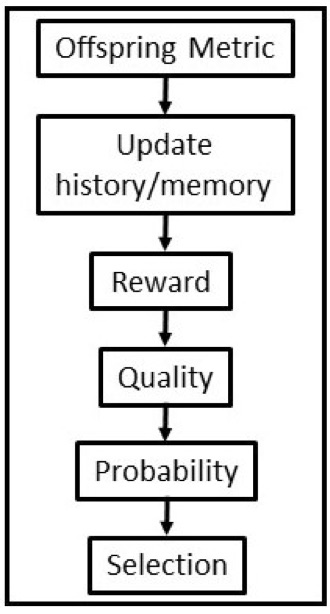}
    \end{figure}
    
    \item A simplified taxonomy is presented that consists of five components with different heuristics as their choices. Thus, the framework consists of an in-depth formulation of AOS components with a generalised structure. In order to make the framework widely applicable we include the choices with diverse properties in each component. Most of the choices are inspired from Evolutionary domain which consists of Reinforcement earning inspired algorithms and thus the architecture can be used in Machine Learning domain as well. We consider immediate progress to far-sighted progress; focusing on the clustered achievement to the outliers. 
    These papers have not explored the possible combinations and their possibility to explore the search space better. There can be many more combinations possible from literature which are never explored and might results in a novel combination that results in much better results. 
    Each AOS component with its choices is shown in figure~\ref{aos_components_choices}
    \begin{figure*}
    \caption{Adaptive Operator Selection component choices} \label{aos_components_choices}
    \centering
    \includegraphics[scale=0.6]{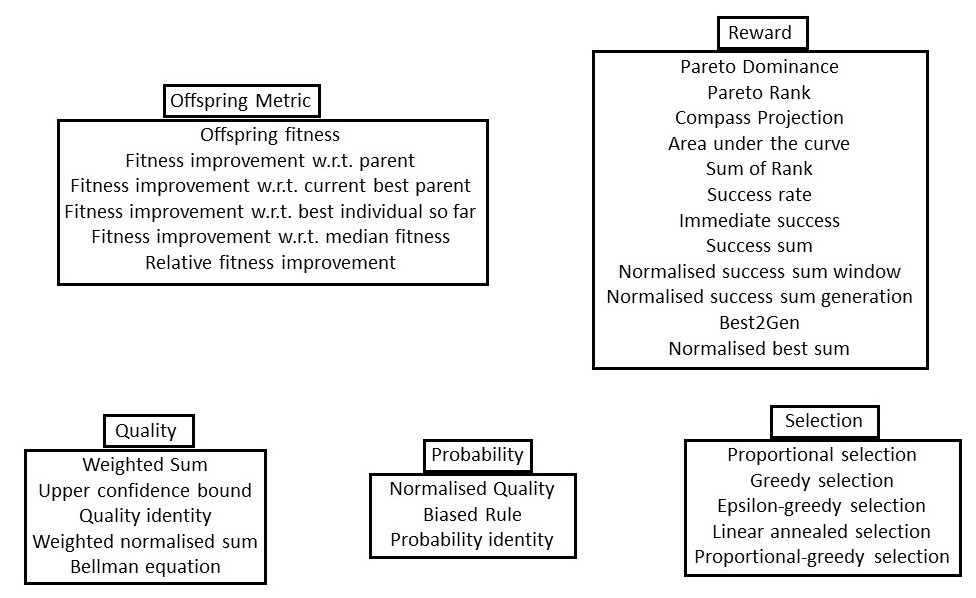}    
    \end{figure*}
    
    \item The resultant framework consists of various AOS designs, out of which one needs to be selected to perform online tuning of parameters in an EA. Thus, a tuner (a meta-leaner) is employed to find a near optimal configuration setting or combination of choices for a given set of problems. A well-known offline configurator known as \irace is used for this purpose. An AOS method has its own parameters to be tuned which do not directly impact the problem solution space. Thus, an offline configurator can be utilised efficiently to give a static value for these hyper-parameters. To tune these hyper-parameters we employ the same tuner in combination with the framework. The role of \irace is to offline select a combination of component choices given a set of problems and tune the hyper-parameters of the selected AOS method along with the parameters of DE.
\end{itemize} 

\begin{figure*}
\caption{Unified Adaptive Operator Selection architecture} \label{fig:training-de-aos}
\centering
\includegraphics[scale=0.4]{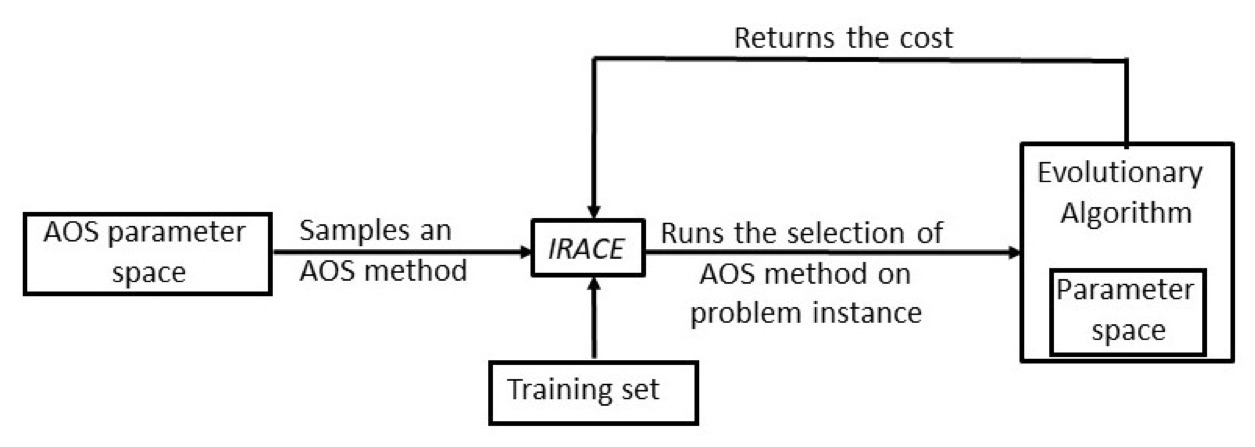}
\end{figure*}

Figure~\ref{fig:training-de-aos} shows the simplified training procedure of the framework using an offline tuner \irace. \irace samples a configuration from the component space and its hyper-parameter space related to the selected choices. It also samples a configuration from the DE parameter space. These choices remain static while the algorithm runs on a selected problem instance selected from a training set. During this run, the AOS method formed by component choices selected by \irace online tunes the mutation strategy of DE. At the end of an EA run, the best seen fitness value is sent to the \irace in the form of cost to make an informed decision on the optimal choices. This describes a single step of \irace. This process is followed repeatedly until a budget given to \irace is exhausted and the configuration that performed best is returned. This configuration consists of a choice from each AOS component, its tuned hyper-parameters and parameter values of DE.  

\section{Preliminaries}
\subsection{Problem definition}
We assume a single-objective problem, where the cost of a solution $x \in X$ is given by $f\colon X \to \mathbb{R}$, which w.l.o.g, must be minimized.

We assume an algorithm with a set of alternative operators $\op \in \Operators$. An operator is a function $\op\colon X \to X$ that takes a number of solutions (parents) and returns one or more solutions (offsprings). 
For instance, crossover operator in Genetic Algorithms is an operator that takes two or more solutions from the parent population to generate offsprings. 

The proposed framework can select different crossover operator as well tune crossover rate. Also, it is independent on the number of solutions the operator deals with. That is, it imposes no restriction on number of solutions used by an operator to generate a certain number of solutions. Thus, this architecture is applicable for any kind of operator which has discrete number of choices, as long as one parent is replaced by one offspring. 

\subsection{Differential Evolution}\label{sec:DE}
To test our architecture we adapt the operators of DE~\cite{PriStoLam2005:book}. It is a population-based algorithm that uses a mutation strategy to create an offspring solution $\vec{u}_i$. A mutation strategy is a linear combination of three or more parent solutions $\vec{x}_i$, where $i$ is the index of a solution in the current population. Some mutation strategies are good at exploration and others at exploitation, and it is well-known that no single strategy performs best for all problems and for all stages of a single run. A survey of development in DE can be found in~\cite{DasMulSug2016de}. In this paper, we consider following frequently used mutation strategies:
\vspace{-1em}
\begin{center}
\begin{tabular}{@{}l@{\hspace{1em}}l}
  \text{``rand/1''}:& $\vec{u}_i = \vec{x}_{r_1} + F \cdot (\vec{x}_{r_2}-\vec{x}_{r_3})$\\
  \text{``rand/2''}:& $\vec{u}_{i}=\vec{x}_{r_1} + F\cdot (\vec{x}_{r_2}-\vec{x}_{r_3} + \vec{x}_{r_4}-\vec{x}_{r_5})$\\
  \text{``rand-to-best/2''}:&  $\vec{u}_{i}=\vec{x}_{r_1} + F\cdot (\vec{x}_\text{best}-\vec{x}_{r_1} + \vec{x}_{r_2}-\vec{x}_{r_3} + \vec{x}_{r_4}-\vec{x}_{r_5})$\\
  \text{``curr-to-rand/1''}:& $\vec{u}_{i}=\vec{x}_{i} + F\cdot (\vec{x}_{r_1}-\vec{x}_{i}+\vec{x}_{r_2}-\vec{x}_{r_3})$\\
  \text{``curr-to-pbest/1''}:& $\vec{u}_{i}=\vec{x}_{i} + F\cdot (\vec{x}_{pbest}-\vec{x}_{i}+\vec{x}_{r_1}-\vec{x}_{r_2})$\\
  \begin{tabular}{@{}c@{}}\text{``curr-to-pbest/1}\\\text{(archived)''}:\end{tabular}& $\vec{u}_{i}=\vec{x}_{i} + F\cdot (\vec{x}_{pbest}-\vec{x}_{i}+\vec{x}_{r_1}-\vec{x}_{archive})$\\
  \text{``best/1''}:& $\vec{u}_{i}=\vec{x}_{best} + F\cdot (\vec{x}_{r_1}-\vec{x}_{r_2})$\\
  \text{``best/2''}:& $\vec{u}_{i}=\vec{x}_{best} + F\cdot (\vec{x}_{r_1}-\vec{x}_{r_2}+\vec{x}_{r_3}-\vec{x}_{r_4})$\\
  \text{``curr-to-best/1''}:& $\vec{u}_{i}=\vec{x}_{i} + F\cdot (\vec{x}_{best}-\vec{x}_{i}+\vec{x}_{r_1}-\vec{x}_{r_2})$\\
\end{tabular}
\end{center}
$F$ is a scaling factor, $\vec{u}_{i}$ and $\vec{x}_{i}$ are the $i$-{th} offspring and parent solutions in the population, respectively, $\vec{x}_\text{best}$ is the best parent in the population, and $r_1$, $r_2$, $r_3$, $r_4$, and $r_5$ are randomly generated mutually exclusive indexes within $[1, NP]$, where $NP$ is the population size. An additional numerical parameter, the crossover rate ($CR \in [0,1]$), determines whether the mutation strategy is applied to each dimension of $\vec{x}_i$ to generate $\vec{u}_i$. At least one dimension of each $\vec{x}_i$ vector is mutated. 

rand/1 and rand/2 are known to explore the search space whereas best/1 and best/2 exploit the neighborhood of the best candidate in the population. curr-to-pbest/1 and curr-to-pbest/1(archived), modifications of curr-to-best/1, are proposed as a novel operator in \JaDE algorithm. We decided to include these mutation strategies involved in \JaDE as they balance the exploration and exploitation of the solution space resulting in competitive performance compared to the other DE variants~\cite{}. We also include two other popular mutation strategies to increase the robustness of the framework. These are curr-to-rand/1 and rand-to-best/2, known to explore the neighborhood of current parent and best parent in the current population respectively.

\section{Components of the proposed framework for AOS}
In this section the components of the proposed framework are discussed in detail. It consists of five components each with a number of generalised choices. The five components are offspring metric ($\Metric$), reward ($r_{g+1,op}$), quality ($q_{g+1,op}$), probability ($p_{g+1,op}$) assignment and selection mechanism ($op_{g+1}$). We consider individual level control that is in a generation an operator is selected for each parent. At the end of the generation, $\Metric$ assigns a value to each offspring according to the improvement gained with the operator application. To prepare the selection of operators for the population in the next generation, reward, quality and probability of each operator are updated, according to the $\Metric$ values. In the end, based on the probability values of each operator, the selection method is invoked to select the operator for each parent to produce offspring. Further, we discuss each of the component choices one by one in detail.

\subsection{Offspring Metric}
We define an $\Metric$ as a function of some statistics on population fitness. The metric, mathematically represented as ($\Metric(g,k,op)$), assigns a $k$-th improved value to the $i$-th improved solution (offspring) $x_{i,op}$ generated after using operator $\op$. If there is no improvement, $0.0$ metric value is assigned to the offspring. That implies that the offspring ($\Metric(g,k,op)$) is as good as its parent.  
$\Metric$ value depends on the parent fitness, the offspring fitness and other significant references shown in table~\ref{OM}. The table shows seven offspring metrics all of which are designed to be maximised when the objective function is a minimisation problem. That is, for an $\Metric(g,k,op)$, the higher the value of $i$-th offspring, the better the offspring is.

We store all $\Metric$ in memory that can be of two types, generation memory and window memory. The generation memory stores six $\Metric$ values for each offspring in each generation. These values are shown in fig.~\ref{OM}. As the algorithm progresses, the size of generation memory grows. A window memory of size $\Window$ is formed using the generation memory. Each entry in the window consists of six values resulting from \Metric. The window memory stores these six metric values of an offspring only if it improves over its parent. In other words, it stores a finite set of $\Metric$ generated by any operator. Initially, the window is filled as the offsprings are generated. Once the window is filled, the new improved offspring generated by an operator is inserted in First In First Out (FIFO) manner such that the offspring from the window generated by the same operator to enter first is removed and new offspring metric data is put at top of the window. If there is no application of that operator present in the window, the worst offspring data is removed. The generation memory and window memory are updated at the end of each generation. The data from generation memory can be utilised either for fix number of generations or for fix number of operator applications described below:

\begin{itemize}
    \item Dynamic number of operator applications ($\maxgen$ as a parameter): The $\Metric$ values produced by an operator in the last fix number of $\maxgen$ generation(s) are taken into account. It is important to note that the number of applications of each operator can vary in each generation. Thus, in $\maxgen$ number of generations the number of total applications of an operator can be different from others. 
    \item Fix number of operator applications ($\fixappl$ as parameter): In this case the last fixed number of operator applications are considered. The generation span depends on the improved offsprings in each generation and is not known in advance.
\end{itemize} 



\begin{table*}[!t]
\renewcommand{\arraystretch}{2.5}
\renewcommand\tabcolsep{1pt}
\caption{Offspring Metrics ($\Metric(g,k,op)$)}\label{OM}
\centering
\begin{tabular}{p{200pt}|c}
\hline
\bfseries Name & \bfseries Definition \\ \hline
Offspring fitness 
& \tableequation{\label{O1} - f(u_{i,op})
}\\\hline


Fitness improvement w.r.t. parent 
& \tableequation{\label{O2}\max\{0, f(x_i) - f(u_{i,op})\}
}\\\hline

Fitness improvement w.r.t. current best parent 
& \tableequation{\label{O3}\max\{0, f_{best} - f(u_{i, op})\}
}\\\hline

Fitness improvement w.r.t. best individual so far
& \tableequation{\label{O4}\max\{0, f_{bsf} - f(u_{i,op})\}
}\\\hline

Fitness improvement w.r.t. median fitness 
& \tableequation{\label{O6}\max\{0, f_{median} - f(u_{i,op})\}
}\\\hline

Relative fitness improvement
& \tableequation{\label{O7}\left(\frac{f_{bsf}}{f(u_{i,op})}\right) \cdot \max\{0, (f(x_i) - f(u_{i,op}))\}
}\\\hline
\end{tabular}
\end{table*}

Equation~\ref{O1}~\cite{lobo1997decision, igel1999using, sakurai2010method, karafotias2014generic, rost2016adaptive, buzdalova2014selecting, karafotias2015evaluating} defines the offspring fitness. For the minimisation problem, this metric assigns the offspring an $\Metric$ as the negative of the raw fitness. In case of the maximisation problem, the raw fitness value should be considered. 


The next four equations~\ref{O2} -- \ref{O6} define the $\Metric$ as the difference between the fitness of an offspring generated by the application of operator $op$ and a reference point. The reference points considered in this study are parent fitness ($f(x_i)$)~\cite{igel1999using, hong2000simultaneously, igel2003operator, fialho2008extreme, fialho2009analysis, pettinger2002controlling, corne1994ga, lobo1997decision, barbosa2000adaptive, maturana2009extreme, maturana2010autonomous, tuson1998adapting, FiaSchoSeb2010toward, maturana2008compass, aleti2015test}, best parent fitness ($f_{best}$)~\cite{igel1999using, igel2003operator, davis1989adapting}, best individual fitness so far ($f_{bsf}$) and median population fitness ($f_{median}$)~\cite{Julstrom1995, julstrom1996inquiry, julstrom1997adaptive}. Getting a value $0$ shows that there is no significant improvement in the offspring from a reference point. These four metrics follow the following rule: the farther the offspring from the reference point, the higher the $\Metric$ will be. 

The offspring fitness improvement shown in equation~\ref{O2} has been most widely used in the literature showing its significance in giving useful information of an operator. Among these four metrics, we propose to include the best seen candidate as a reference point shown in equation \ref{O4}. It is an important reference as it gives a search direction for the exploitation in the neighborhood of the best so far candidate. In this case, an operator that produce an offspring with higher fitness in reference to the best so far candidate gets higher reward value compared to the operator producing offspring with relatively lower improvement.

Relative fitness improvement defined in equation~\ref{O7} was originally proposed in~\cite{ong2004meta} and later used in~\cite{GonFiaCai2010adaptive} as part of an AOS method. It takes into account the fitness improvement from parent to offspring along with the best so far candidate fitness. 

\subsection{Reward}
The reward given to an operator $\op$ at generation $g+1$, represented as $r_{g+1,op}$, gives a measure of achievement of an operator. It goes beyond the current performance of an operator using either generation or memory window. In a run of EA, it is a function of one of the selected $\Metric$. The reward assigned to an operator is maximised as the $\Metric$ is designed to be maximised. That means if an operator has performed better than another operator, then former should get higher reward compared to the latter for any selected $\Metric$. 

We present a generalised classification of the reward definitions from the literature shown in table~\ref{tab:Reward}. They utilise an $\Metric$ definition for an operator in a specifically defined manner. Some make the use of direct fitness values such as diversity and quality, weighted fitness average and best $\Metric$; others using fitness based score such as ranking and count of improved $\Metric$. Reward combines one or two of these statistics, thus learning from limited amount of information available from the fitness landscape.

Some authors consider rewarding ancestors of a well-performing operator~\cite{whitacre2006credit, barbosa2000adaptive} in addition to rewarding the current operator itself; others do not consider such a case~\cite{lobo1997decision}. By ancestor we mean the operators that lead to the good performance of current operator. We decided not to include that option in the classification as~\cite{barbosa2000adaptive} suggests that it sometimes degrades the results. 
In the framework, any method that involves clock time~\cite{sakurai2010method} is replaced by function evaluations in some cases. 
The framework updates reward at the end of each generation that is contrary to the methods that chose to update reward after a certain number of generations~\cite{qian2018differential}. This implies that there is no update on AOS components for few generations because this can lead to loss of information. 
\cite{maturana2010autonomous} introduces the idea of removing and adding operators from a storage called credit registry in an AOS. We do not maintain such a registry in the framework with the intuition that eventually an AOS can learn to select the optimal operator from a list of operators depending on the current stage of an EA. 

The reward choices are divided into the following five categories: fitness diversity and quality, comparison based, successful operator applications, fitness sum and best offspring. The choices under these categories share similar properties. Fitness diversity and quality reward choices are made with the selected $\Metric$ diversity and quality for a fix number of applications, choices under comparison based category involve ranking the $\Metric$ from the window memory, successful operator applications comprise of the number of $\Metric$ resulting from successful operator application, choices in fitness sum simply add the raw $\Metric$ values and lastly best offspring category consists of choices formed with the best solution fitness in a generation combined from a certain number of generations.  

\onecolumn
\begin{longtable}{ p{12.5cm}|p{1.5cm} }
\caption{Reward ($r_{g+1,op}$)}\label{tab:Reward}
\\ \hline
\multicolumn{1}{c}{\textbf{Definition}} & \multicolumn{1}{|c}{\textbf{Parameters}} \\ \hline

\multicolumn{2}{c}{\textbf{Fitness Diversity and Quality}}\\ \hline

\begin{tabular}{@{}c@{}}Pareto Dominance\\\vbox{\begin{equation}\label{DQ1} \frac{\mbox{PD } (\oopfixappl)}{\sum_{j=1}^{K}\mbox{PD } (o_j^{\fixappl})} 
\end{equation}}
\end{tabular} 
& $\fixappl$ \\ \hline

\begin{tabular}{@{}c@{}}Pareto Rank\\\vbox{\begin{equation}\label{DQ2} \frac{\mbox{PR } (\oopfixappl)}{\sum_{j=1}^{K}\mbox{PR } (o_j^{\fixappl})} \end{equation}} 
\end{tabular} & $\fixappl$ \\ \hline

\begin{tabular}{@{}c@{}}Compass Projection\\\vbox{\begin{equation}\label{DQ3} |\oopfixappl| \cdot \cos(\alpha_{op}) - \min_{j = 1}^K |\oopfixappl| \cdot \cos(\alpha_{j})\end{equation}} 
\end{tabular} & \begin{tabular}{@{}c@{}}$\theta$,\\$\fixappl$\end{tabular} \\ \hline

\multicolumn{2}{c}{\textbf{Comparison (Rank)}}\\ \hline

\begin{tabular}{@{}c@{}}Area Under the Curve\\\vbox{\begin{equation}\label{R1} \mbox{Area under the curve based on the rank of } \Metric(g,k,op) \end{equation}} 
\end{tabular} & $D$,$\Window$ \\ \hline

\begin{tabular}{@{}c@{}}Sum of Rank\\\vbox{\begin{equation}\label{R2} \frac{\sum_{r(\Metric(g,k,op))} D^{r(\Metric(g,k,op))} (\Window-r(\Metric(g,k,op)))}{\sum_{r=1}^{\Window} D^{r(\Metric(g,k,op))} (\Window-r(\Metric(g,k,op)))} \end{equation}} 
\end{tabular} & $D$,$\Window$ \\ \hline

\multicolumn{2}{c}{\textbf{Successful operator applications}}\\ \hline

\begin{tabular}{@{}c@{}}Success Rate\\\vbox{\begin{equation}\label{SR1}\sum_{t = g}^{g-(\maxgen-1)}\left(\frac{{(\nsuccopt)}^\gamma + Frac * \sum_{j=1}^{K} n_{succ, j}^t}{\nsuccopt + \nfailopt}\right) + \epsilon\end{equation}} 
\end{tabular}& \begin{tabular}{@{}c@{}}$\maxgen, \epsilon$\\$\gamma$,$Frac$\end{tabular} \\ \hline

\begin{tabular}{@{}c@{}}Immediate Success\\\vbox{\begin{equation}\label{SR2} \frac{n_{succ, op}^g}{NP} \end{equation}} 
\end{tabular} & - \\ \hline

\multicolumn{2}{c}{\textbf{Fitness Sum}}\\ \hline

\begin{tabular}{@{}c@{}}Success Sum \\ \vbox{\begin{equation}\label{W1}\frac{\sum_{t=g}^{g-(\maxgen-1)} \sum_{i=1}^{\nsuccopt} \Metric(t,i,op)}{\sum_{t=g}^{g-(\maxgen-1)} \nsuccopt + \nfailopt} \end{equation}} 
\end{tabular}& $\maxgen$\\ \hline

\begin{tabular}{@{}c@{}}Normalised Success Sum Window \\ \vbox{\begin{equation}\label{W2}\frac{\frac{\sum_{k=1}^{n_\op} \Metric(t,k,op)}{n_\op}}{\left(Best_{j=1}^{K} \frac{\sum_{k=1}^{n_j} \Metric(t,k,j)}{n_j} \right)^\omega} \end{equation}} 
\end{tabular}& $\omega$,$\Window$ \\ \hline

\begin{tabular}{@{}c@{}}Normalised Success Sum Generation\\\vbox{\begin{equation}\label{W3} \sum_{t = g}^{g-(\maxgen-1)} {\frac{\sum_{i=1}^{\nsuccopt} \Metric(t,i,op)}{\nsuccopt + \nfailopt}} \end{equation}} 
\end{tabular}& $\maxgen$ \\ \hline\hline

\multicolumn{2}{c}{\textbf{Best offspring}}\\ \hline

\begin{tabular}{@{}c@{}}Best2Gen\\\vbox{\begin{equation}\label{B1} C * \frac{Best_{i=1}^{n_{succ,op}^g} \Metric(g,i,op) - Best_{i=1}^{n_{succ,op}^{g-1}} \Metric(g-1,i,op)}
{(Best_{i=1}^{n_{succ,op}^{g-1}} \Metric(g-1,i,op))^\alpha * |(n_{succ,op}^g + n_{fail,op}^g) - (n_{succ,op}^{g-1} + n_{fail,op}^{g-1})|^\beta} \end{equation}} 
\end{tabular}& \begin{tabular}{@{}c@{}}$C$\\$\beta,\alpha$\end{tabular} \\ \hline

\begin{tabular}{@{}c@{}}Normalised Best Sum\\\vbox{\begin{equation}\label{B2}\frac{\frac{1}{\maxgen}\sum_{t=g}^{g-(\maxgen-1)} {Best_{i=1}^{n_{succ,op}^t} \Metric(t,i,op)}^\rho}{(Best_{j=1}^{K} \{\sum_{t=g}^{g-(\maxgen-1)} Best_{i=1}^{n_{succ,k}^t} \Metric(t,i,j)\})^\alpha} \end{equation}} 
\end{tabular}& \begin{tabular}{@{}c@{}}$\alpha,\rho$\\ $\maxgen$ \end{tabular} \\ \hline
\end{longtable}

\twocolumn
\textbf{Fitness Diversity and Quality} 
This category includes the diversity (standard deviation) and quality (average) over an $\Metric$, combined in different manner. The three definitions consider fix number of operator applications, $\fixappl$, extracted from generation memory. These two statistics are represented as a coordinate ($\oopfixappl$) in two dimensional space for each operator $\op$ shown in the equation below:
\begin{equation}\label{coor} 
\begin{split}
\oopfixappl &= (div_{i=k}^{k-({\fixappl}-1)} \Metric(g,i,op),\\
            &\qquad qual_{i=k}^{k-(\fixappl-1)} \Metric(g,i,op)) 
\end{split}
\end{equation}
The diversity $div$ and quality $qual$ of $\Metric$ for an operator $\op$ are calculated for last $\fixappl$ number of applications.

Pareto Dominance (PD) on ($\oopfixappl$) shown in equation~\ref{DQ1}~\cite{maturana2010autonomous} counts the number of operators that are dominated by \op. It is normalised by the sum of $k$ operators' reward values. 
The best operator corresponds to the highest value of PD indicating that it generated maximum number of offspring in last certain number of generations compared to other operators.

Pareto Rank (PR) on (\oopfixappl) shown in equation~\ref{DQ2}~\cite{maturana2010autonomous} method counts the number of operators that dominate \op. The operator with the least PR value is the best of all operators. Operators with a PR value of $0$ belong to the Pareto frontier. Both PD and PR encourage non-dominated operators.

In equation~\ref{DQ3}, proposed in~\cite{maturana2008compass}, the projection of a coordinate \oopfixappl is taken on a plane represented by an angle $\theta$ (a hyper-parameter). This angle defines the trade-off between exploration and exploitation. Thus, the mathematical formulation of projection is given by $|\oopfixappl| \cdot \cos(\alpha_{\op})$ where, 
%

\begin{equation}
\begin{split} 
|\oopfixappl| &= \biggl\{({div_{i=k}^{k-(\fixappl - 1)} \Metric(g,i,op)})^2 \\
                &\qquad+ ({qual_{i=k}^{k-(\fixappl - 1)} \Metric(g,i,op)})^2\biggl\}^{1/2}
\end{split}
\end{equation}
\begin{equation} 
\alpha_{\op} = atan\left(\frac{qual_{i=k}^{k-(\fixappl - 1)} \Metric(g,i,\op)}{div_{i=k}^{k-(\fixappl - 1)} \Metric(g,i,\op)}\right) - \theta \mbox{ and }
\end{equation}
$\alpha_{\op}$ is the angle between the plane and the coordinate. Compass evaluates the performance of operators by considering not only the fitness improvements from parent to offspring, but also the way they modify the diversity of the population. This was later combined with dynamic multi-armed bandit~\cite{maturana2009extreme, maturana2010autonomous} for adaptive selection of the operators in differential evolution.

\textbf{Comparison (Rank)}
The two definitions in this category assign a rank to each operator in the window of size \Window according to \Metric values. These ranks are then decayed using hyper-parameter $D$ to prioritise the operators according to their ranks. As ranking of \Metric is involved and not the direct \Metric values, both methods under this category are invariant with respect to the linear scaling of the fitness function. That is, their behavior, when applied on a given fitness function $f$, is exactly the same when applied to a fitness function defined by $(a \cdot f)$, for any $a > 0$.

Area Under the Curve (AUC)~\ref{R1}~\cite{FiaSchoSeb2010toward} plots a Receiver Operator Characteristic (ROC) curve for each operator by scanning the decayed ranking of \Metric from a window memory. The area is taken as a reward of each operator.

Sum of rank (SR)~\ref{R2}~\cite{FiaSchoSeb2010toward} assigns the operators with the sum of the decayed ranks of the \Metric in the window memory. This sum is normalised by the reward sum of all operators. 

\textbf{Successful operator applications}
Under this category, the number of successful and unsuccessful applications of each operator are considered. We design two definitions under this category, both using generation memory. For a particular \Metric under consideration, we count the total number of non-zero values (\nsuccopg) as the successful applications of operator \op in generation $g$. Similarly, the number of zeros shows the number of unimproved applications (\nfailopg) of operator \op in generation $g$. These counts are recorded for a fix number of generations represented by hyper-parameter \maxgen. Thus there are \maxgen number of values each representing number of successful and unsuccessful applications coming from last \maxgen generations. Mathematically, the number of successes and failures of an operator \op in generation $g$ of population size $NP$ is defined as follows:
\begin{equation*} 
\nsuccopg = \sum_{i=1}^{NP}\begin{cases} 1, & \quad\text{ if } \Metric(g,i,op) > 0\\ 0, & \quad\text{ otherwise } \end{cases} 
\end{equation*}
\begin{equation*} 
\nfailopg = \sum_{i=1}^{NP}\begin{cases} 1, & \quad\text{ if } \Metric(g,i,op) = 0\\ 0, & \quad\text{ otherwise } \end{cases} 
\end{equation*}

To calculate the reward of an operator, equation~\ref{SR1}~\cite{Julstrom1995} takes into account a fraction ($Frac$) of sum of successes of all operators in a generation along with the linear or quadratic contribution of the success of an operator (power is a parameter represented as $\gamma$) in the same generation. This term is divided by the total number of applications of the operator in that generation. The total number of applications of an operator in a generation is the total number of successes and failures seen by the operator. This fraction is aggregated for the last \maxgen generations. In the end, an error value ($\epsilon$) is added that perturbs the resultant reward value. 
Equation~\ref{SR1} is part of the method proposed in~\cite{qin2009differential, qin2005self} and the potential choices of power for $\gamma$ come from \cite{niehaus2001adaption, qian2018differential}.
Another research comes from paper~\cite{consoli2014diversity} that utilises a similar choice. It proposes MAENSm method that selects a crossover operator among a set of operators.
The second term in the numerator of this choice is coming from extension of the ADOPP algorithm in~\cite{julstrom1997adaptive}.

Next equation~\ref{SR2}, modified from paper~\cite{MudLopKaz2018ppsn}, ignores the achievements of the operator in the past. The complete reward definition in the paper can be derived as a combination of \ref{SR2} and \ref{Q8}. It is a simple idea defined as the fraction of the immediate or current success with respect to the population size. Here, current success refers to the number of improved \Metric applications of an operator in the current generation. 

\textbf{Fitness Sum}
In the previous two categories, the reward definitions have utilised metric values for ranking and calculating successes and failures. In further categories, we will use direct values of metrics to calculate reward. 

In fitness sum category, we took three definitions from literature. Equations~\ref{W1}~\cite{igel2003operator} and~\ref{W3}~\cite{igel1999using} sum the $\Metric$ values in the last $\maxgen$ generations from the generation window. The only difference between them is that in the former, once the $\Metric$ data from all $\maxgen$ is summed, it is divided by the number of applications in all $\maxgen$ generations whereas in the latter this division is performed per generation.
\cite{vafaee2008dynamic} and~\cite{hong2000simultaneously} considers~\ref{W1} for one generation (that is $\maxgen = 1$). The only difference is that the latter does not include the denominator part.

Equation~\ref{W2} uses the data stored in the window memory. It simply sums the $\Metric$ values for an operator present in the window of size $\Window$ divided by its number of applications of the operator present in the window. As the window comprises of successful applications, $n_op$ denotes the number of applications of \op present in the window. This definition for this choice is proposed by~\cite{GonFiaCai2010adaptive} known as Average Absolute Reward (AAR). Average Normalised Reward (ANR) proposed in the same research normalises AAR by the best AAR seen by any operator. Thus, we give the normalisation as a choice decided by a hyper-parameter $\omega$. 

\textbf{Best Offspring} 
This category has definitions which consider outliers, that is, it takes into account the \OM generated by \op that have given best or extreme performance 
$Best\mbox{}\OM(g,i,\op)$ in a generation $g$. 
This metric value is selected among the successful application from \op, \nsuccopg. It can easily get trapped in local optima if the landscape is too ragged. 

RL-based adaptive methods could not be part of framework fully, as their design is different from AOS in general. RL methods have a concept to reward the operators that produce good offsprings. We have included these reward definitions in the framework that are based on the best generated offspring in a generation and Best2Gen~\ref{B1} is the only reward definition inspired from RL design. It is commonly used within the RL design to learn the selection of the operator for each parent. It takes the difference of best seen $\Metric$ by an operator in the two consecutive generations~\cite{lobo1997decision,igel1999using,buzdalova2014selecting}. The two terms in the denominator, separated by product, are considered by one RL method but discarded by other. Thus, to achieve a general equation we decided to include them with the decision hyper-parameters $\alpha$ and $\beta$. \cite{rost2016adaptive, sakurai2010method} consider the best seen $\Metric$ produced by an operator in the previous generation $g-1$ along with the difference in the numerator. A hyper-parameter $C$ is multiplied and divided by the difference in applications of operator in the last two generations~\cite{karafotias2015evaluating, karafotias2014generic}. 
 
Based on the similar idea, equation~\ref{B2} has been used in non-RL context. It uses the best $\Metric$ value seen in the last $\maxgen$ number of generations generated by an operator~\cite{fialho2008extreme}. Thus, it not only looks for the best candidate produced by an operator in current and last generation but is far-sighted to combine best fitness in certain number of generations. The contribution of best seen value in a generation is either linear or quadratic decided by hyper-parameter $\rho$ in~\cite{vafaee2008dynamic}. Optionally, it can be normalised~\cite{GonFiaCai2010adaptive} by best $\Metric$ value seen by any operator in last $\maxgen$ generations decided by $\alpha$. $\alpha = 0, 1$ corresponds to extreme absolute reward and extreme normalised reward respectively. We extend the resultant equation by multiplying it with $\frac{1}{\maxgen}$~\cite{igel1999using}.
Both equations in this category have a decision parameter $\alpha$.

\subsection{Quality}
Assigning quality to each operator is an important task involved in AOS method. A quality definition is dependent on current reward and can also include any of the following: reward and quality achieved in previous generation. Thus, to calculate quality, we keep a memory of reward and quality from the previous generation. 
In the end the quality values are normalised to prevent probability to explode or go out of range.
Table~\ref{tab:quality} shows five choices for quality.

\begin{table*}[!t]
\renewcommand{\arraystretch}{1.3}
\renewcommand\tabcolsep{1pt}
\caption{Quality ($q_{g+1,op}$)}\label{tab:quality}
\centering
\begin{tabular}{p{12.0cm}|c}
\hline
\multicolumn{1}{c}{\textbf{Definition}} & \multicolumn{1}{|c}{\textbf{Parameters}} \\ \hline

\begin{tabular}{@{}c@{}}Weighted Sum\\\vbox{\begin{equation}\label{Q1} \delta * r_{g+1,op} + (1 - \delta) * q_{g,op} \end{equation}} 
\end{tabular} & $\delta$ \\ \hline

\begin{tabular}{@{}c@{}}Upper Confidence Bound\\\vbox{\begin{equation}\label{Q2} r_{g+1,op} + C \cdot \sqrt{\frac{\log{\sum_{j = 1}^{K}n_j}}{n_{op}}} \end{equation}} 
\end{tabular} & $C$ \\ \hline

\begin{tabular}{@{}c@{}}Quality Identity\\\vbox{\begin{equation}\label{Q4} r_{g+1,op} \end{equation}} 
\end{tabular} & - \\ \hline


\begin{tabular}{@{}c@{}}Weighted Normalised sum\\\vbox{\begin{equation}\label{Q7} \delta * \max\left\{\qmin, \frac{r_{g+1,op}}{\sum_{j=1}^K r_{g+1,j}}\right\} + (1-\delta) * q_{g,op}  \end{equation}} 
\end{tabular} & $\delta, \qmin$ \\\hline

\begin{tabular}{@{}c@{}}Bellman equation\\\vbox{\begin{equation}\label{Q80} (1-\gamma P)^{-1} Q'_{g+1}\end{equation}}\\\vbox{\begin{equation}\label{Q8}\mbox{where, }q'_{g+1,op} = c_1 * r_{g+1,op} + c_2 * r_{g,op}\end{equation}} 
\end{tabular} & $c_1, c_2, \gamma$ \\ \hline

\end{tabular}
\end{table*}

\textbf{Weighted Sum}
Equation~\ref{Q1} in table~\ref{tab:quality} is the weighted sum of current reward and previous quality. This is part of Probability matching which is originally proposed in~\cite{Gold1990pm} and later used as operator selector in AOS~\cite{lobo1997decision, Thierens2005:gecco, Thie2007adaptive}. This is the commonly used quality choice in the literature to assign the quality for the adaption of parameters~\cite{fialho2008extreme, gong2011adaptive, fialho2009analysis}.

\textbf{Weighted Normalised Sum}
Equation~\ref{Q7} has the same definition as equation~\ref{Q1} except that the current reward is normalised by the sum of the reward values of all operators to bring the value in the range [$0$, $1$]. This is obtained directly from~\cite{igel1999using} which involves two hyper-parameters $\qmin$ and $\delta$. This definition is lower bounded by $\qmin$. \cite{igel2003operator} does not consider a lower bound in the quality, that is $\qmin = 0$ given that the reward is positive real value. In case where a reward attained by an operator is zero, the quality becomes a fraction of previous quality. 
The parameter ($\delta$) in equations~\ref{Q1} and~\ref{Q7} plays the role to act as a weight for current/normalised reward and previous quality. 

\textbf{Upper confidence Bound}
(UCB) is a well-known algorithm, originally proposed in~\cite{AueCesFis2002finite}. It is known to achieve a compromise between exploitation and exploration. 
\cite{dacosta2008adaptive} proposes dynamic multi-armed bandit (DMAB), a selection strategy based on UCB. To control the exploration strength, it includes a hyper-parameter $C$ in UCB as shown in equation~\ref{Q2}. This definition has only been used for window memory where $n_j$ represents the total number of applications of $j$-th operator present in the window. However, we have made this definition flexible enough such that if a reward choice utilising generation memory is selected, $n_{succ,j}^t$ counts the successful operator applications in generation $t$ by operator $j$. The formulae calculating UCB value for \maxgen generations is shown below:
\begin{equation}
    r_{g+1,op} + C \cdot \sqrt{\frac{\log{\sum_{j = 1}^{K}\sum_{t=g}^{g-(\maxgen - 1)} n_{succ,j}^t}}{\sum_{t=g}^{g-(\maxgen - 1)} \nsuccopt}} 
\end{equation}
In the case of fitness diversity and quality reward definitions, $\fixappl$ value for an \op is same as $n_\op$. 
This method is also used in~\cite{maturana2009extreme, fialho2008extreme, fialho2009analysis, FiaSchoSeb2010toward, maturana2010autonomous} either within DMAB or for comparison. 
UCB variants such as UCB-Tuned~\cite{AueCesFis2002finite} and KUCBT~\cite{hussain2006exploration} are not included as part of the UCB definition to keep the formula simple.

\textbf{Quality Identity}
Looking deeply into the AOS methods~\cite{qin2009differential, qian2018differential, corne1994ga, barbosa2000adaptive, niehaus2001adaption, maturana2008compass, maturana2010autonomous, julstrom1997adaptive, Julstrom1995}, we identified some existing methods have mixed the definitions of reward and quality. There is no quality component that shares the properties with any of the quality definitions in table~\ref{tab:quality}. Rather these methods could simply be divided into AOS without quality. For instance, SaDE~\cite{qin2009differential} assigns probability to each operator according to the normalised success rate for \maxgen~\ref{SR1} and there is no quality definition involved. That is it directly maps success rate reward value of an operator to its probability. Thus, equation~\ref{Q4} represents an identity function that maps reward of an operator directly to its quality. In this definition, we do not include previous reward or quality that helps to clearly distinguish the reward from probability definition. It becomes an important choice to determine whether quality has important role in the AOS process. 

\textbf{Bellman Equation}
Equation~\ref{Q80}~\cite{MudLopKaz2018ppsn} represents the bellman equation where each entry in vector Q' is the weighted sum of the reward values of an operator from previous generations shown in~\ref{Q8}~\cite{vafaee2008dynamic}. The hyper-parameters $c_1$ and $c_2$ denote the weights of these rewards. The adaptive method in~\cite{lobo1997decision} considers the sum of $c_1$ and $c_2$ to be $1$. We lift this condition and each of them can attain a value between $0$ and $1$. It should be noted that if hyper-parameter $\gamma = 0$ then this definition reduces to just weighted sum of rewards otherwise the bellman equation is used to calculate the final quality.

\subsection{Probability}
The three probability definitions shown in table~\ref{tab:probability} are used to assign probability to each operator to get selected in the next generation. These definitions use the current quality of operator and optionally previous probability. Each of the probability choice is lower bounded by a minimum probability of selection \pmin to avoid probability of any operator becoming zero. An operator showing weak performance in current generation can become useful in later generation. Thus, \pmin plays important role in allowing an operator to get selected after a certain number of generations. 
\cite{corne1994ga, igel2003operator, whitacre2006credit}  
consider generation gap in updating operator probability. That means the probability is updated after certain number of generations. However, in the current framework we have not included this case for simplicity. Thus, we update the probability of each operator at the end of each generation. This helps the method to be up-to-date with the operator performance according to the current landscape.
We also eliminate the case where each candidate solution in the population is assigned with the probabilities of getting selected by each operator in the population~\cite{stanczak1999genetic}. Instead, we assign probability to each operator and employ a selection mechanism to select an operator for each offspring based on the selection probability of the operator. All the probability definitions are normalised in the end to bring the sum of the probabilities of all operators to $1$. 

\begin{table*}[!t]
\renewcommand{\arraystretch}{1.3}
\renewcommand\tabcolsep{1pt}
\caption{Probability ($p_{g+1,op}$)}
\label{tab:probability}
\centering
\begin{tabular}{p{12.0cm}|c}
\hline
\multicolumn{1}{c}{\textbf{Definition}} & \multicolumn{1}{|c}{\textbf{Parameters}} \\ \hline

\begin{tabular}{@{}c@{}}Normalised Quality \\\vbox{\begin{equation}\label{S1} \pmin + (1 - K * \pmin) \left(\frac{q_{g+1,op} + \epsilon_p}{\sum_{j=1}^{K}q_{g+1,j} + \epsilon_p}\right)\end{equation}}
\end{tabular} & $\pmin,\epsilon_p$ \\ \hline

\begin{tabular}{@{}c@{}}Biased rule\\\vbox{\begin{equation}\label{S2} \begin{cases} \mu * \pmax + (1-\mu) * p_{g,\op}, & for \mbox{ } op = max_{j=1}^K \{q_{g+1,j}\} \\ \mu * \pmin + (1-\mu) * p_{g,j}, & \forall j \neq \op \end{cases}\end{equation}} 
\end{tabular} & \begin{tabular}{@{}c@{}}$\mu$,\\$\pmin$,\\$\pmax$\end{tabular} \\ \hline


\begin{tabular}{@{}c@{}}Probability Identity\\\vbox{\begin{equation}\label{S3} q_{g+1, op}\end{equation}} 
\end{tabular} & - \\ \hline 

\end{tabular}
\end{table*}

\textbf{Normalised quality} shown in equation~\ref{S1} is the most widely used probability definition in the literature to assign the selection probability to each operator. It is part of Probability matching (PM) originally proposed in~\cite{Gold1990pm}. \cite{Thierens2005:gecco, gong2011adaptive, igel2003operator, Thie2007adaptive,  qian2018differential,  fialho2008extreme, GonFiaCai2010adaptive, MudLopKaz2018ppsn, niehaus2001adaption, fialho2009analysis, maturana2010autonomous, fialho2008extreme, FiaSchoSeb2010toward, lobo1997decision, davis1989adapting, Julstrom1995, corne1994ga} used normalised value of quality, lower bounded by $\pmin$. \cite{barbosa2000adaptive,qin2009differential, julstrom1997adaptive, maturana2008compass, vafaee2008dynamic} used the normalised value of quality but with $\pmin = 0$ and the latter two papers also added an error value ($\epsilon_p$) to this quantity. The error value is used to prevent the quality to become $0$. A generalised form of these is shown in~\ref{S1} with two hyper-parameters, $\pmin$ and $\epsilon_p$. In the term $(1-K*\pmin)$, $K$ denotes the total number of operators. The value for $K*\pmin$ should be less than $1$ to avoid this term becoming negative. Thus, $\pmin$ is dependent on the number of operators employed.

Normalised quality has a disadvantage that it allocates the probabilities to the operators directly proportional to the quality which makes the convergence slow and prevents the exploitation of the operator with maximum probability. To overcome this issue, \textbf{biased rule}~\ref{S2}~\cite{Thierens2005:gecco} was proposed as part of the Adaptive Pursuit algorithm. It increases the probability of selection of the operator with best quality. This is done by assigning probability to this operator as the weighted sum of the upper bound of probability ($\pmax$) and previous probability of the operator. To maintain a large gap between best operator and others, biased rule assigns the latter operators with the weighted sum of $\pmin$ and previous probability. It is interesting to note that this definition does not directly include qualities of operators in calculating the probability of an operator. There are three hyper-parameters involved in the biased rule probability definition, namely $\pmin$, $\pmax$ and $\mu$ with $\pmin < \pmax$. This definition is also utilised in~\cite{Thie2007adaptive, fialho2008extreme}.

\textbf{Probability Identity} The last proposed probability definition is shown in equation~\ref{SL5}. It simply maps the quality of an operator to its probability obtained in the current generation.



\subsection{Selection}
The selection component consists of various choices that are used to select the operator for an individual in the population given the selection probability of each operator. The most commonly used selection method is proportional selection which is popular in Probability Matching technique. Adaptive Pursuit is the only method that utilises greedy selection. In addition to proportional and greedy selection, we present three proposed selection choices based on the combinations of greedy, proportional and linear decay. 

\begin{table*}[!t]
\renewcommand{\arraystretch}{1.5}
\caption{Selection(\op)}\label{tab:selection}
\centering
\begin{tabular}{p{2.1cm}|p{10cm}|c}
\hline
\multicolumn{1}{l|}{\textbf{Name}} & \multicolumn{1}{c}{\textbf{Definition}} & \multicolumn{1}{|c}{\textbf{Parameter}} \\ \hline
Proportional 
&\tableequation{\label{SL1} \frac{p_{g+1,\op}}{\sum_{j=1}^{K}p_{g+1,j}} } & - \\\hline
Greedy 
&\tableequation{\label{SL2} \max_{j=1}^{K}\{p_{g+1,j}\} } & - \\\hline
Epsilon-Greedy 
&\tableequation{\label{SL3} \begin{cases} rand[1,K] , & if\mbox{ }uniform(0, 1) < eps\\ \max_{j=1}^{K}\{p_{g+1,j}\} , & else \end{cases}} & $eps$ \\\hline
Linear-Annealed 
&\tableequation{\label{SL4} \begin{cases} rand[1,K] , & if\mbox{ }uniform(0,1) < Anneal_{eps}(0, 1)\\ \max_{j=1}^{K}\{p_{g+1,j}\} , & else \end{cases}} & - \\\hline
Proportional-Greedy 
&\tableequation{\label{SL5} \begin{cases} \frac{p_{g+1,\op}}{\sum_{j=1}^{K}p_{g+1,j}} , & if\mbox{ }uniform(0,1) < eps\\ \max_{j=1}^{K}\{p_{g+1,j}\} , & else \end{cases}} & $eps$ \\
\bottomrule
\end{tabular}
\end{table*}

\textbf{Proportional Selection} also known as roulette-wheel selection is shown in equation~\ref{SL1}. The normalised probability of an operator defines its chances of getting selected proportional to its quality in the next generation. That is, the operator with high normalised value has greater chance to get selected compares to the other operators.

\textbf{Greedy Selection} As the name indicates, in~\ref{SL2} the operator with the maximum probability is selected in the next generation. This choice is known to bring least exploration in the task.

\textbf{Epsilon-Greedy Selection} Greedy selection has a drawback that it neglects the exploration aspect which plays an important role in searching the unexplored parts of the search space. Thus, we introduced a novel selection choice named epsilon greedy selection~\ref{SL3}. The hyper-parameter $eps$ ensures that a random strategy (rand[$1$,$K$]) is selected (exploration). $uniform(0,1)$ indicates a uniformly selected number between $0$ and $1$.

\textbf{Linear-Annealed Selection} The epsilon-greedy definition in~\ref{SL4} keeps the $eps$ value static during the whole run of an EA. However, adapting $eps$ can ensure a right balance between exploration and exploitation. This is so because at different stages of algorithm, importance of exploration and exploitation varies. The higher the value of $eps$, the greater the exploration will be. Thus, we propose a linear decay of $eps$ where vaue of $eps$ decreases from $1$ to $0$ as the algorithm progresses represented by $Annealed_{eps}(0,1)$. Thus, this ensures that algorithm explores in the initial runs and exploits towards the end.

\textbf{Proportional-Greedy Selection} This selection choice~\ref{SL5} is a hybrid of proportional and greedy selection. If a random number between $0$ and $1$ is smaller than the hyper-parameter, $\epsilon$, proportional selection is performed otherwise greedy is performed. Here the random operator is not selected for exploration but the one that has shown good performance (and not necessarily the best) recently. This brings a restricted exploration of the search space compared to the equation~\ref{SL3}.

\section{AOS methods utilised to build the framework}
In the previous section we presented the proposed framework utilising various AOS methods from literature. The component choices sharing same properties are generalised with the introduction of a number of hyper-parameters. It is possible to replicate many AOS methods existing in literature using the designed framework. A particular AOS method is obtained by setting the component choices along with their hyper-parameter values. The table~\ref{tab:AOS from literature} shows the AOS methods from the literature that can be replicated from the unified framework. The table shows a method as a combination of component choices from the framework with their hyper-parameter values as proposed in the literature. The first column gives the method name and the rest of the columns indicate a choice from each component setting their hyper-parameter values as decided in their respective papers. We decided to name some algorithms if they are not named in the paper they are proposed. These names are chosen to best suit the description of the algorithm.

\onecolumn
\begin{longtable}{ p{2.5cm}|p{0.4cm}|p{4.0cm}|p{3.0cm}|p{3.5cm}|p{1.2cm} }
\caption{Relevant literature}\label{tab:AOS from literature}
\\ \hline
\bfseries AOS Methods & \bfseries $\Metric$ & \bfseries Reward & \bfseries Quality & \bfseries Probability & \bfseries Selection \\ \hline 
Hybrid
& \ref{O1} & \ref{B1} ($\alpha = \beta =0$,$C = 1$) & \ref{Q80} ($c_1 + c_2 = 1, \gamma = 0$) & \ref{S1} ($\pmin = \epsilon_p = 0$) & \ref{SL1} \\ \hline 

Op-adapt
& \ref{O3} & \ref{W3} & \ref{Q7} & \ref{S3} & \ref{SL1} \\ \hline

PDP
& \ref{O2} & \ref{SR1} ($\gamma = 2$, $\maxgen = 1$, $Frac = \epsilon = 0$) & \ref{Q4} & \ref{S1} ($\epsilon_p = 0, p_{min} = \left\lfloor\frac{20}{K}\right\rfloor$) & \ref{SL1} \\ \hline 

ADOPP
& \ref{O6} & \ref{SR1} ($\epsilon = 0$, $\gamma = 1$,$\maxgen = 1$, $Frac = 0$) & \ref{Q4} & \ref{S1} ($\pmin = \epsilon_p = 0$) & \ref{SL1} \\ \hline 
ADOPP-ext
& \ref{O6} & \ref{SR1} ($\epsilon = 0$,$\gamma = 1$, $\maxgen = 1$) & \ref{Q4} & \ref{S1} ($\pmin = \epsilon_p = 0$) & \ref{SL1} \\ \hline 

Adapt-NN
& \ref{O2} & \ref{W1} & \ref{Q7} ($\qmin = 0$) & \ref{S1} ($\epsilon_p = 0$) & \ref{SL1} \\ \hline 

Dyn-GEPv1
& \ref{O2} & \ref{W1} ($\maxgen = 1$) & \ref{Q80} ($c_1 = 1, \gamma = 0$) & \ref{S1} ($\pmin = 0$) & \ref{SL1} \\ \hline 
Dyn-GEPv2
& \ref{O2} & \ref{B2} ($\rho = 3$, $\alpha = 0$,$\maxgen = 1$) & \ref{Q80} ($c_1 = 1, \gamma = 0$) & \ref{S1} ($\pmin = 0$) & \ref{SL1} \\ \hline 

SaDE
& \ref{O2} & \ref{SR1} ($\gamma = 1$,$Frac = 0$) & \ref{Q4} & \ref{S1} ($\pmin = 0, \epsilon_p = 0$) & \ref{SL1} \\ \hline 

MMRDE
& \ref{O2} & \ref{SR1} ($\maxgen = \gamma = 1$,$Frac = \epsilon = 0$) & \ref{Q4} & \ref{S1} ($\epsilon_p = 0$) & \ref{SL1} \\ \hline 

Compass
& \ref{O2} & \ref{DQ3} & \ref{Q4} & \ref{S1} ($\pmin = 0$) & \ref{SL1} \\ \hline 

PD-PM
& \ref{O2} & \ref{DQ1} & \ref{Q1} & \ref{S1} ($\epsilon_p = 0$) & \ref{SL1} \\ \hline 
PR-PM
& \ref{O2} & \ref{DQ2} & \ref{Q1} & \ref{S1} ($\epsilon_p = 0$) & \ref{SL1} \\ \hline 
Proj-PM
& \ref{O2} & \ref{DQ3} & \ref{Q1} & \ref{S1} ($\epsilon_p = 0$) & \ref{SL1} \\ \hline 

F-AUC-MAB
& \ref{O2} & \ref{R1} & \ref{Q2} & \ref{S1} ($\epsilon_p = \pmin = 0$) & \ref{SL2} \\ \hline 

F-SR-MAB
& \ref{O2} & \ref{R2} & \ref{Q2} & \ref{S1} ($\epsilon_p = \pmin = 0$) & \ref{SL2} \\ \hline 
F-AUC-AP
& \ref{O2} & \ref{R1} & \ref{Q1} & \ref{S2} & \ref{SL1} \\ \hline 
F-SR-AP
& \ref{O2} & \ref{R2} & \ref{Q1} & \ref{S2} & \ref{SL1} \\ \hline 
F-AUC-AP
& \ref{O2} & \ref{R1} & \ref{Q1} & \ref{S1} & \ref{SL1} \\ \hline 
F-SR-PM
& \ref{O2} & \ref{R2} & \ref{Q1} & \ref{S1} & \ref{SL1} \\ \hline 

RecPM
& \ref{O2} & \ref{SR2} & \ref{Q80} ($c1 = 1$, $c2 = 0.5$, $\gamma = 0.46$) & \ref{S1} ($\epsilon_p = 0, \pmin = 0.11$) & \ref{SL1} \\ \hline 

MAENSm
& \ref{O1} & \ref{SR1} ($\maxgen = \gamma = 1$,$Frac = \epsilon =0$) & \ref{Q2} & \ref{S1} & \ref{SL1} \\ \hline 

PM-AdapSS-AA
& \ref{O7} & \ref{W2} ($\omega = 0$) & \ref{Q1} & \ref{S1} ($\epsilon_p = 0$) & \ref{SL1} \\ \hline 
PM-AdapSS-N
& \ref{O7} & \ref{W2} ($\omega = 1$) & \ref{Q1} & \ref{S1} ($\epsilon_p = 0$) & \ref{SL1} \\ \hline 

Ex-PM
& \ref{O2} & \ref{B2} ($\rho = 1, \alpha = 0$) & \ref{Q1} & \ref{S1} & \ref{SL1} \\ \hline 
Ex-AP
& \ref{O2} & \ref{B2} ($\rho = 1, \alpha = 0$) & \ref{Q1} & \ref{S2} & \ref{SL1} \\ \hline
Ex-MAB
& \ref{O2} & \ref{B2} ($\rho = 1, \alpha = 0$) & \ref{Q2} & \ref{S1} ($\epsilon_p = \pmin = 0$) & \ref{SL2} \\


\bottomrule
\end{longtable}

\twocolumn
\cite{vafaee2008dynamic} proposes an AOS method, named as Dyn-GEP, in the context of Gene Expression programming (GEP). It assigns probabilities to the operators as follows:
\begin{equation}\label{eqn:dyn-gep}
    p_i(t) = \frac{d_i(t)}{\sum_{i=1}^n d_i(t)}\mbox{ and }
    d_i(t) = d_0 + m_i(t) + \alpha * m_i(t-1)
\end{equation}
where, $p_i(t)$, $d_i(t)$ and $m_i(t)$ is the probability, improvement and mean value of reward assigned to an operator $i$ at generation $t$ respectively; $\alpha$ represents the forgetting factor; $n$ is the number of operators and $d_0$ is the minimum value attained by $d_i(t)$. We simplify this formula to map to each component. It assigns fitness improvement w.r.t. parent (eq.~\ref{O2}) as offspring metric. This method runs the algorithm $20$ times and considers two rewards, best fitness value in $20$ runs and mean of best value found in each $20$ runs. As per the design of the framework, instead of running the algorithm multiple times, we run the method only once. Thus, we assign reward as mean of fitness values (eq.~\ref{W1}) produced by the application of the operator in the current generation (not run) and best operator application in the current generation (eq.~\ref{B2}). We consider Dyn-GEP with two versions named Dyn-GEPv1 and Dyn-GEPv2 for two reward definitions, respectively. Dyn-GEPv1 calculates mean fitness value from current generation ($\maxgen$ = 1) where the sum in the denominator of equation~\ref{W1} is the total number of application of an operator. Equation~\ref{B2} in Dyn-GEPv2 sets $\maxgen$ as $1$ with $\rho = 3$ without normalisation ($\alpha = 0$). These two definitions represent $m_i(t)$ in equation~\ref{eqn:dyn-gep}. The quality can be extracted from the above equation to map to equation~\ref{Q80} where hyper-parameters $\gamma$ and $c_1$ are set as $0$ and $1$ respectively and $c_2$ represents $\alpha$ in the above equation. Moving to probability, it is represented by equation~\ref{S1} with $\epsilon_p$ set as $d_0$ and $\pmin = 0$. Proportional selection is used to select operators based on the probability.

Hybrid algorithm, presented in~\cite{lobo1997decision}, is an algorithm that picks among two elitist algorithms. This approach is no different than AOS methods that are used to select operators. Thus, we decided to include it in the framework as a method to select among discrete choices. It uses raw fitness values (eq.~\ref{O1}) as offspring metric and their difference in two consecutive generations to assign reward to each choice. As soon as one of the algorithms is applied to evolve two individuals from the population, reward is assigned as a choice in reward component, Best2Gen (eq.~\ref{B1}) is utilised with the hyper-parameters $C = 1$, $\alpha$ and $\beta = 0$. It then combines current and previous reward (eq.~\ref{Q80}) of an algorithm as weighted geometric average with coefficient $c_1$ and $c_2$ summing to $1$. It does not consider future reward, so $\gamma$ is taken as $0$. Finally, probability (eq.~\ref{S1}) and selection (eq.~\ref{SL1}) definitions according to the probability matching is used as an assignment rule to probability and selection of an algorithm for next evolution.

ADOPP~\cite{Julstrom1995} is one of the early AOS methods proposed in 1995. It assigns each operator a probability proportional to the contribution it has made in producing an offspring better than population median. This operator receives a reward of 1.0 and assigns a decayed reward to its ancestors. However, we do not consider the possibility of rewarding the ancestors. To avoid the probability becoming zero, an extension of ADOPP is proposed, ADOPP-ext~\cite{julstrom1997adaptive}. In ADOPP-ext, a fraction of the sum of all operators' reward is added to each operator's accumulated reward.
Population-level Dynamic probabilities (PDP)~\cite{niehaus2001adaption} is a similar approach to ADOPP where instead of offspring comparison to median population fitness, it considers operation applications that improved fitness w.r.t parent. Other component choices in PDP are matched with ADOPP with the variation in their hyper-parameter values.

The SaDE algorithm~\cite{qin2009differential} combines adaptive control of the mutation and crossover rate with adaptive selection of two mutation operators. As our proposed framework is concerned with unifying AOS methods, we only show to replicate the AOS method involved in SaDE from the framework. Operators are assigned proportional to the probability of operator selection. The probability is based on the success and failure rate in previous fix number of generation. SaDE and MMRDE~\cite{qian2018differential} share the same component choices. The only difference is that latter assigns reward based on the applications in certain number of generations whereas the former only considers the applications in current generation. 

\RecPMaos~\cite{MudLopKaz2018ppsn} is a recently proposed AOS method that has shown promising results on \bbob test bed. It is based on the idea of maximising the future reward utilising probability matching mechanism. \Compass, \FAUCMAB and \PMAdapSS are among the popular AOS methods that can also be replicated by fixing choices for each component in the framework. 

\begin{figure*}[tbp]
\caption{Component choices with their usage frequency in the literature}\label{fig:choice-frequency}
\begin{minipage}{0.5\textwidth}
\centering
\begin{tabular}{@{}c@{}c@{}}
Offspring metrics & Quality choice \\
\includegraphics[scale=0.3]{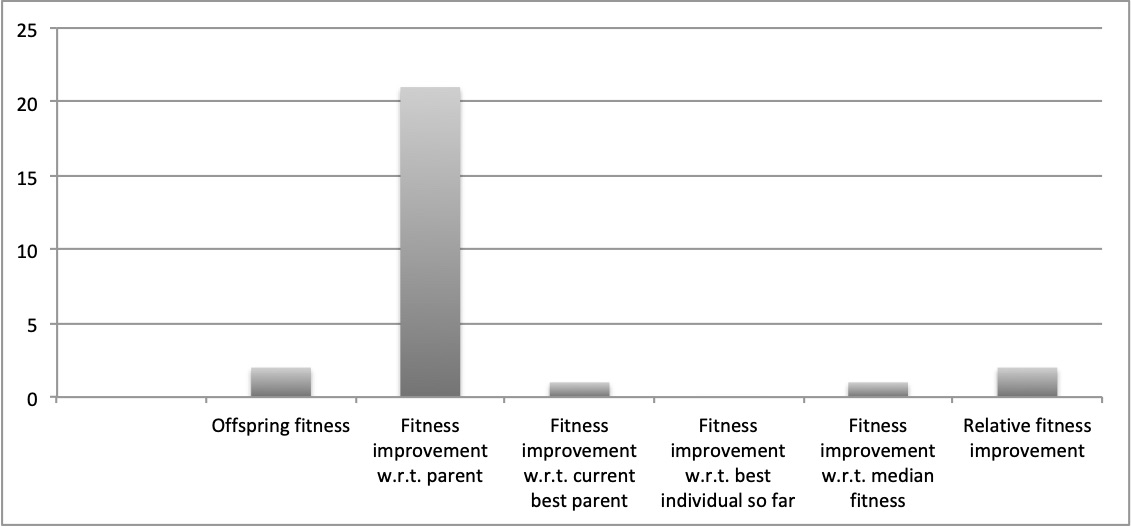} & \includegraphics[scale=0.3]{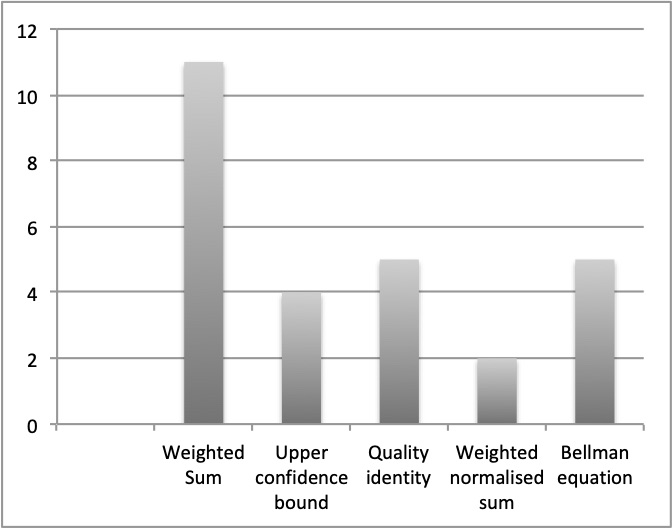} \\ 
\end{tabular}
\end{minipage}%
\\
\begin{minipage}{0.5\textwidth}
\centering
\begin{tabular}{@{}c@{}c@{}}
Selection choice & Probability choice \\
\includegraphics[scale=0.3]{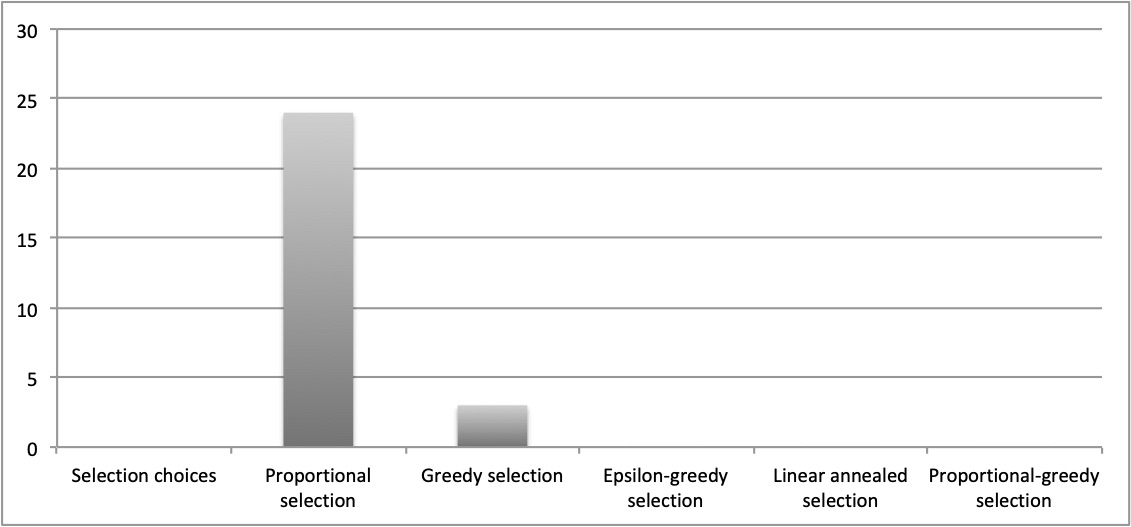} &
\includegraphics[scale=0.3]{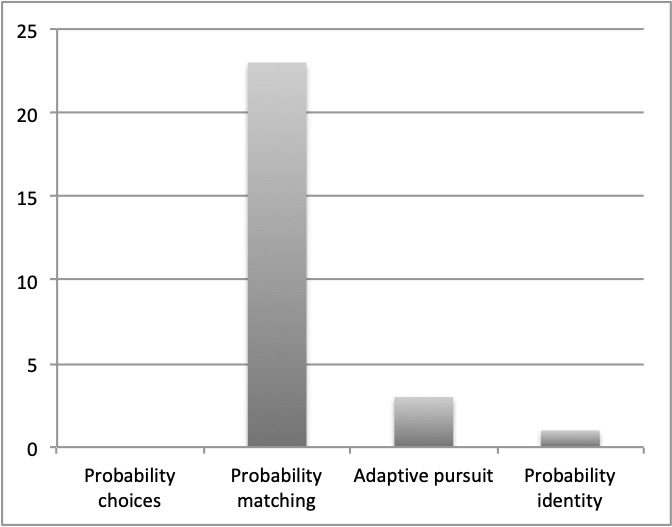} \\
\end{tabular}
\end{minipage}%
\\
\begin{minipage}{1.0\textwidth}
\centering
\begin{tabular}{@{}c@{}}
Reward metrics \\
\includegraphics[scale=0.3]{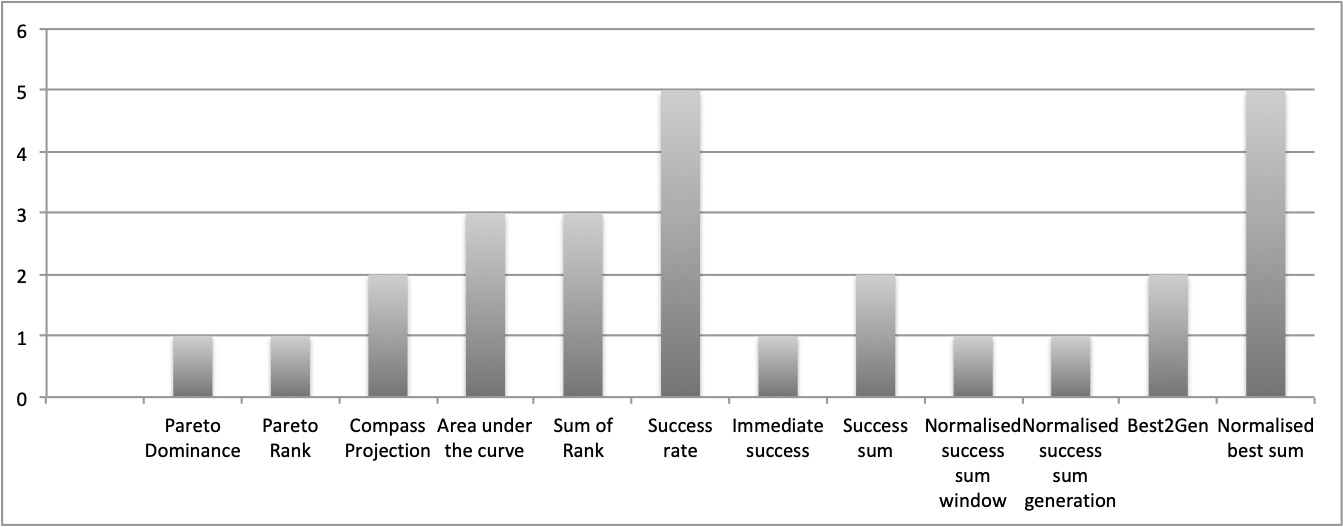} \\ 
\end{tabular}
\end{minipage}%
\end{figure*}

The usage of different component choices involved in adaptive methods vary greatly in the literature. We attempt to analyse the trend of utilising different choices in each component extracted from table~\ref{tab:AOS from literature}. The summary of choice count considered in literature is shown in figure~\ref{fig:choice-frequency}. 
It can be clearly seen that a lot of emphasis is given on exploring novel reward choices among all components while proposing an AOS method. Success rate (eq.~\ref{SR1}) and normalised best sum (eq.~\ref{B2}) are among the popular choices for reward. In other components, Fitness improvement w.r.t. parent (eq.~\ref{O2}) is frequently used offspring metric and probability matching is a common choice by researchers that assign quality (eq.~\ref{Q1}), probability (eq.~\ref{S1}) and selection mechanism (eq.~\ref{SL1}).
The combinations of these popular component choices have been explored in literature. \cite{maturana2010autonomous} considers three reward definitions based on fitness improvement w.r.t. parent namely, pareto rank, pareto dominance and projection. To form a complete AOS method, these are combined with PM. Six combinations are tested in~\cite{FiaSchoSeb2010toward}, area under the curve and sum of rank reward choices paired with dynamic multi-armed bandit (eq.~\ref{Q2} + eq.~\ref{S1} + eq.~\ref{SL2}), adaptive pursuit (eq.~\ref{Q1} + eq.~\ref{S2} + eq.~\ref{SL1}) and probability matching. We have not included statistical factor in the framework, and have only considered MAB instead of D-MAB. The papers testing other combinations include~\cite{fialho2008extreme, GonFiaCai2010adaptive, maturana2011adaptive}.

Besides these combinations, many other novel AOS methods can be designed using the framework. These combinations can be easily tested with differential evolution using the tool available on Github~\footnote{https://github.com/mudita11/Tune-AOS-bbob}. The framework can be utilised to control the discrete parameters of not only DE but also any other evolutionary algorithm that has discrete parameters.



\section{Experimental Design}
\subsection{Problem set}
We use the \bbob (Black-box optimisation benchmarking)~\cite{HanAugMer2016coco} problem suite to train and test the proposed algorithms. \bbob provides an easy to use tool-chain for benchmarking black-box optimisation algorithms for continuous domains and to compare the performance of numerical black-box optimisation algorithms. It consists of $24$ noiseless continuous benchmark functions~\cite{HanFinRosAug2009bbob} shown in table~\ref{tab:class and function}.
Each function consists of $15$ different instances, totalling to $360$ function instances. An instance of a function is a rotation and/or translation of the original function leading to a different global optimum. These $24$ functions are grouped in five classes, namely, separable functions ($f01$ -- $f05$), function with low or moderate conditioning ($f06$ -- $f09$), functions with high conditioning and uni modal ($f10$ -- $f14$), multi modal functions with adequate global structure ($f15$ -- $f19$) and multi modal functions with weak global structure ($f20$ -- $f24$). 

\begin{table*}[!tp]
\renewcommand{\arraystretch}{2.5}
\caption{\bbob function class and their functions}\label{tab:class and function} 
\centering
\begin{tabular}{p{4cm}|p{8cm}}
\toprule
\bfseries{Class name} & \bfseries{Function name} \\ 
\midrule
\bfseries{Separable functions} & Sphere function ($f01$), Ellipsoidal Separable function ($f02$), Rastrigin Separable function ($f03$), B\"{u}che-Rastrigin function ($f04$), Linear Slope ($f05$) \\
\bfseries{Function with low or moderate conditioning} & Attractive Sector function ($f06$), Step Ellipsoidal function ($f07$), Rosenbrock Original function ($f08$), Rosenbrock Rotated function ($f09$) \\
\bfseries{Functions with high conditioning and uni modal} & Ellipsoidal non-Separable function ($f10$), Discus function ($f11$), Bent Cigar function ($f12$), Sharp Ridge function ($f13$), Different Powers function ($f14$)\\
\bfseries{Multi modal functions with adequate global structure} & Rastrigin non-Separable function ($f15$), Weierstrass function ($f16$), Schaffers $f7$ function ($f17$), Schaffers $f7$ Moderately Ill-conditioned function ($f18$), Composite Griewank-Rosenbrock function $f8f2$ ($f19$) \\
\bfseries{Multi modal functions with weak global structure} & Schwefel function ($f20$), Gallagher’s Gaussian $101$-me Peaks function ($f21$), Gallagher’s Gaussian 21-hi Peaks function ($f22$), Katsuura function ($f23$), Lunacek bi-Rastrigin function ($f24$) \\
\bottomrule
\end{tabular}
\end{table*}

\subsection{Parameter tuning}
The framework consists of $6\times 12\times 5\times 3 \times 5 = 5,400$ AOS methods. These methods have their own hyper-parameters which increase the number of unique combinations possible. As the parameter space is huge, we decided to combine the framework with an offline method that tunes the framework and returns a combination of choices for a problem set. We use the offline automatic configurator \irace to tune the framework. \irace is a racing algorithm that saves the hassle of manual tuning and allows for a fully specified and reproducible procedure. The input given to \irace is the set of component choices along with the range of all hyper-parameters that needs tuning and a set of training function instances. In addition to tuning the framework and its hyper-parameters, it also tunes the parameters of the DE algorithm $F$, $NP$, $CR$ and $top_{NP}$. The parameter ranges/choices given to \irace are shown in table~\ref{component choices with hyper-parameters}. The name, type (Real, Integer, categorical) and range of each parameter tuned by \irace. The real-valued parameters take a two-point float value, integer type represents integer value in the given range and categorical can only select from the choices given. \irace starts the tuning procedure by sampling a number of candidate parameter configurations from either a randomly initialised probability distribution or from a population of starting configurations given to it. The candidate configuration sampled by \irace is an AOS method with a set of its hyper-parameter values and the DE parameters. The following steps are repeated until a budget given to \irace is exhausted. The generated candidate configurations are evaluated on a sequence of problem instances. This process of evaluation on a sequence of instances is known as racing. During the racing process, poor performing candidate configurations are discarded and elite ones survive. A race terminates once the allocated computation budget is exhausted or the number of surviving candidate configurations is below some specific number. The best candidate configurations are then used to update the probability distribution. The distribution is biased towards good candidate configurations and is used to generate new candidate configurations for the next iteration. 

\begin{table*}[tp]
\caption{Hyper-parameter choices given to \irace}\label{component choices with hyper-parameters} 
\centering
\begin{tabular}{c | c | c | c}
\hline
\bfseries Parameter Name & \bfseries Type & \bfseries Range & \bfseries Notes\\ \hline
\multicolumn{4}{c}{\textbf{DE parameters}} \\ \hline
F & Real & [0.1, 2.0] & Mutation Rate\\ 
CR & Real & [0.1, 1.0] & Crossover Rate\\ 
NP & Integer & [50, 400] & Population Size\\ 
$top_{NP}$ & Real & [0.02, 1.0] & Top $p$ candidates\\ \hline
\multicolumn{4}{c}{\textbf{Component choices}} \\ \hline
Offspring Metric & Categorical & [0, 6] & Type of $\Metric$\\ 
Reward Type & Categorical & [0, 11] & Type of Reward\\ 
Quality Type & Categorical & [0, 4] & Type of Quality\\ 
Probability Type & Categorical & [0, 2] & Type of Probability\\ 
Selection Type & Categorical & [0, 4] &Type of Selection\\ \hline
\multicolumn{4}{c}{\textbf{Reward Choice parameters}} \\ \hline
$\fixappl$ & Integer & [10, 50] & Fix number of applications\\ 
$\maxgen$ & Integer & [1, 50] & Maximum number of generations\\ 
$\theta$ & Categorical & (36, 45, 54, 90) & Projection angle\\ 
$\Window$ & Integer & (20, 150)	& Size of window\\ 
$D$ & Real & [0.0, 1.0] & Decay factor\\ 
$\gamma$ & Categorical & (1, 2) & Success choice\\ 
$Frac$ & Real & [0.0, 1.0] & Fraction of overall success\\ 
$\epsilon$ & Real & [0.0, 1.0] & Noise\\ 
$\omega$ & Categorical & (0, 1) & Normalisation choice\\ 
$C$ & Real & [0.001, 1.0] & Scaling constant\\ 
$\alpha$ & Categorical & (0, 1) & Decision parameter\\ 
$\beta$ & Categorical & (0, 1) & Decision parameter\\ 
$\rho$ & Categorical & (1, 2, 3) & Intensity\\ \hline 
\multicolumn{4}{c}{\textbf{Quality Choice parameters}} \\ \hline
$\delta$ & Real & (0.0, 1.0) & Decay rate\\ 
$C$ & Real & (0.0, 1.0) & Scaling Factor\\ 
$q_{min}$ & Real & (0.01, 1.0) & Minimum quality attained\\ 
$c_1$ & Real & (0.0, 1.0) & Memory for current reward\\ 
$c_2$ & Real & (0.0, 1.0) & Memory for previous reward\\ 
$\gamma$ & Real & (0.01, 1.0) & Discount rate\\ \hline
\multicolumn{4}{c}{\textbf{Probability Choice parameters}} \\ \hline
$\pmin$ & Real & [0.0, 1.0] & Minimum selection probability\\ 
$\epsilon_p$ & Real & (0.0, 1.0) & Noise\\ 
$\mu$ & Real & (0.0, 1.0) & Learning rate\\ 
$\pmax$ & Real & (0.0, 1.0)	& Maximum selection probability\\ \hline
\multicolumn{4}{c}{\textbf{Selection Choice parameters}} \\ \hline
$eps$ & Real & [0.0, 1.0] & Random probability of selection\\ \hline
\end{tabular}
\end{table*}

The budget given to \irace is $10^4$ and the budget assigned to a DE run is $10^4 \cdot n$ function evaluations where $n=20$ is the dimension of sampled function. All algorithms in this work focus on $n=20$ for all functions. We want to generalise on different classes of functions in the \bbob noiseless functions of dimension $20$, thus we select function instances across different function classes. To prevent over-fitting, the training set contains two randomly selected function instances out of $15$ from each of the $24$ functions. Thus, training set consists of total $48$ out of $360$ function instances. These $48$ function instances are shown in table~\ref{tab:training set}.

\begin{table*}[!tp]
\renewcommand{\arraystretch}{2.5}
\caption{Training set. $fx iy$ denotes a function instance $iy$ that is obtained by a transformation of original function $fx$.}\label{tab:training set} 
\centering
\begin{tabular}{p{4cm}|p{8cm}}
\hline
\bfseries{Function class} & \bfseries{Function instance} \\ \hline
\bfseries{Separable functions} & $f01 i01$, $f01 i07$, $f02 i09$, $f02 i15$, $f03 i10$, $f03 i05$, $f04 i08$, $f04 i06$, $f05 i07$, $f05 i01$ \\
\bfseries{Function with low or moderate conditioning} & $f06 i13$, $f06 i07$, $f07 i02$, $f07 i05$, $f08 i06$, $f08 i03$, $f09 i10$, $f09 i03$ \\
\bfseries{Functions with high conditioning and uni modal} & $f10 i11$, $f10 i04$, $f11 i09$, $f11 i02$, $f12 i01$, $f12 i03$, $f13 i13$, $f13 i12$, $f14 i12$, $f14 i11$ \\
\bfseries{Multi modal functions with adequate global structure} & $f15 i07$, $f15 i15$, $f16 i02$, $f16 i14$, $f17 i12$, $f17 i15$, $f18 i09$, $f18 i15$, $f19 i01$, $f19 i09$ \\
\bfseries{Multi modal functions with weak global structure} & $f20 i10$, $f20 i06$, $f21 i05$, $f21 i11$, $f22 i01$, $f22 i08$, $f23 i03$, $f23 i15$, $f24 i08$, $f24 i04$ \\
\bottomrule
\end{tabular}
\end{table*}

As the parameter space given to \irace is huge, we give four tuned AOS methods as the starting configurations to \irace. \Compass~\cite{maturana2008compass}, \PMAdapSS~\cite{GonFiaCai2010adaptive}, \FAUCMAB~\cite{FiaSchoSeb2010toward} and \RecPMaos~\cite{MudLopKaz2018ppsn} are among the popular AOS methods that act as initial population for \irace to explore the parameter search space. Their hyper-parameters along with DE parameters are tuned using \irace. The tuned AOS methods within DE are shown in table~\ref{tab:initial and returned configuration}. The first column shows the parameter name involved in the starting configurations followed by four AOS methods. 

\begin{table*}[tp]
\caption{Starting configurations and the configuration returned by \irace. The symbol `-' in the table means that the parameter is not applicable to the AOS method.}\label{tab:initial and returned configuration} 
\centering
\begin{tabular}{c | c | c | c | c | p{2.5cm}}
\hline
\textbf{Parameter name} & \textbf{\RecPMaos} & \textbf{\emph{PM-AdapSS-NN}} & \textbf{\emph{F-AUC-MAB}} & \textbf{\emph{Compass}} & \textbf{Configuration returned by \irace (\UAOSFW)}\\ \hline

\multicolumn{6}{c}{\textbf{DE parameters}} \\ \hline
F & 0.57 & 0.47 & 0.45 & 0.51 & 0.41 \\ 
CR & 0.93 & 0.96 & 0.21 & 0.95 & 0.91 \\ 
NP & 154 & 329 & 57 & 163 & 262 \\ 
p in pbest & 0.05 & 0.07 & 0.73 & 0.64 & 0.02 \\ \hline

\multicolumn{6}{c}{\textbf{Component choices}} \\ \hline
Offspring Metric & \ref{O2} & \ref{O2} & \ref{O2} & \ref{O2} & \ref{O2} \\ 
Reward Type & \ref{SR2} & \ref{W2} & \ref{R1} & \ref{DQ3} & \ref{SR2} \\ 
Quality Type & \ref{Q80} & \ref{Q1} & \ref{Q2} & \ref{Q4} & \ref{Q80} \\ 
Probability Type & \ref{S1} & \ref{S1} & \ref{S1} & \ref{S1} & \ref{S1} \\ 
Selection Type & \ref{SL1} & \ref{SL1} & \ref{SL2} & \ref{SL1} & \ref{SL1} \\ \hline

\multicolumn{6}{c}{\textbf{Reward Choice hyper-parameters}} \\ \hline
$\fixappl$ & - & - & - & 66 & - \\ 
$\theta$ & - & - & - & 90 & - \\ 
$\Window$ & - & 73 & 138 & - & - \\ 
$D$ & - & - & 0.47 & - & - \\ 
$\omega$ & - & 1 & - & - & - \\ \hline

\multicolumn{6}{c}{\textbf{Quality Choice hyper-parameters}} \\ \hline
$\delta$ & - & 0.07 & - & - & - \\ 
$c$ & - & - & 0.04 & - & 0.54 \\ 
$c_1$ & 0.57 & - & - & - & 0.66 \\ 
$c_2$ & 0.96 & - & - & - & 0.45 \\ 
$\gamma$ & 0.43 & - & - & - & - \\ 

\multicolumn{6}{c}{\textbf{Probability Choice hyper-parameters}} \\ \hline
$\pmin$ & 0.08 & 0.06 & 0.02 & 0.08 & 0.04 \\ 
$\epsilon_p$ & 0.26 & 0.53 & 0.72 & 0.55 & 0.22 \\ \hline

\end{tabular}
\end{table*}

\section{Testing phase}
After tuning, \irace returns an AOS method along with its tuned hyper-parameter values and tuned DE parameter values given the starting configurations. The last column in table~\ref{tab:initial and returned configuration} shows the configuration returned by \irace, abbreviated as \UAOSFW. The returned configuration is a variant of \RecPMaos~\cite{MudLopKaz2018ppsn} that acts as a starting configuration given to \irace. \RecPMaos assigns reward to an operator depending on the short term success of that operator and estimates quality based on the expected quality of possible selection of operators in the past. It shares similarities with \PMAdapSS AOS method, one of the starting configuration. \PMAdapSS utilises \PM to select an operator whereas \RecPMaos uses a variant of \PM known as \RecPM. Both AOS methods use reward based on the number of improvements from parent to offspring, however, \PMAdapSS uses average relative fitness improvement as immediate reward without using accumulated reward, whereas \RecPMaos uses offspring survival rate as immediate reward combined with a fraction of its previous accumulated reward.

Algorithm~\ref{alg:test-de-aos} shows the working steps of AOS within DE in the testing phase. For a given test problem, this is simply the working of DE with multiple mutation operators where each parent is evolved using an operator selected with the selection method employed in the AOS method. Testing starts by initialising and evaluating the parent population. The $\Metric$ values are calculated for each offspring to initialise the generation and window memory. The probability for each operator is initialised as $\frac{1}{K}$ 
where $K$ is the total number of mutation strategies. This gives every operator an equal chance to get selected in the beginning. Once the initialisation phase is over, the following steps are repeated until the stopping criteria are not satisfied. The parent population is evolved using a mutation strategy selected for each parent by selection definition in the AOS method. This improvement per parent is stored in $\Metric$. The memory is updated based on $\Metric$. The offspring population is evaluated and the solution among parent and offspring with better fitness survives. This is followed by reward, quality and probability update according to the AOS method. The selection of operator is performed based on the probability of each operator. Once the algorithm terminates, best fitness value is returned.

\begin{algorithm}[!t]
\caption{AOS method formed with the component choices from the framework coupled with DE}\label{alg:test-de-aos}
\begin{algorithmic}[1]
\STATE Given: Tuned $CR$, $F$ and $NP$; Component choices from AOS framework
\STATE Initialise and evaluate fitness of each individual $x_i$ in the population
\STATE $g=0$ (generation number)
\STATE Calculate $\Metric(0, k, op)$, $\forall$ k 
\STATE Initialise generation and window memory using $\Metric(0, k, op)$, $\forall$ k
\STATE Initialise probability $P_{0,op}$, $\forall\op\in\Operators$ 
\WHILE{stopping condition is not satisfied}
    \FOR{\textbf{each} $x_i$, $i=1,\dotsc, NP$}
        \IF{one or more operators not yet applied}
            \STATE $\op = \text{Uniform selection among operator(s) not yet applied}$
        \ELSE
            \STATE $\op = \text{Select a mutation strategy based on the selection choice}$
        \ENDIF
        \STATE Generate offspring using selected operator $\op$ 
    \ENDFOR
    \STATE Calculate $\Metric(g, k, op)$
    \STATE Update generation and window memory using $\Metric(g, k, op)$, $\forall$ k
    \STATE Evaluate offspring population
    \STATE Perform survival selection
    \STATE Calculate reward for each operator $R_{g+1,op}$
    \STATE Estimate quality for each operator $Q_{g+1,op}$
    \STATE Update probability for each operator $P_{g+1,op}$
    \STATE $g=g+1$
\ENDWHILE
\end{algorithmic}
\end{algorithm}

We select four tuned AOS methods within DE \PMAdapSS~\cite{GonFiaCai2010adaptive}, \FAUCMAB~\cite{FiaSchoSeb2010toward}, \Compass~\cite{maturana2008compass} and \RecPMaos~\cite{MudLopKaz2018ppsn}; two non-AOS DE algorithms with one mutation strategy, \JaDE~\cite{ZhaSan2009jade} and \RSHADE~\cite{tanabe2015tuning} 
to compare with the returned configuration. \PMAdapSS and \FAUCMAB were introduced in the context of selecting a mutation strategy in DE~\cite{PriStoLam2005:book}. In particular, \PMAdapSS uses probability matching as the method for operator selection, whereas \FAUCMAB employs a method inspired by multi-armed bandits. \Compass evaluates an operator's impact using two measures, mean fitness and population fitness diversity. It measures how well the operator balances the exploration and exploitation. 
The two non-AOS algorithms employ different ways to self-adapt crossover and mutation rate in DE. \JaDE is a DE variant that adapts the crossover probability $CR$ and mutation factor $F$ using the values which proved to be useful in the current generation. \RSHADE is an improvement upon \JaDE which employs a restart mechanism and uses a parameter adaptation mechanism based on a historical record of successful parameter settings to adapt $CR$ and $F$. \JaDE and \RSHADE utilise same mutation strategy ``current-to-pbest'' to evolve population and it is one of the operators adapted by \UAOSFW and other AOS methods in comparison.

The data for non-AOS methods is taken from the \textsc{coco} website.\footnote{\url{http://coco.gforge.inria.fr/doku.php?id=algorithms-bbob}} 
The returned configuration and other six algorithms from the literature are evaluated on the remaining $312$ \bbob noiseless benchmark set on dimension $20$. We use plots of the Empirical Cumulative Distribution Function (ECDF) to assess their performance (Fig.~\ref{fig:ECDF-UFW1}). The ECDF displays the proportion of problems solved within a specified budget of function evaluations (FEvals) for different targets $f_\text{target} = \fopt + \Delta f$, where $\fopt$ is an the optimum function value to reach with some precision $\Delta f \in [10^{-8}, 10^2]$. In the plots, FEvals is given on the x-axis and y-axis represents the fraction of problems solved. A large symbol `$\times$' shows the maximum number of function evaluations (budget) given to each algorithm. We give ${10}^{5}\cdot 20$ FEvals as the budget to each algorithm with AOS method. The results reported after this symbol use bootstrapping to estimate the number of evaluations to reach a specific target for a problem~\cite{efron1994introduction}. The results denoted with \texttt{best 2009} correspond to the artificial best algorithm from the \bbob-2009 workshop constructed from the data of the algorithm with the smallest $\mathrm{aRT}$ (average Run Time) for each set of problems with the same function, dimension and target. The $\mathrm{aRT}$ is calculated as the ratio of the number of function evaluations for reaching the target value over successful runs (or trials), plus the maximum number of evaluations for unsuccessful runs, divided by the number of successful trials. The trials that reached $f_\text{target}$ within the specified budget are termed as successful trials, $\#succ$. The $\mathrm{aRT}$ tables for eight algorithms are shown in tables~\ref{tab:art-ufw1} and~\ref{tab:art-ufw2}. 

\section{Experiments and Results}
\begin{figure*}[tp]
\caption{Parameter sampling frequency. FF for F, top\_NP for p in pbest, $\theta$ for theta, \Window for window, D for decay, $\gamma$ for succ\_lin\_quad, Frac for frac, $\epsilon$ for noise, $\omega$ for normal\_factor, C for scaling\_constant, $\alpha$ for alpha, $\beta$ for beta, $\rho$ for intensity, c for scaling\_factor, $\delta$ for decay\_rate, c1 for weight\_reward, c2 for weight\_old\_reward, $\gamma$ for discount\_rate, $\mu$ for learning\_rate, $\epsilon_p$ for error\_prob, eps for sel\_eps} \label{fig:parameter frequency}
\begin{tabular}{@{}c@{}}
\includegraphics[height = 7.7cm, width = 15cm]{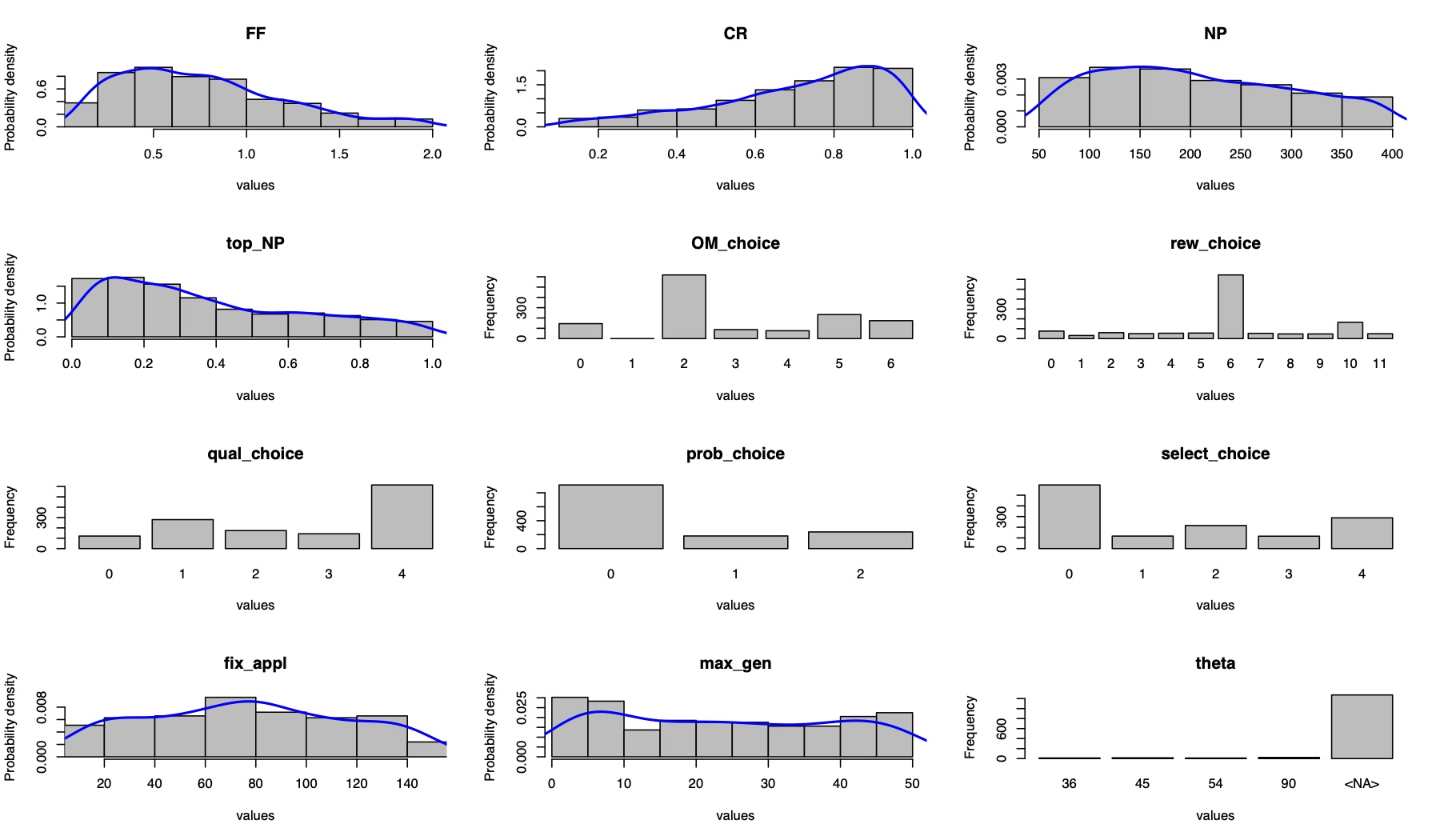} \\ \includegraphics[height = 7.7cm, width = 15cm]{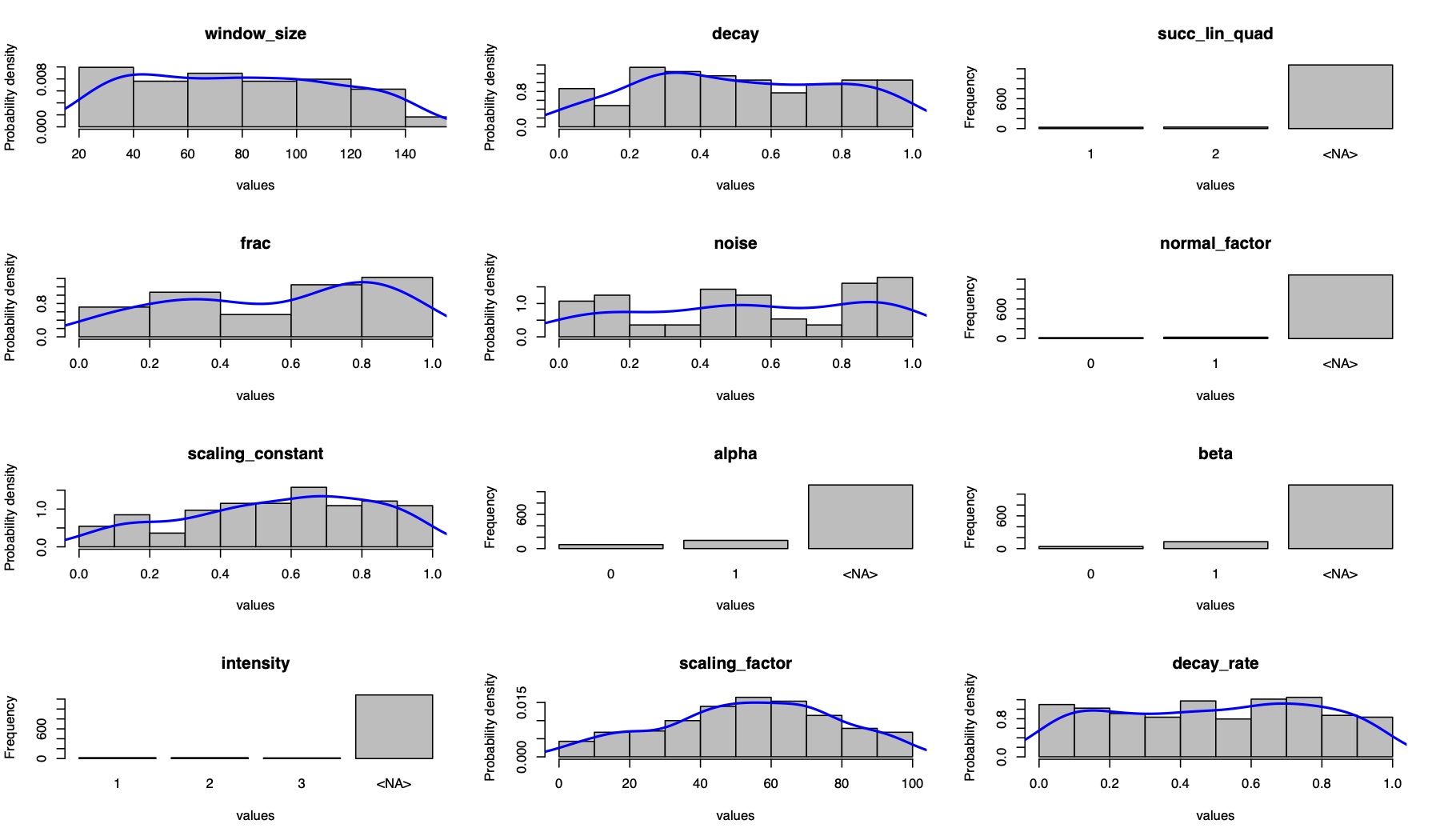} \\ \includegraphics[height = 7.7cm, width = 15cm]{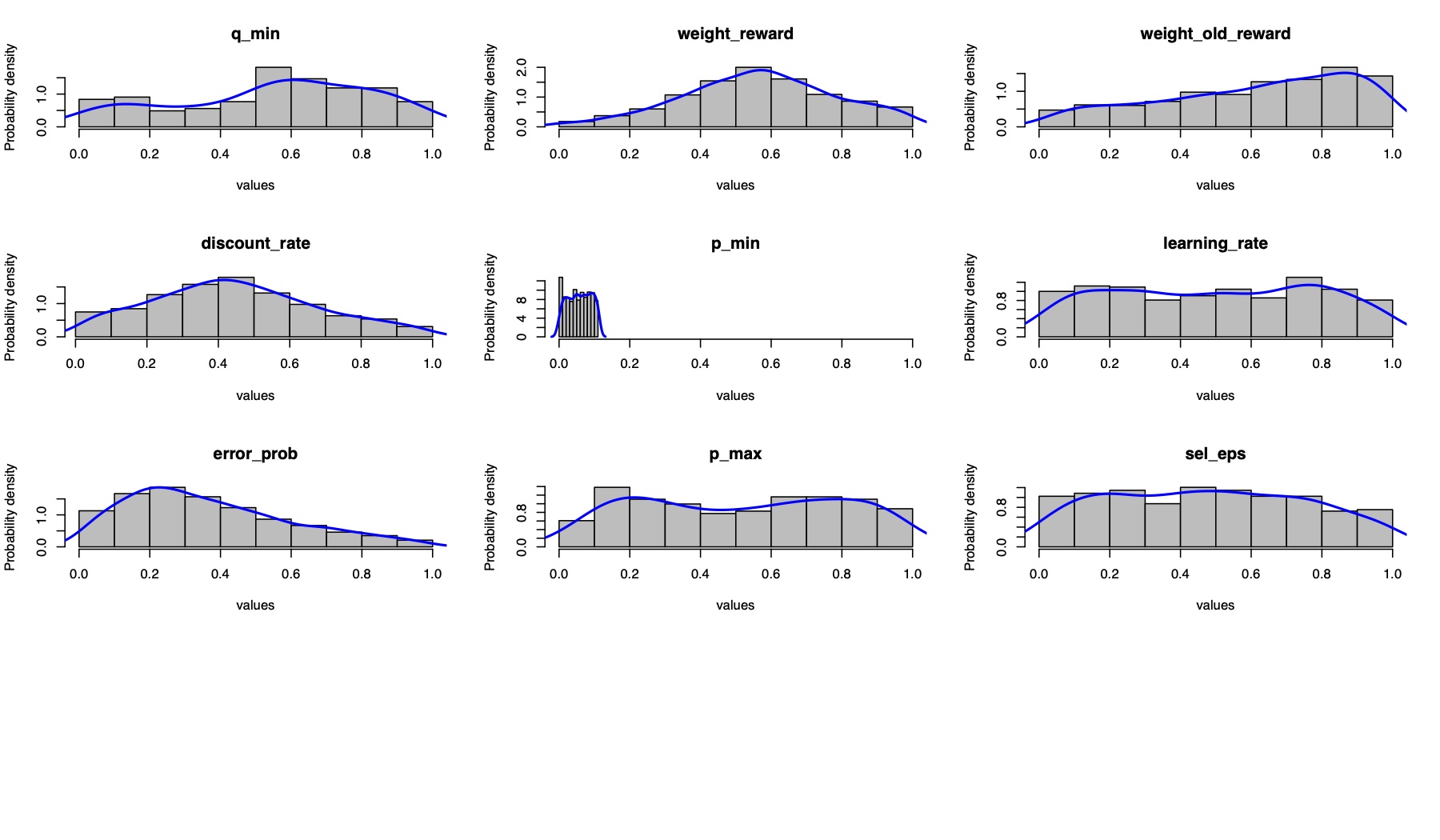}
\end{tabular}
\end{figure*}

During the tuning phase, \irace samples parameters from their individual parameter distribution. This distribution is updated to focus on the efficient parameter range. The frequency of sampled frequency of the parameters is shown in figure~\ref{fig:parameter frequency}. Categorical parameters that take values from a set of values are shown in the figure such as OM\_choice and frac. Parameters that take floating-values such as $FF$ and decay sample from a probability distribution such as a normal distribution. 

\subsection{Comparison of \UAOSFW with other tuned AOS methods}
ECDF graphs on 24 functions each with 13 test instances are shown in figure~\ref{fig:ECDF-UFW1}. It also shows class-wise and overall performance on the test set. These graphs clearly show that the configuration returned by \irace, \UAOSFW reaches all targets with competitive speed for functions f001, f002 and f005-f014 in comparison with \RecPMaos, \PMAdapSS, \Compass and \FAUCMAB. 
Looking at the separable class \Compass and \UAOSFW solves problems with the same speed solving the same number of problems as \PMAdapSS. Although \UAOSFW is a variant of \RecPMaos, the former solves more problems than the latter with faster speed. \FAUCMAB has shown exceptional performance in terms of speed and problems solves compared to other AOS methods for separable class of problems.
\UAOSFW performed best for low/moderate conditioning problems reaching all targets for all 4$\times$13 test instances falling under this category and 
the same performance of all AOS methods can be seen in the case of high conditioning class of problems. \FAUCMAB performed worse on low/moderate and high conditioning class, solving the least number of problems.
\UAOSFW and \RecPMaos excelled in the multi model (adequate structure) class of problems. They showed similar performance with former solving faster than any other algorithm.
\UAOSFW came second in the multi model (weak structure) problems after \FAUCMAB which solved more than 30\% problems.
Overall, \UAOSFW solved approximately 65\% of total problems which is more than the number of problems solved by any tuned AOS method and also with faster speed compared to any AOS method.

\begin{figure*}[tbp]
\caption{ECDFs on test set. F-AUC for \FAUCMAB, UFW for \UAOSFW, RecPM for \RecPMaos, AdapSS for \PMAdapSS}\label{fig:ECDF-UFW1}
\begin{tabular}{@{}c@{}c@{}}
\bfseries f001 & \bfseries f002\\
\includegraphics[scale=0.25]{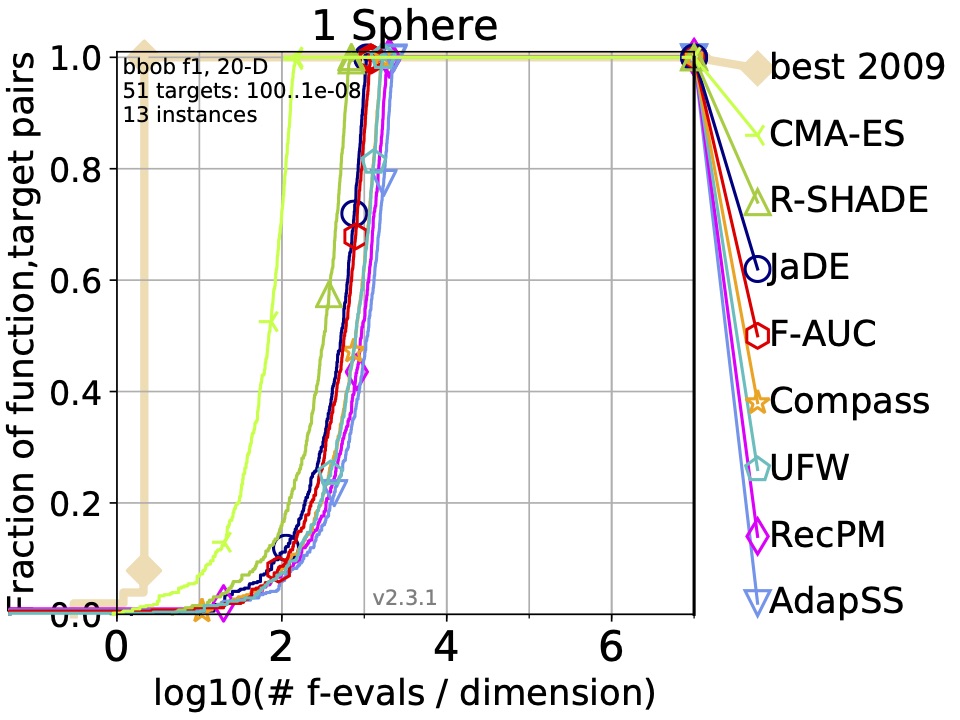} & \includegraphics[scale=0.25]{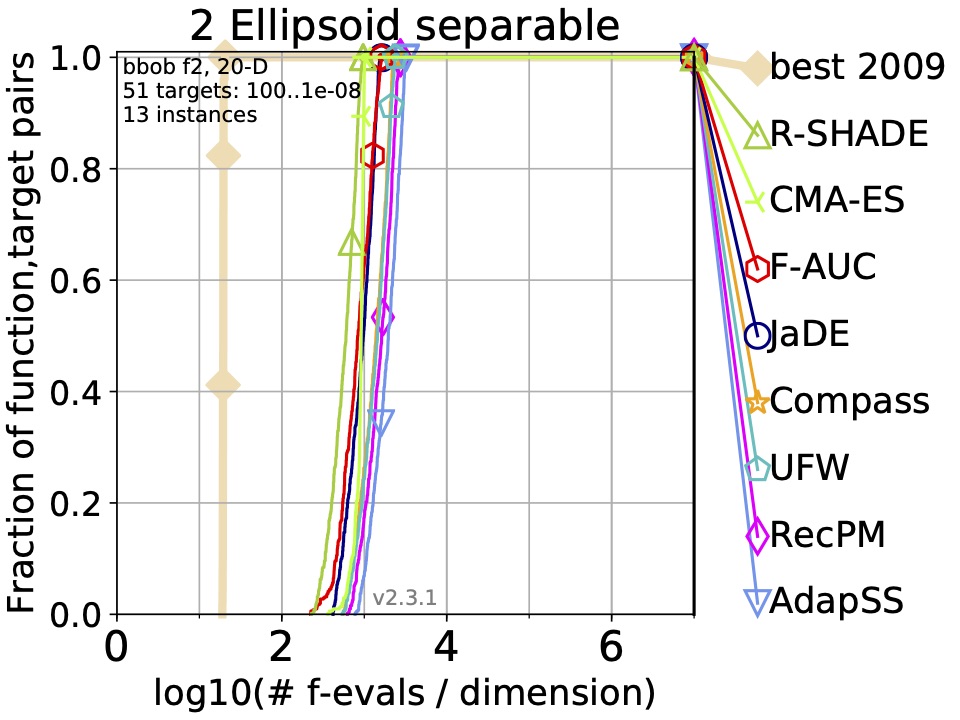} \\
\bfseries f003 & \bfseries f004 \\
\includegraphics[scale=0.25]{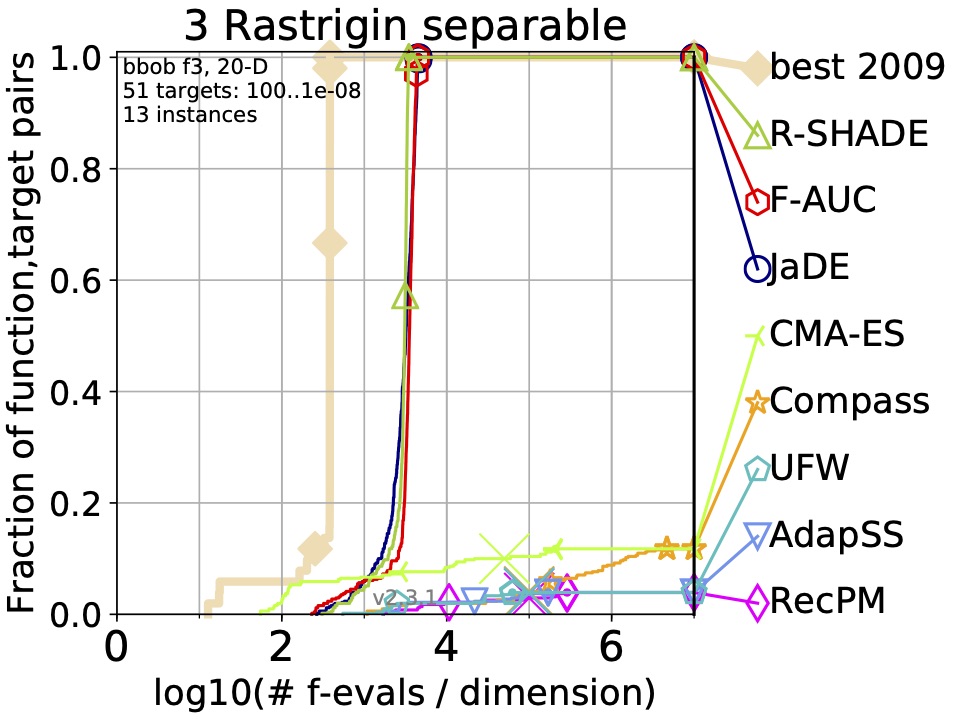} & \includegraphics[scale=0.25]{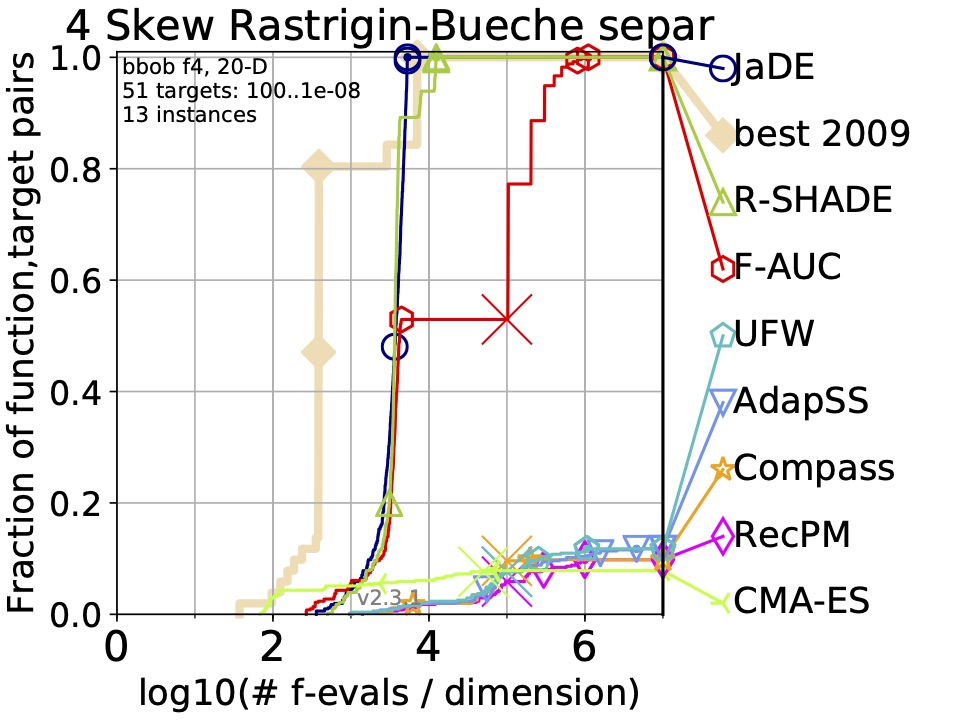}\\
\bfseries f005 & \bfseries f006\\
\includegraphics[scale=0.25]{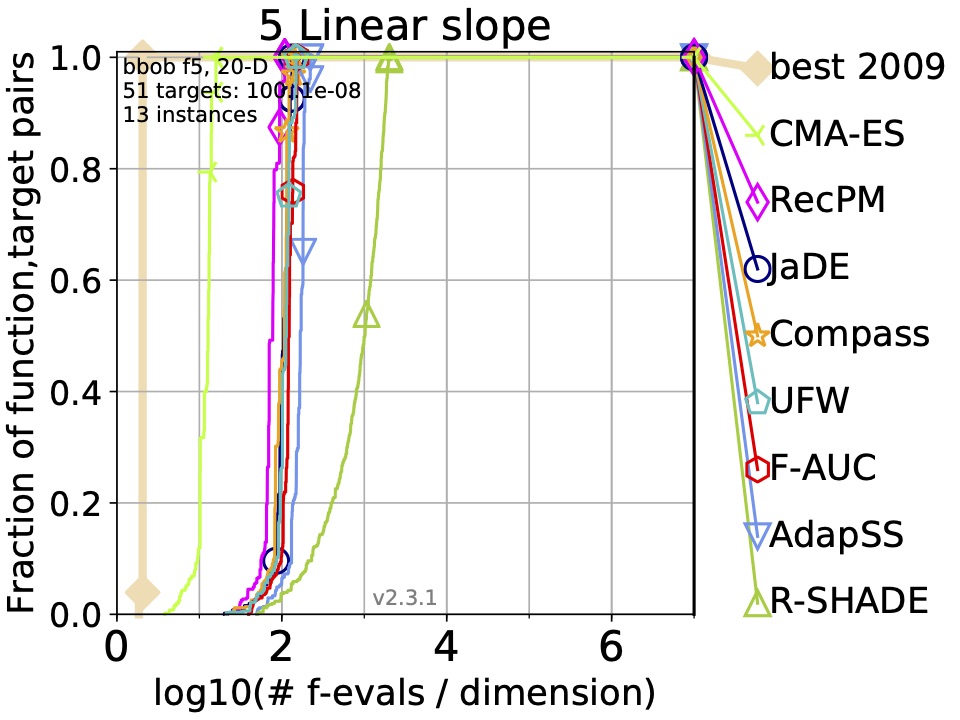} & \includegraphics[scale=0.25]{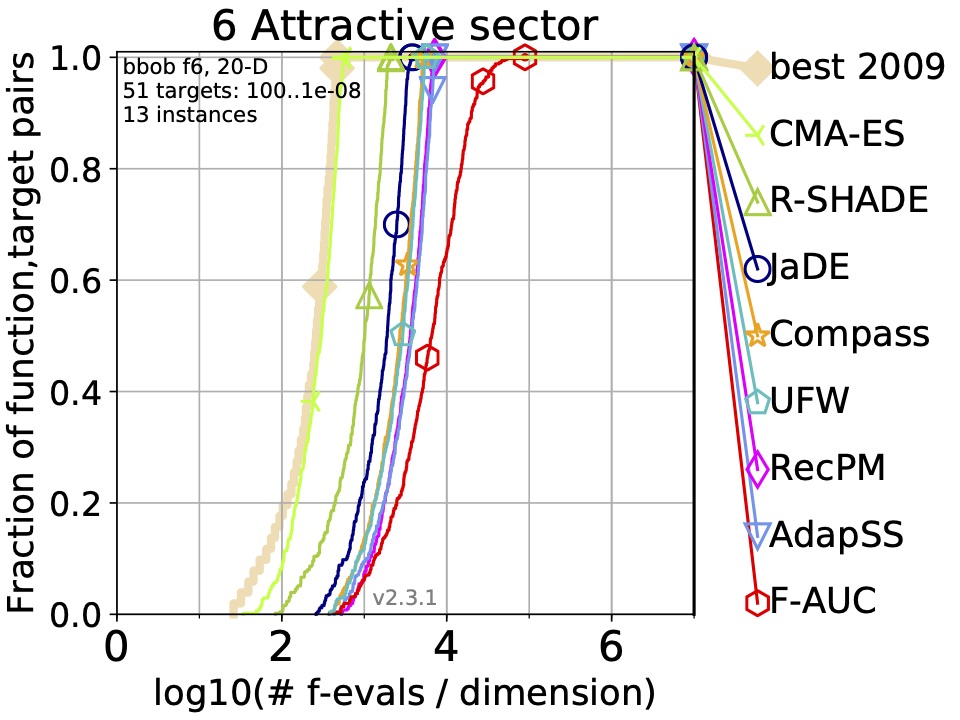} \\
\end{tabular}
\end{figure*}

\begin{figure*}[tbp]
\captionsetup{list=off}
\ContinuedFloat
\caption{ECDFs on test set. F-AUC for \FAUCMAB, UFW for \UAOSFW, RecPM for \RecPMaos, AdapSS for \PMAdapSS}\label{fig:ECDF-UFW2}
\begin{tabular}{@{}c@{}c@{}}
\bfseries f007 & \bfseries f008 \\
\includegraphics[scale=0.25]{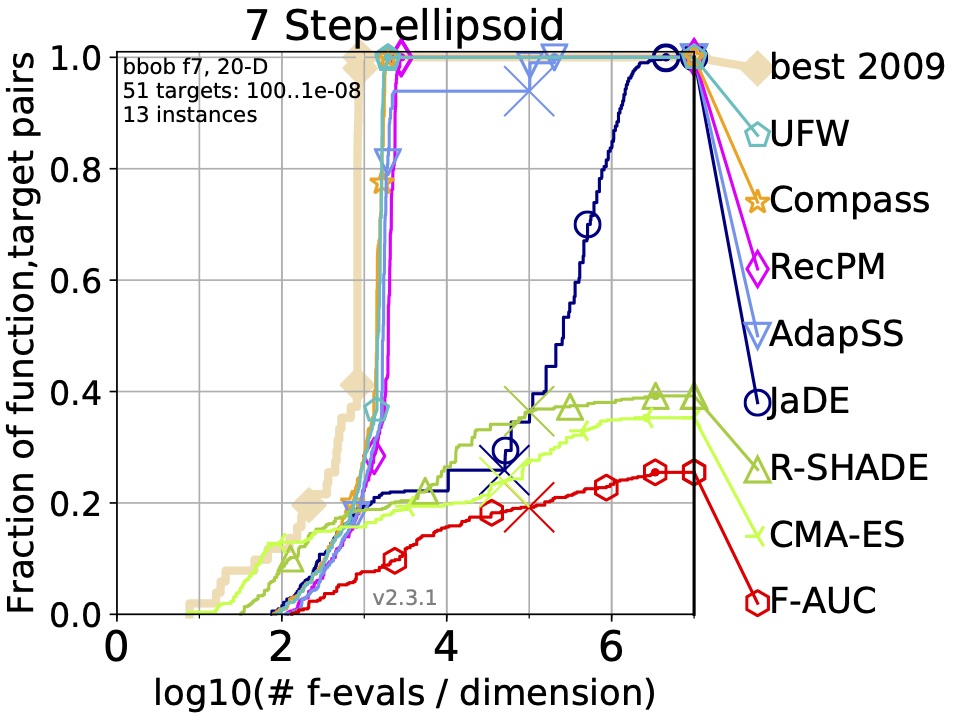} & \includegraphics[scale=0.25]{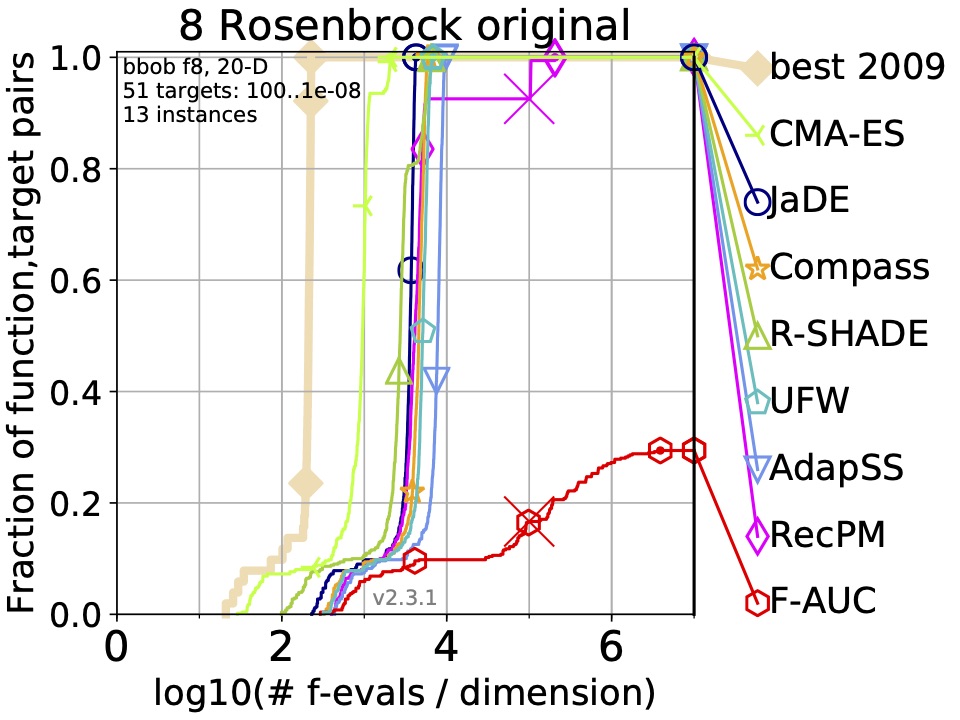}\\ 
\bfseries f009 & \bfseries f010\\
\includegraphics[scale=0.25]{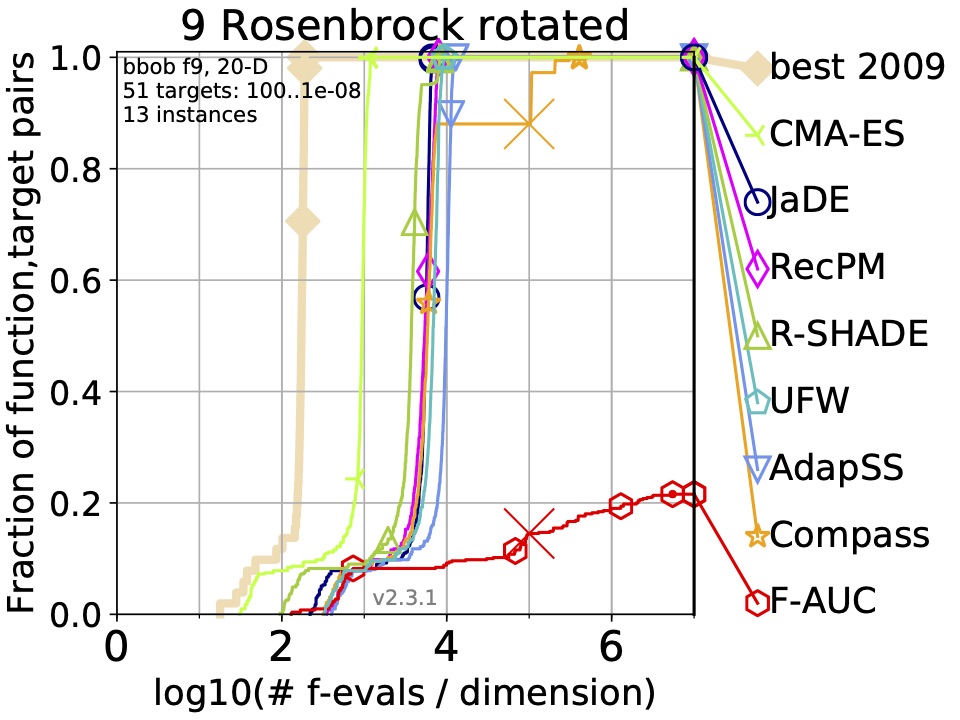} & \includegraphics[scale=0.25]{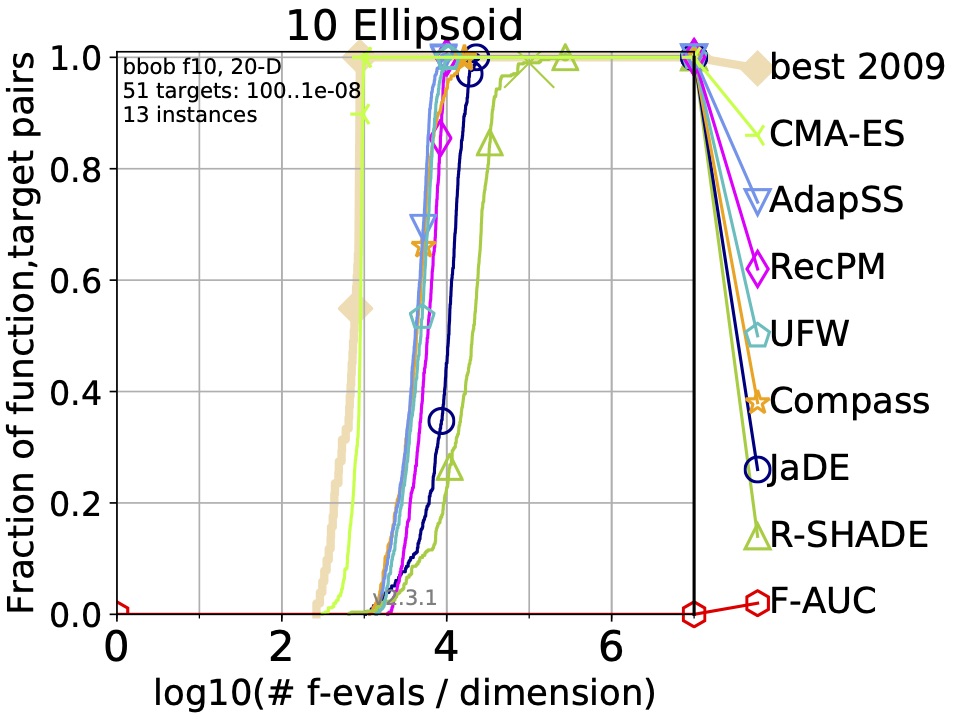} \\
\bfseries f011 & \bfseries f012\\
\includegraphics[scale=0.25]{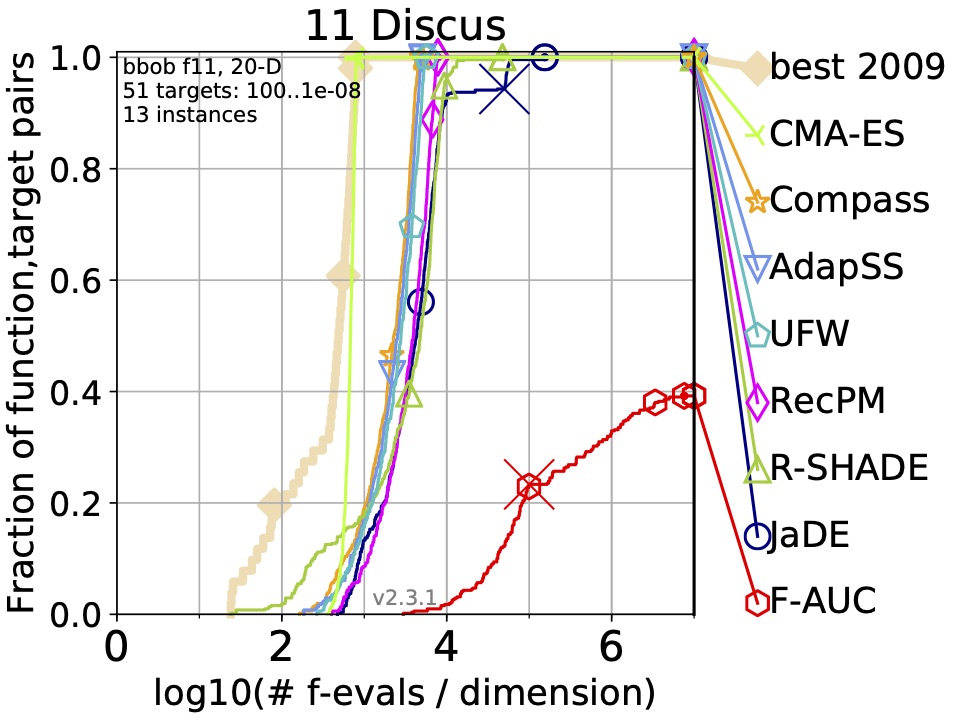} & \includegraphics[scale=0.25]{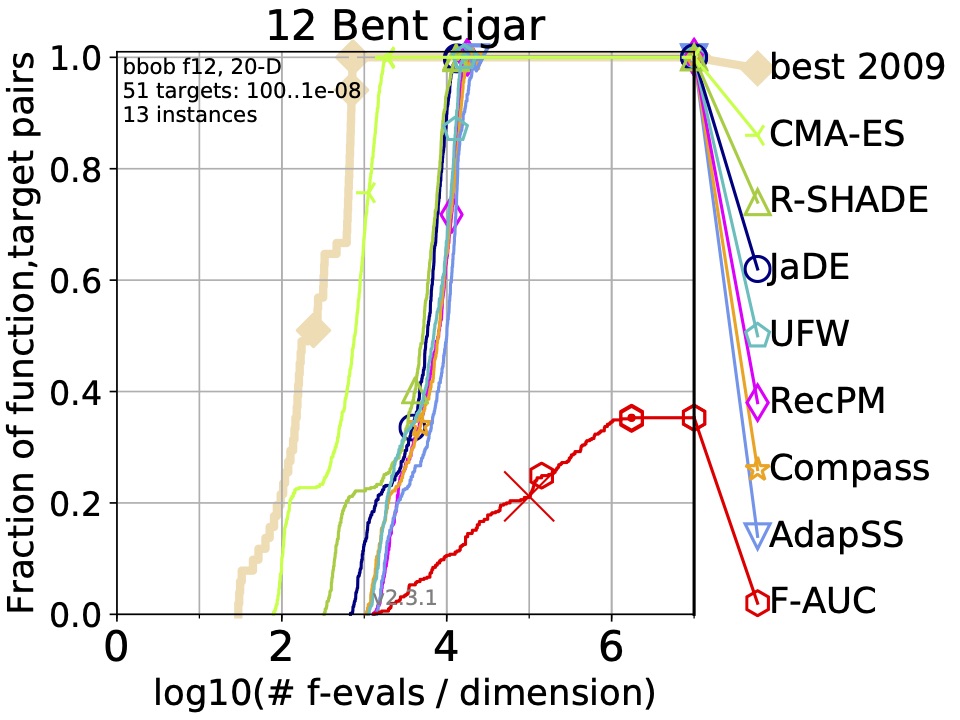}\\
\end{tabular}
\end{figure*}

\begin{figure*}[tbp]
\captionsetup{list=off}
\ContinuedFloat
\caption{ECDFs on test set. F-AUC for \FAUCMAB, UFW for \UAOSFW, RecPM for \RecPMaos, AdapSS for \PMAdapSS}\label{fig:ECDF-UFW2}
\begin{tabular}{@{}c@{}c@{}}
\bfseries f013 & \bfseries f014 \\
\includegraphics[scale=0.25]{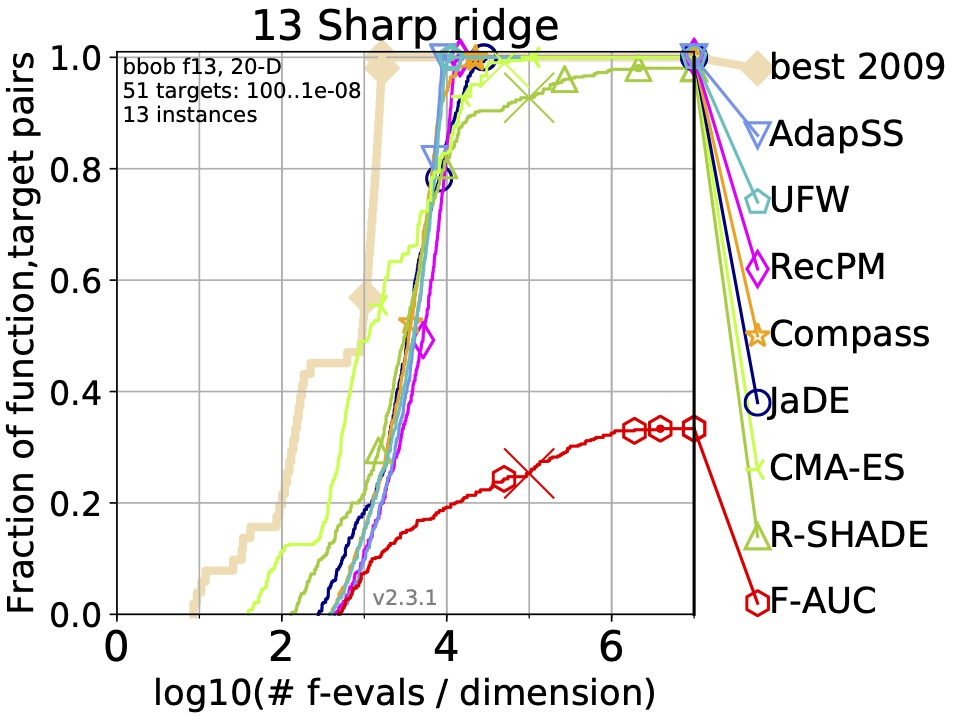} & \includegraphics[scale=0.25]{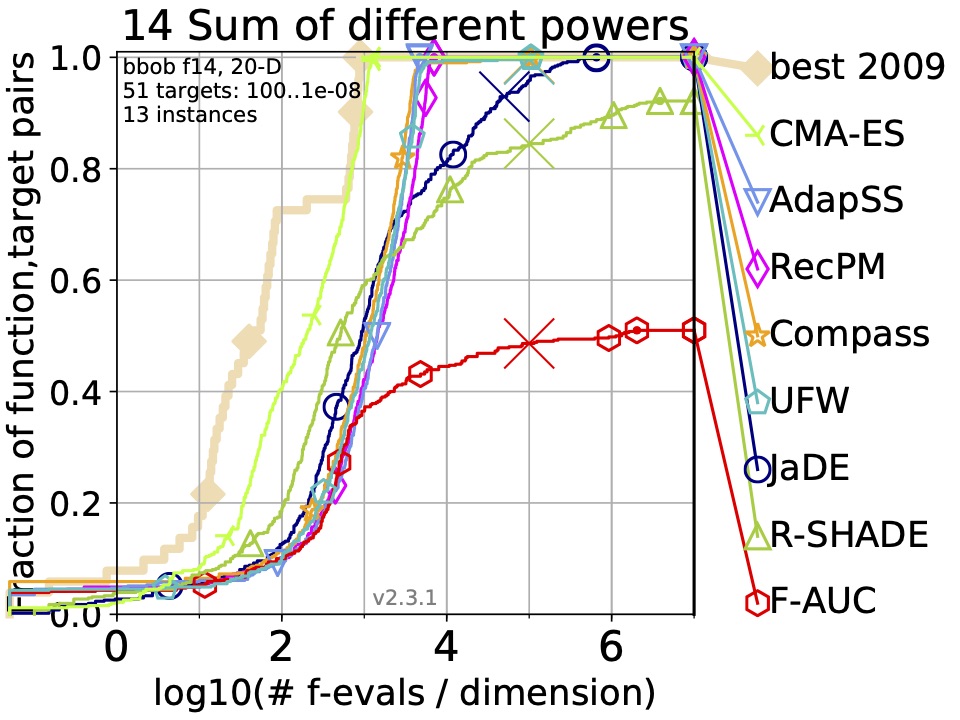} \\
\bfseries f015 & \bfseries f016\\
\includegraphics[scale=0.25]{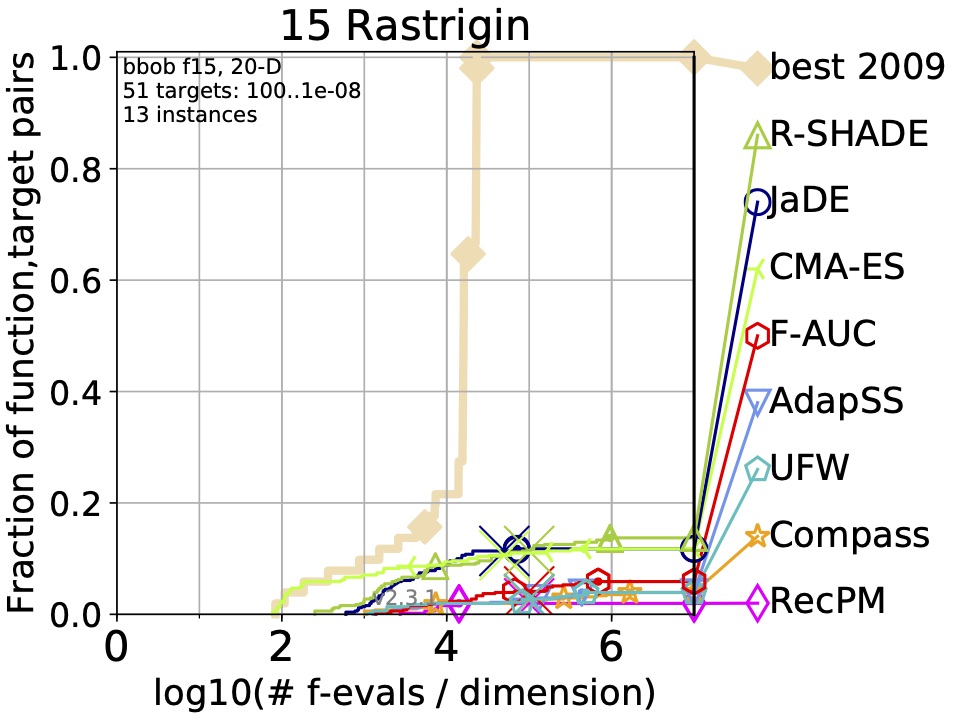} & \includegraphics[scale=0.25]{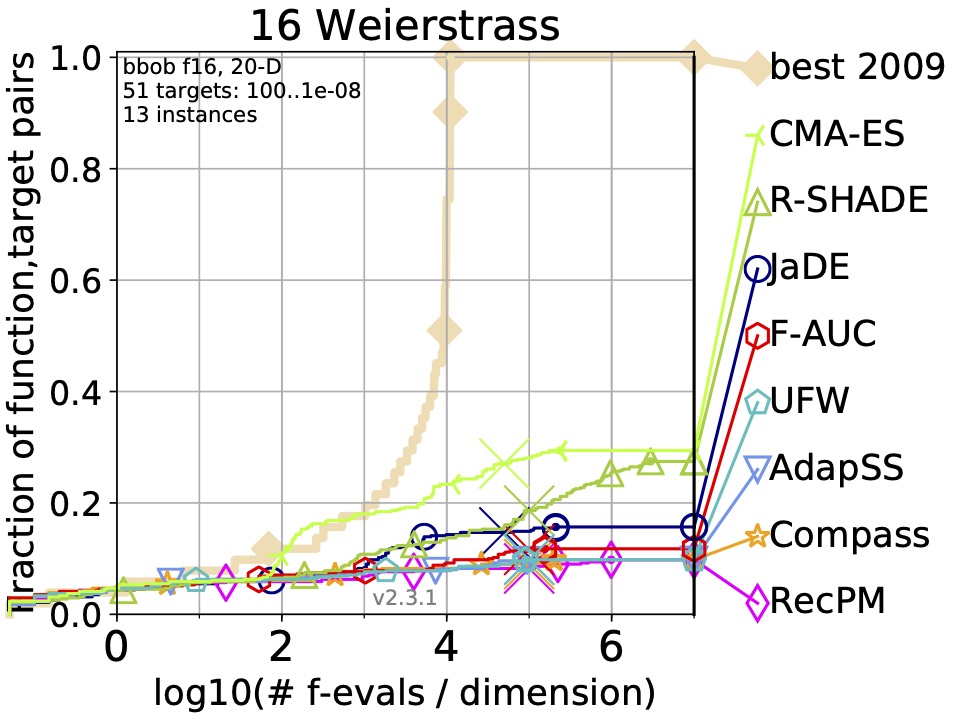}\\ 
\bfseries f017 & \bfseries f018 \\
\includegraphics[scale=0.25]{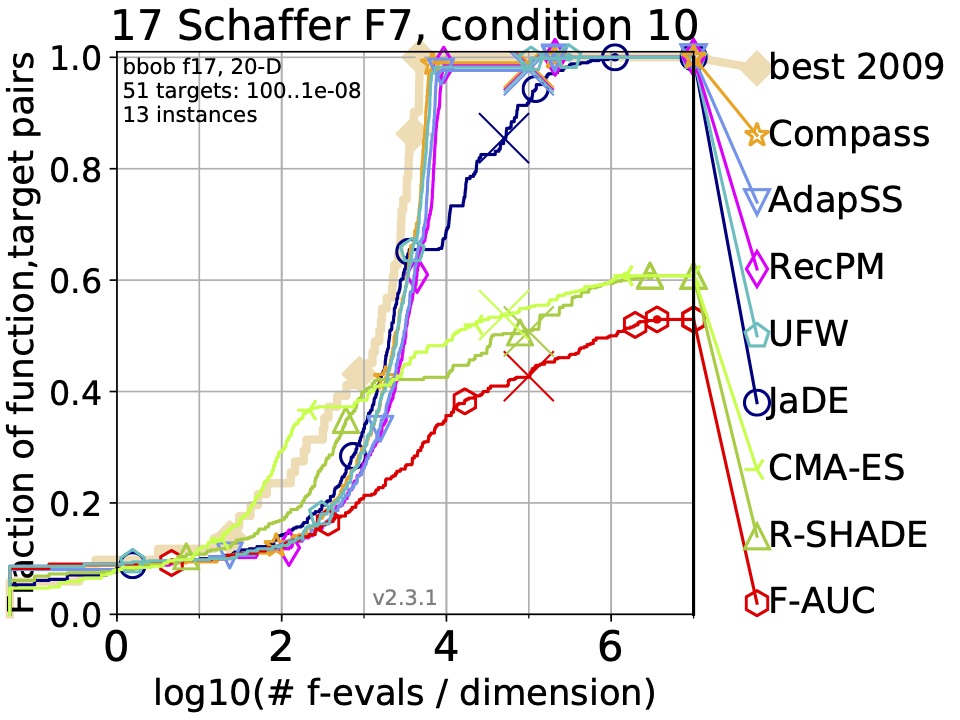} & \includegraphics[scale=0.25]{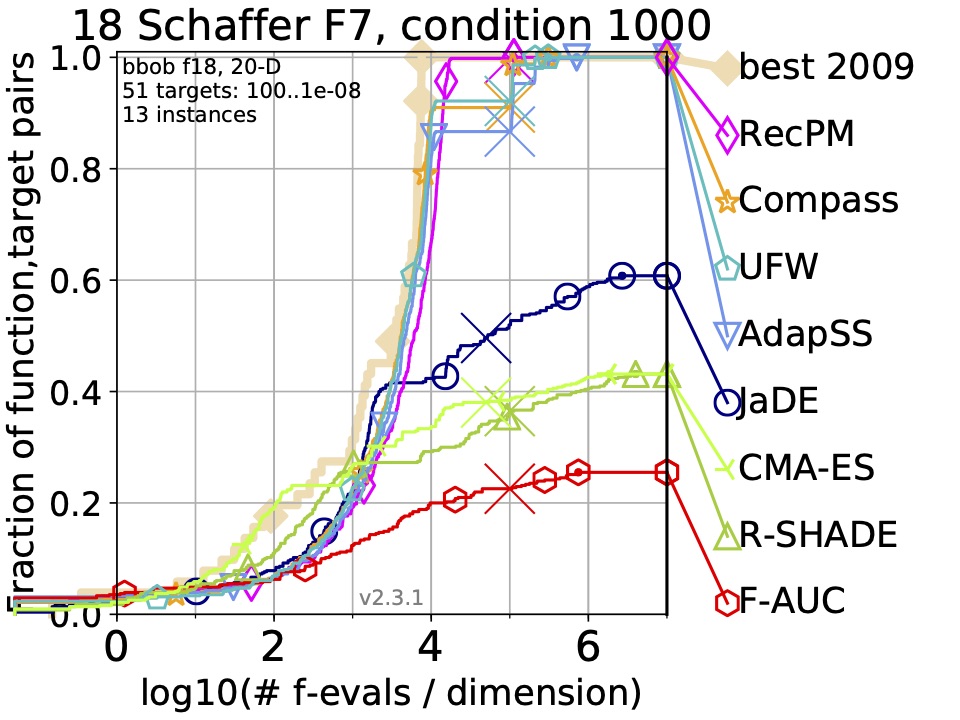} \\
\end{tabular}
\end{figure*}

\begin{figure*}[tbp]
\captionsetup{list=off}
\ContinuedFloat
\caption{ECDFs on test set. F-AUC for \FAUCMAB, UFW for \UAOSFW, RecPM for \RecPMaos, AdapSS for \PMAdapSS}\label{fig:ECDF-UFW2}
\begin{tabular}{@{}c@{}c@{}}
\bfseries f019 & \bfseries f020\\
\includegraphics[scale=0.25]{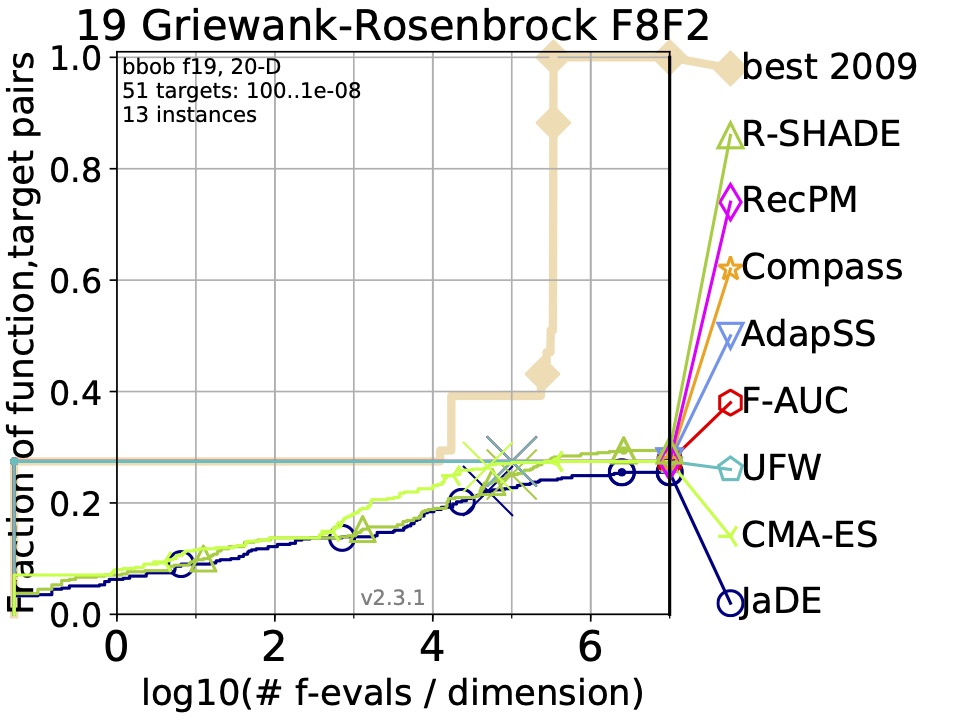} & \includegraphics[scale=0.25]{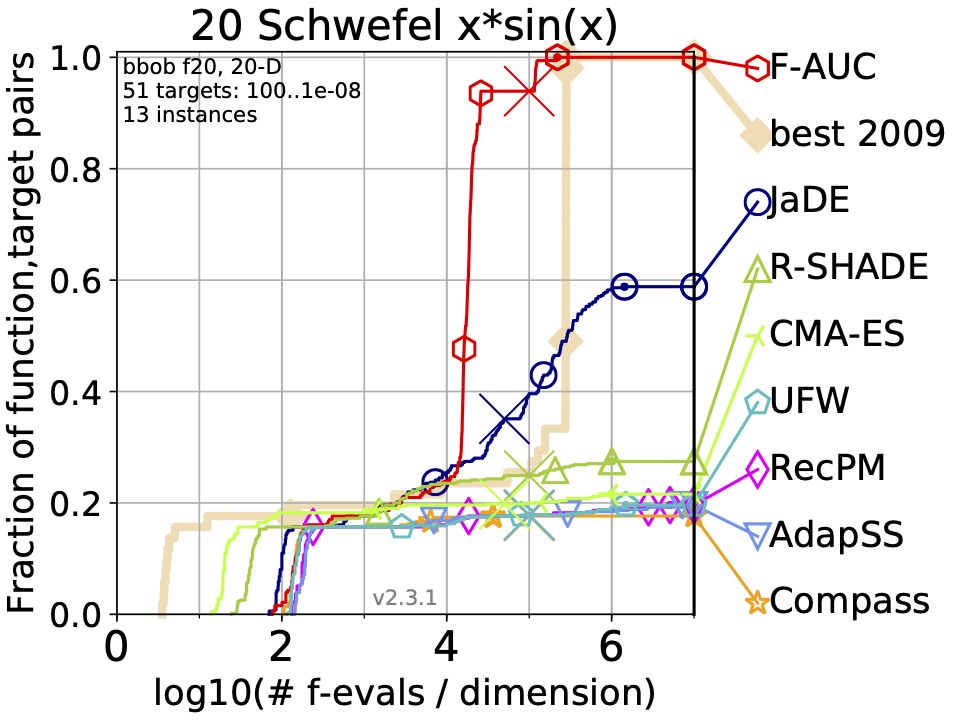}\\
\bfseries f021 & \bfseries f022\\
\includegraphics[scale=0.25]{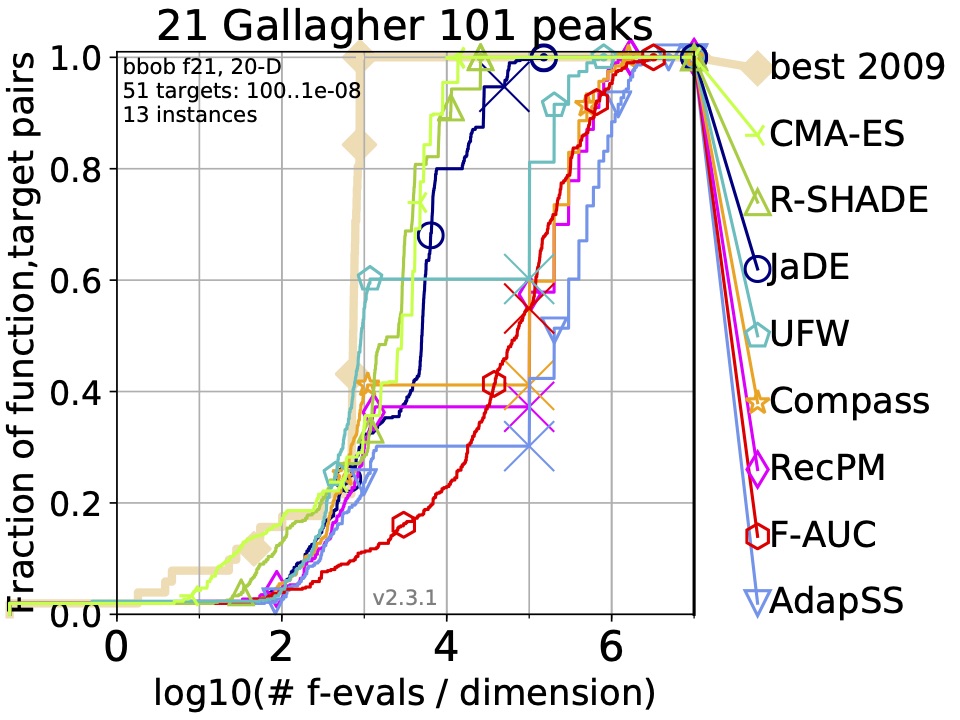} & \includegraphics[scale=0.25]{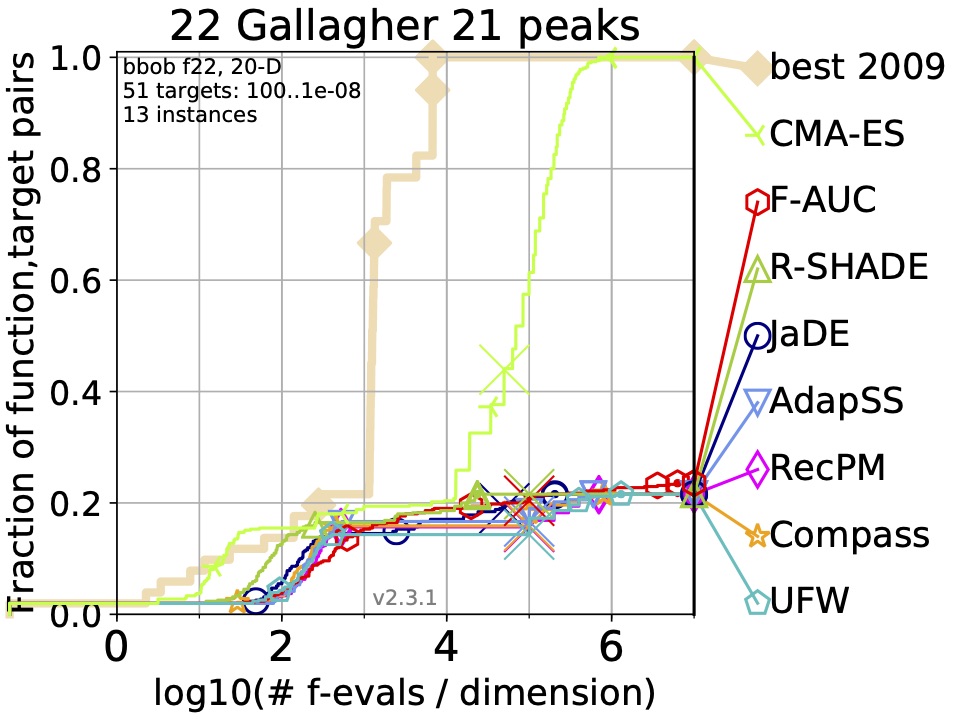} \\
\bfseries f023 & \bfseries f024\\
\includegraphics[scale=0.25]{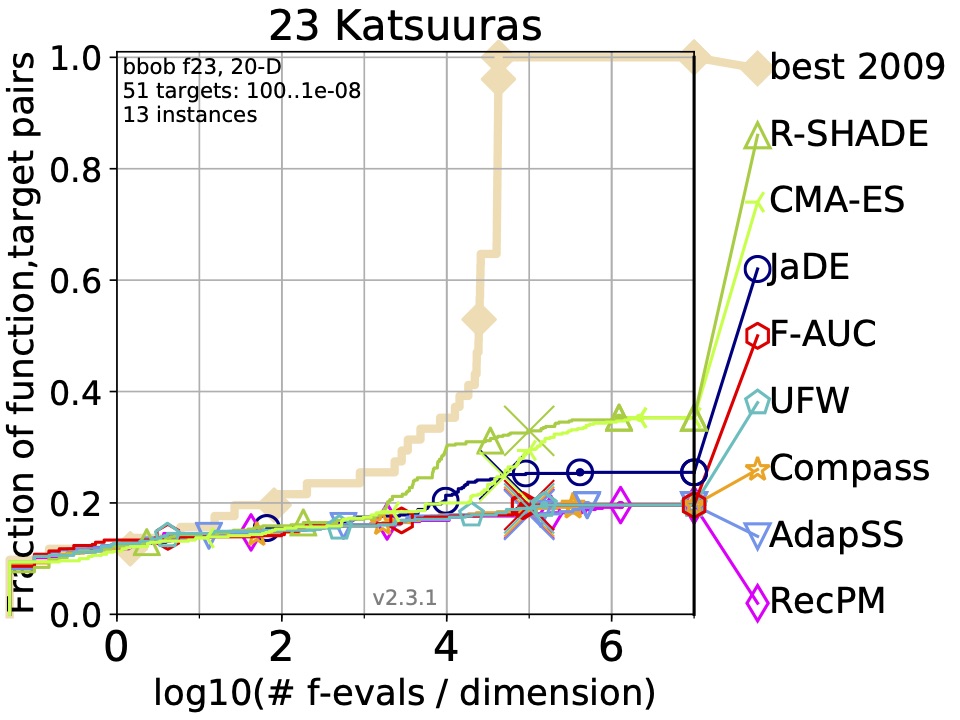} & \includegraphics[scale=0.25]{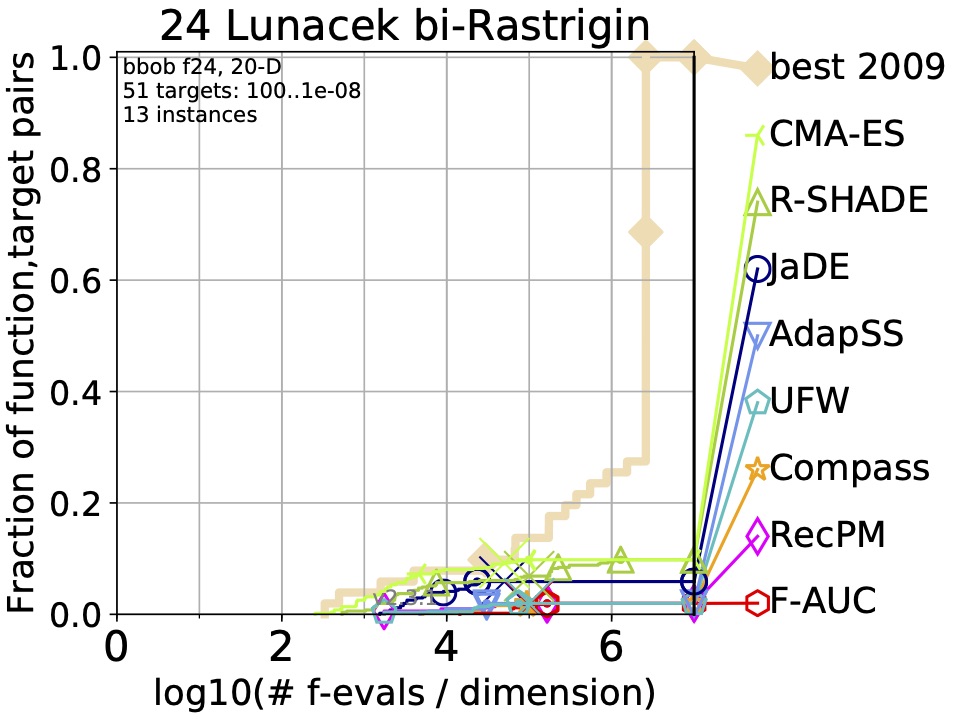}\\
\end{tabular}
\end{figure*}

\begin{figure*}[tbp]
\captionsetup{list=off}
\ContinuedFloat
\caption{ECDFs on test set. F-AUC for \FAUCMAB, UFW for \UAOSFW, RecPM for \RecPMaos, AdapSS for \PMAdapSS}\label{fig:ECDF-UFW2}
\begin{tabular}{@{}c@{}c@{}}
\bfseries Separable & \bfseries Low/moderate conditioning \\
\includegraphics[scale=0.25]{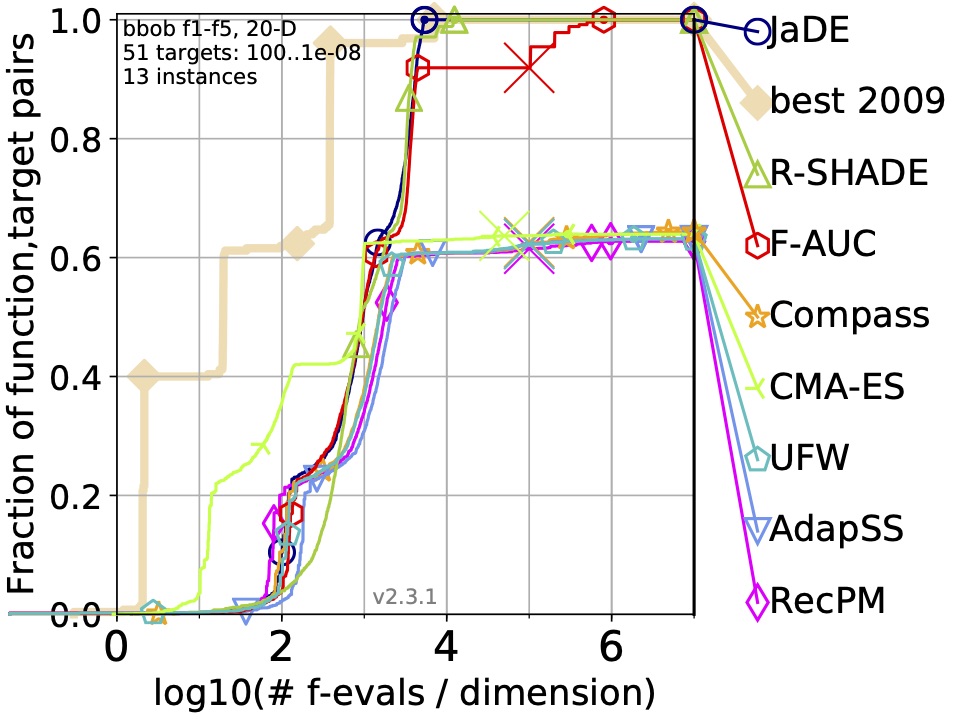} & \includegraphics[scale=0.25]{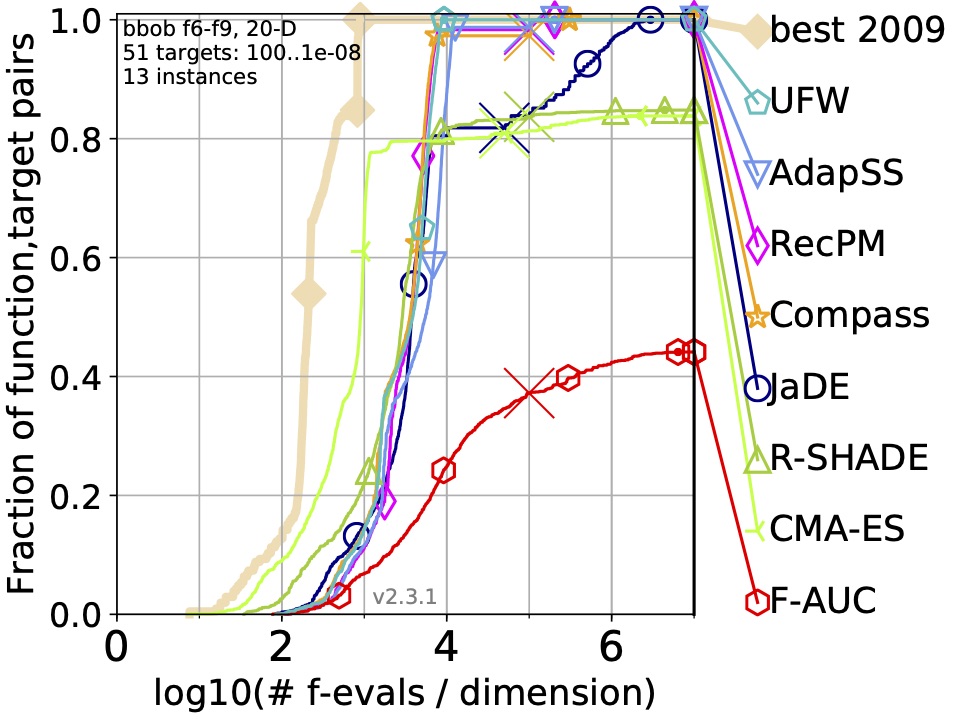} \\

\bfseries High conditioning & \bfseries multi modal (adequate structure)\\
\includegraphics[scale=0.25]{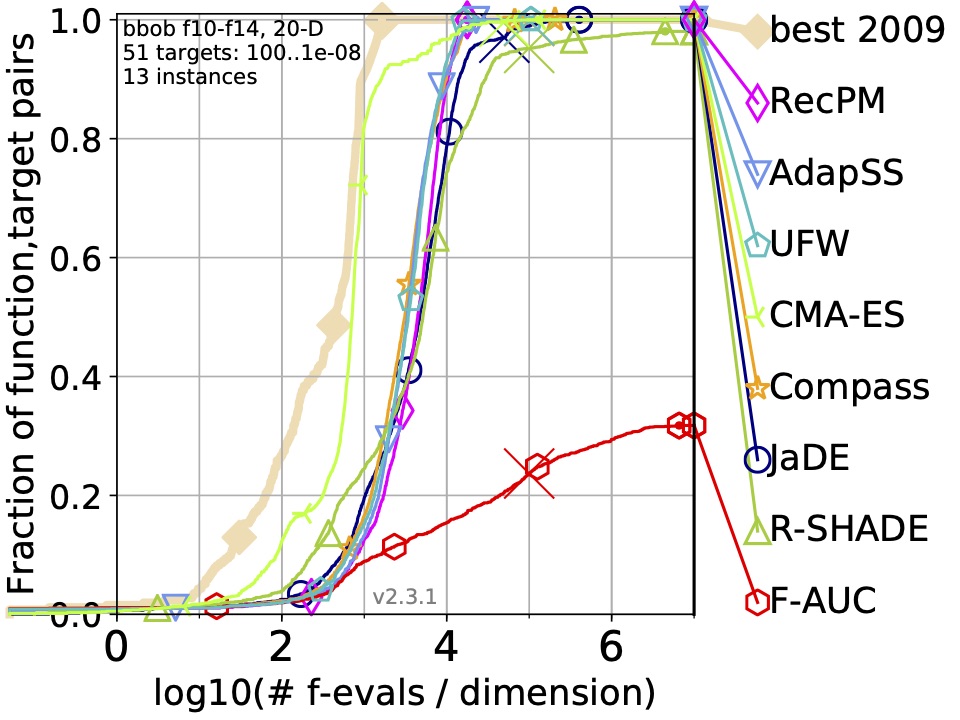} & \includegraphics[scale=0.25]{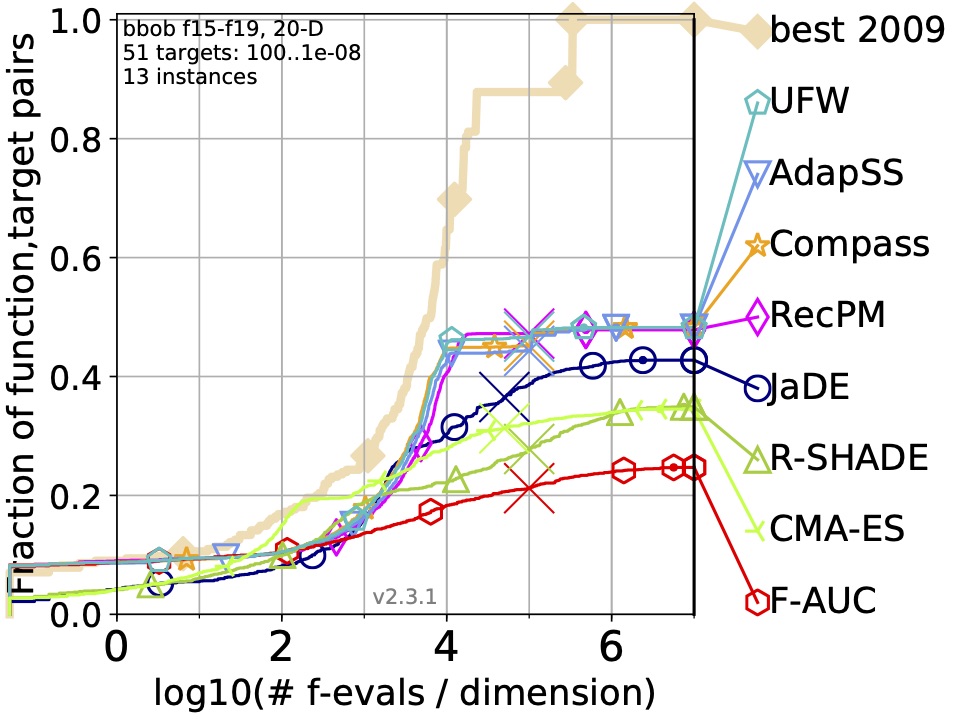} \\

\bfseries multi modal (weak structure) & \bfseries Overall \\
\includegraphics[scale=0.25]{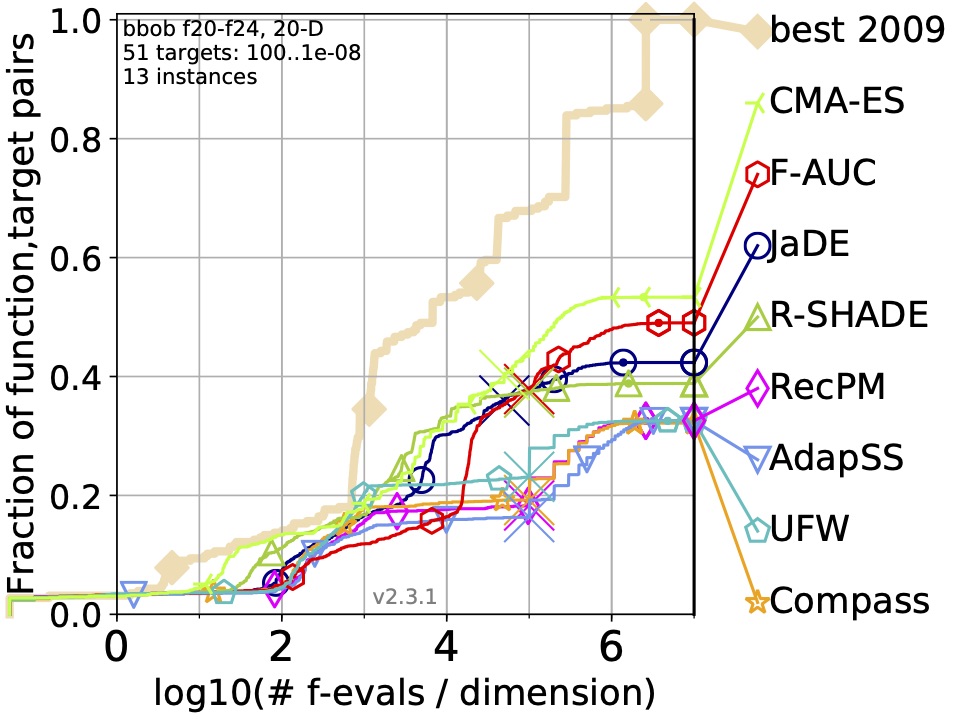} & \includegraphics[scale=0.25]{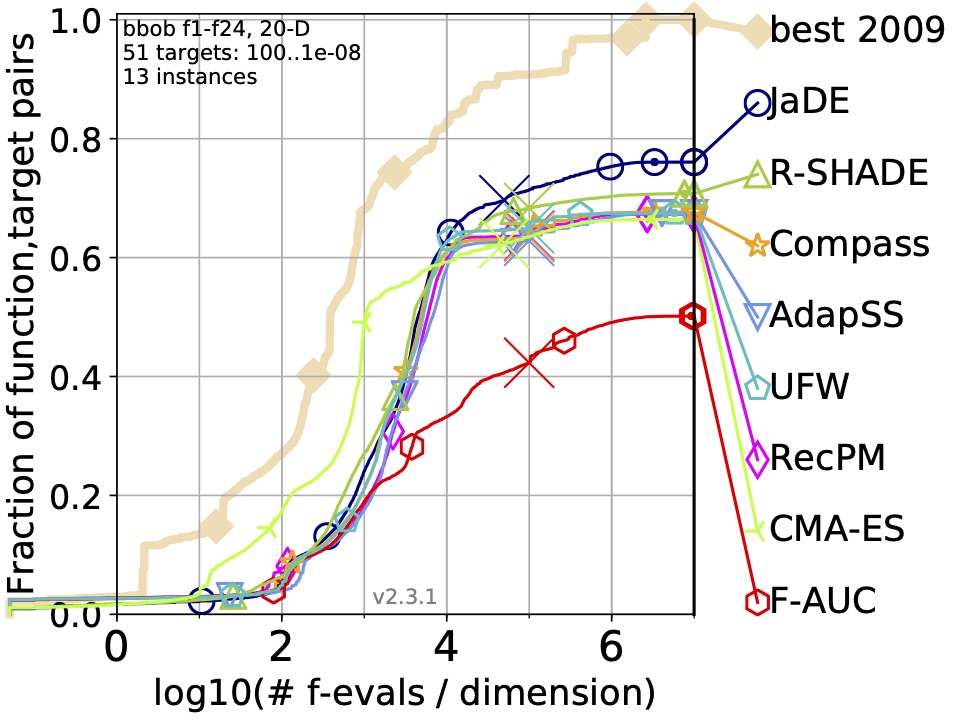} \\
\end{tabular}
\end{figure*}

\subsection{Comparison of \UAOSFW with non-AOS methods}
We compare the performance of \UAOSFW with non-AOS methods that is 
\JaDE and \RSHADE. The latter two are variants of DE utilising one strategy to evolve population of candidate solutions. 
\UAOSFW outperformed the two algorithms in three classes namely multi model (adequate structure), low/moderate and high conditioning problems. It excelled not only in solving maximum number of problems in these three class but also with higher speed. 
Overall, \UAOSFW is ranked third after \JaDE and \RSHADE in solving a number of problems. 

\subsection{Comparison of \UAOSFW trained on nine operators with \RecPMaos on four operators}
We are interested in analysing whether adding new operators in the set of operators for AOS method to choose from lead to any improvement in performance. Thus, we made an attempt in analysing the difference in performance of \UAOSFW trained on nine operators to \RecPMaos trained on four operators (\emph{RecPM1}). To analyse this, we compare the class-wise ECDF graphs of \UAOSFW shown in figure~\ref{fig:ECDF-UFW2} and \RecPMaos from paper~\cite{MudLopKaz2018ppsn}. There does not seem be any difference in performance of \UAOSFW and \emph{RecPM1} in solving separable, multi modal (adequate), low and high conditioning class problems, that is they solve the same number of problems. However, \UAOSFW solves the function instances of low/moderate conditioning and multi modal (adequate structure) with greater speed. In solving multi modal (weak structure), \UAOSFW outperforms RecPM1 in terms of speed and different targets. Thus, adding more operators helped solving more problems with higher speed, showing a better arrangement for exploration and exploitation of search space. 

\begin{figure*}[tbp]
\caption{Operator application and best fitness graphs for \DERecPMaos. Op1: \text{``rand/1''}, Op2: \text{``rand/2''}, Op3: \text{``rand-to-best/2''}, Op4: \text{``current-to-rand/1''}} \label{fig:recpm best fitness and operator four operators}
\begin{tabular}{@{}c@{}}
\textbf{$f05 i01$} \\
\includegraphics[width = 13cm, height = 5.8cm]{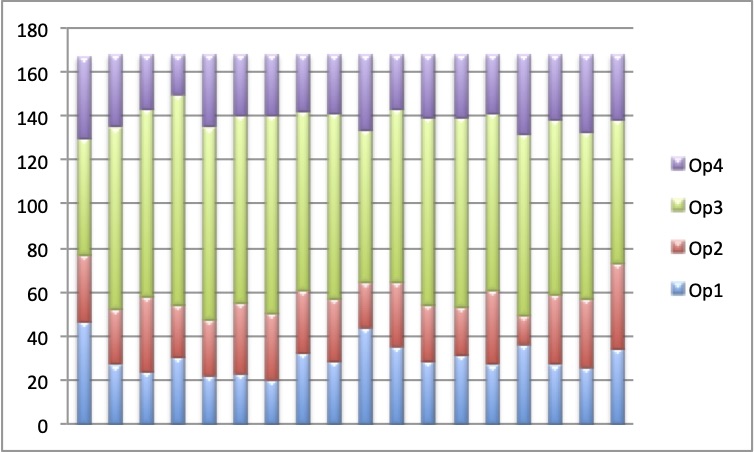} \\
\includegraphics[width = 13cm, height = 5.8cm]{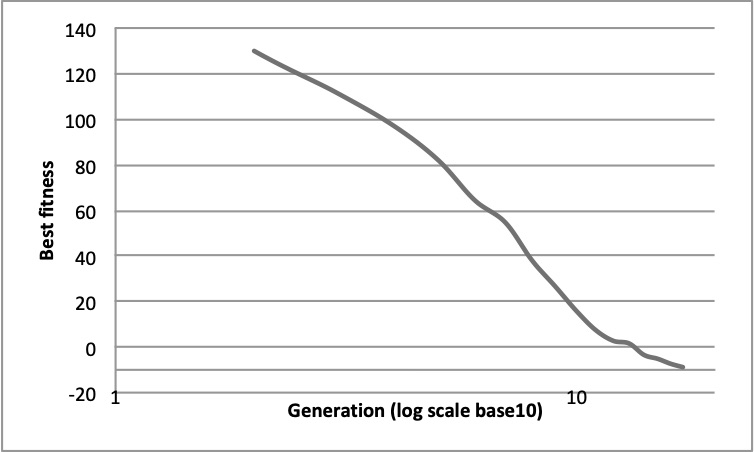} \\
\textbf{$f07 i01$}\\
\includegraphics[width = 13cm, height = 5.8cm]{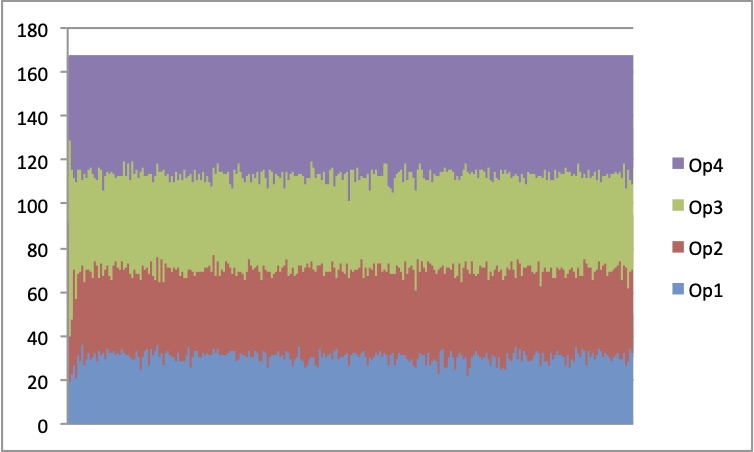} \\
\includegraphics[width = 13cm, height = 5.8cm]{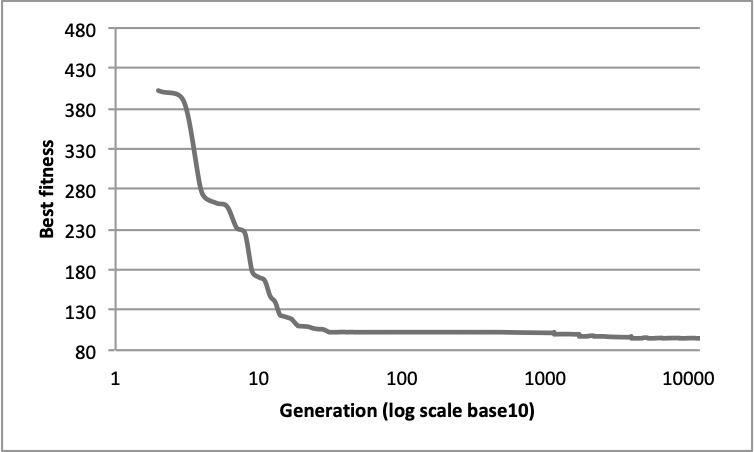} 
\end{tabular}
\end{figure*}

Next we try to identify the operators that lead to differences in performance of these two algorithms. Thus, further to analysing ECDF graphs we plan to better understand the impact of the selected operators by comparing the graphs generated as a result of running algorithms. We compare the performance of \RecPMi from paper~\cite{MudLopKaz2018ppsn} (figure~\ref{fig:recpm best fitness and operator four operators}) and by \UAOSFW (figure~\ref{fig:ufw best fitness and operator}) on functions $f05$ and $f07$ each with instance $i01$. These function instances are selected at random among the test instances of functions. Figure~\ref{fig:recpm best fitness and operator four operators} is a result of four operator applications by \RecPMi and figure~\ref{fig:ufw best fitness and operator} is a result of utilising nine different operators by \UAOSFW in different generations. A value of \pmin = $0.17$ and $0.04$ controls the level of adaptation of operators in algorithms \RecPMi and \UAOSFW respectively. To distinguish the operators, we represent them with different colors. The operators utilised by different candidate solutions to produce offspring in a generation are represented by a bar, all raising to the same level. For \RecPMi and \UAOSFW, population size is $168$ and $262$ respectively. The second figure for a function instance shows the best fitness seen as the generation progresses. 
To show the operator applications from each class, we have included the operator selections and best fitness progress for function instances $f04 i02$, $f08 i10$, $f13 i04$, $f17 i02$ and $f23 i04$, shown in figure~\ref{fig:ufw best fitness and operator}.
\begin{figure*}[tbp]
\caption{Operator application and best fitness graphs for \UAOSFW. Op1: \text{``rand/2''}, Op2: \text{``best/1''}, Op3: \text{``current-to-best/1''}, Op4: \text{``best/2''}, Op5: \text{``rand/1''}, Op6: \text{``rand-to-best/2''}, Op7: \text{``curr-to-rand/1''}, op8: \text{``curr-to-pbest/1''}, Op9: \text{``curr-to-pbest/1(archived)''}}\label{fig:ufw best fitness and operator}
\begin{tabular}{@{}c@{}}
\textbf{$f05 i01$} \\
\includegraphics[width = 13cm, height = 5.8cm]{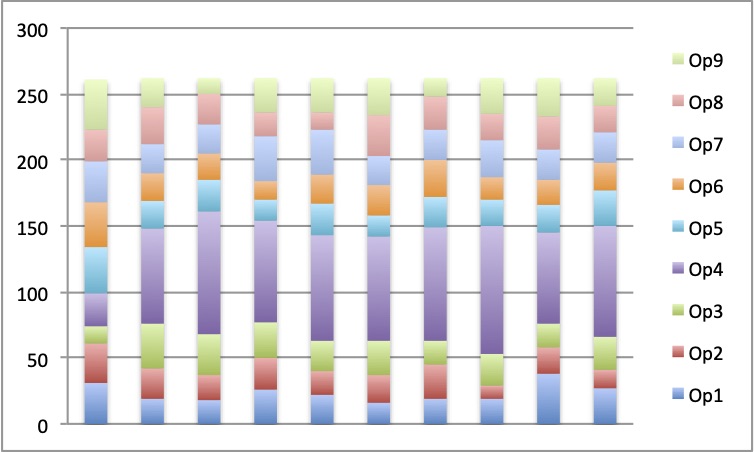} \\
\includegraphics[width = 13cm, height = 5.8cm]{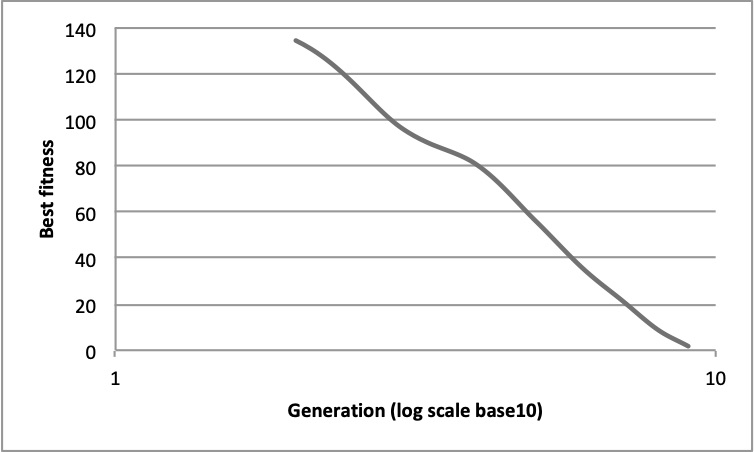} \\
\textbf{$f07 i01$}\\
\includegraphics[width = 13cm, height = 5.8cm]{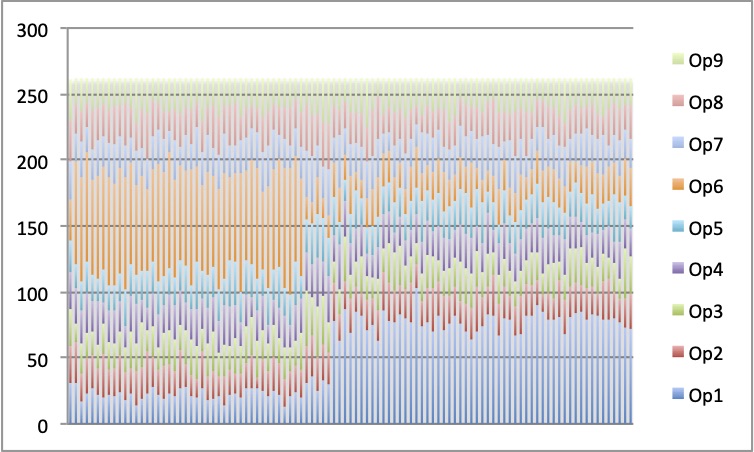} \\
\includegraphics[width = 13cm, height = 5.8cm]{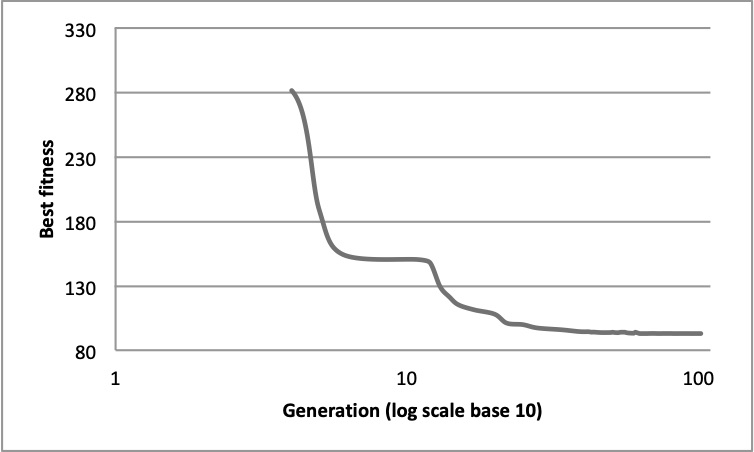} \\
\end{tabular}
\end{figure*}
\begin{figure*}[tbp]
\captionsetup{list=off}
\ContinuedFloat
\caption{Operator application and best fitness graphs for \UAOSFW. Op1: \text{``rand/2''}, Op2: \text{``best/1''}, Op3: \text{``current-to-best/1''}, Op4: \text{``best/2''}, Op5: \text{``rand/1''}, Op6: \text{``rand-to-best/2''}, Op7: \text{``curr-to-rand/1''}, op8: \text{``curr-to-pbest/1''}, Op9: \text{``curr-to-pbest/1(archived)''}}\label{fig:ufw best fitness and operator}
\begin{tabular}{@{}c@{}}
\textbf{$f04 i02$} \\
\includegraphics[width = 13cm, height = 5.8cm]{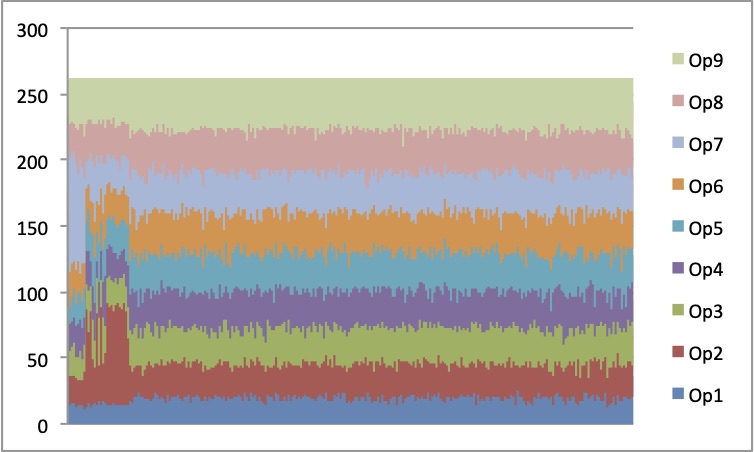} \\
\includegraphics[width = 13cm, height = 5.8cm]{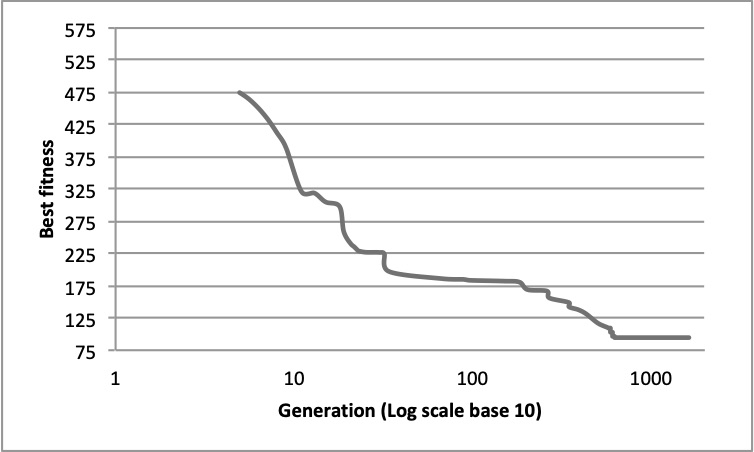} \\
\textbf{$f08 i10$}\\
\includegraphics[width = 13cm, height = 5.8cm]{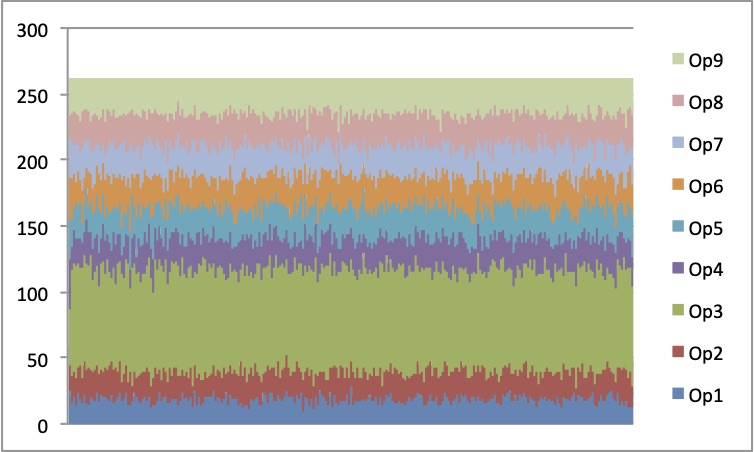} \\
\includegraphics[width = 13cm, height = 5.8cm]{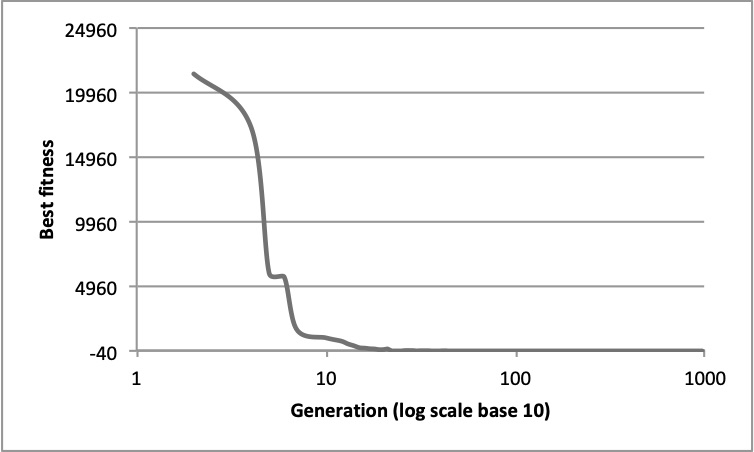} \\
\end{tabular}
\end{figure*}
\begin{figure*}[tbp]
\captionsetup{list=off}
\ContinuedFloat
\caption{Operator application and best fitness graphs for \UAOSFW. Op1: \text{``rand/2''}, Op2: \text{``best/1''}, Op3: \text{``current-to-best/1''}, Op4: \text{``best/2''}, Op5: \text{``rand/1''}, Op6: \text{``rand-to-best/2''}, Op7: \text{``curr-to-rand/1''}, op8: \text{``curr-to-pbest/1''}, Op9: \text{``curr-to-pbest/1(archived)''}}\label{fig:ufw best fitness and operator}
\begin{tabular}{@{}c@{}}
\textbf{$f13 i14$} \\
\includegraphics[width = 13cm, height = 5.8cm]{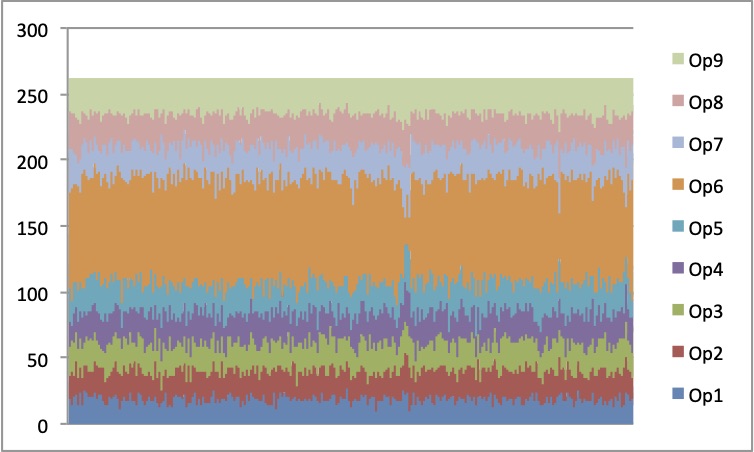} \\
\includegraphics[width = 13cm, height = 5.8cm]{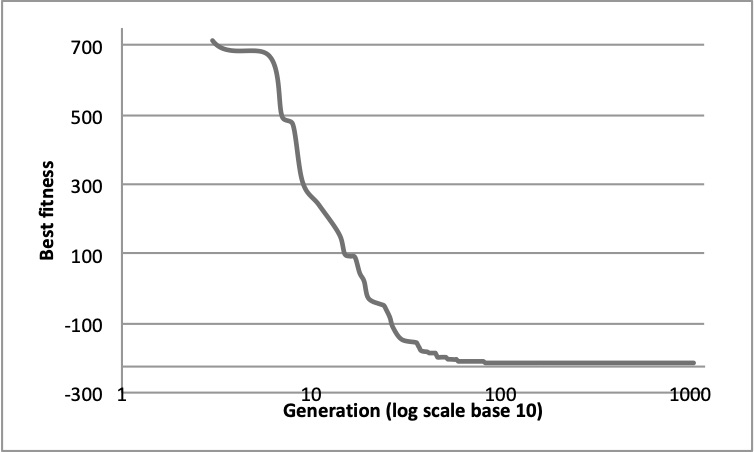} \\
\textbf{$f17 i02$}\\
\includegraphics[width = 13cm, height = 5.8cm]{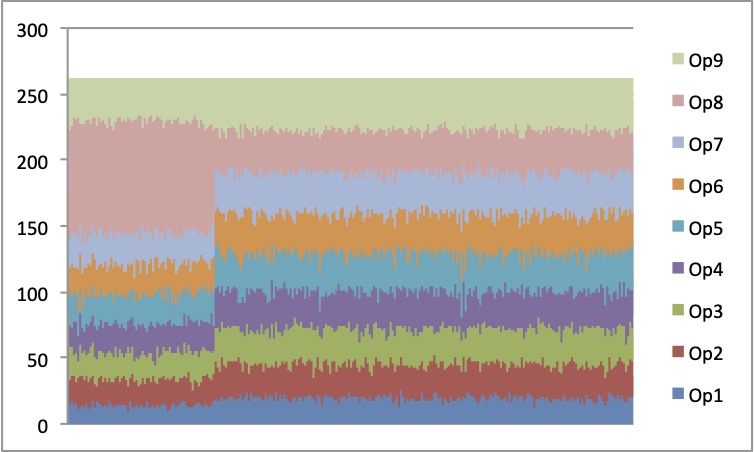} \\
\includegraphics[width = 13cm, height = 5.8cm]{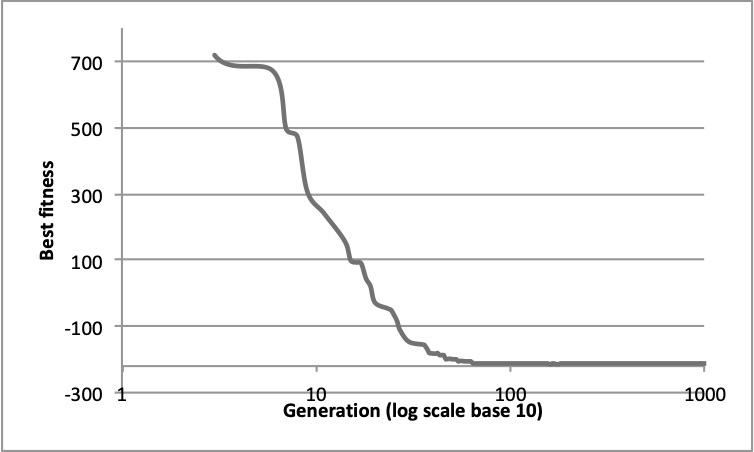} \\
\end{tabular}
\end{figure*}
\begin{figure*}[tbp]
\captionsetup{list=off}
\ContinuedFloat
\caption{Operator application and best fitness graphs for \UAOSFW. Op1: \text{``rand/2''}, Op2: \text{``best/1''}, Op3: \text{``current-to-best/1''}, Op4: \text{``best/2''}, Op5: \text{``rand/1''}, Op6: \text{``rand-to-best/2''}, Op7: \text{``curr-to-rand/1''}, op8: \text{``curr-to-pbest/1''}, Op9: \text{``curr-to-pbest/1(archived)''}}\label{fig:ufw best fitness and operator}
\begin{tabular}{@{}c@{}}
\textbf{$f23 i04$} \\
\includegraphics[width = 13cm, height = 5.8cm]{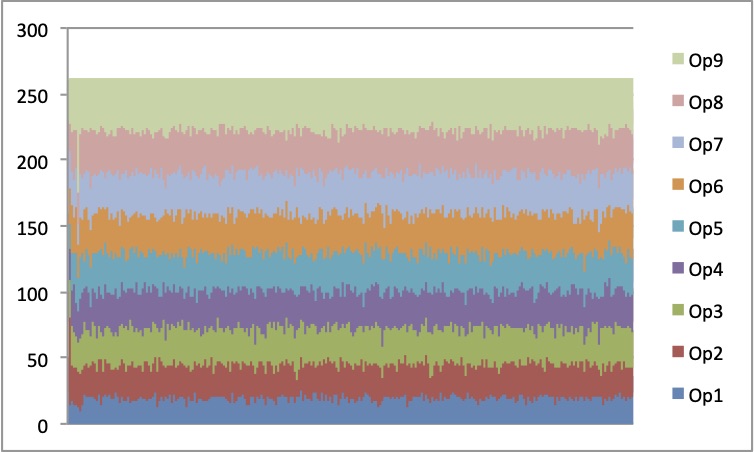} \\
\includegraphics[width = 13cm, height = 5.8cm]{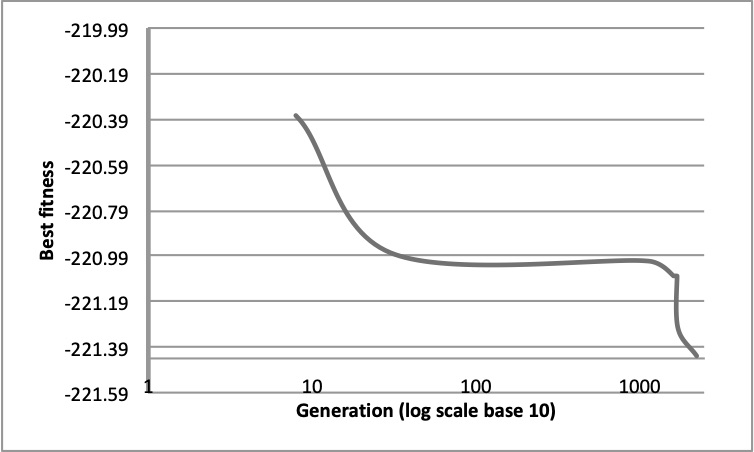} \\
\end{tabular}
\end{figure*}

\newcommand{\bbobdatapathuaosfw}{figures/} 
\providecommand{\bbobecdfcaptionsinglefunctionssingledim}[1]{
Empirical cumulative distribution of simulated (bootstrapped)
             runtimes, measured in number of objective function evaluations,
             divided by dimension (FEvals/DIM) for the $51$ 
             targets $10^{[-8..2]}$ in dimension #1.
}
\providecommand{\cocoversion}{\hspace{\textwidth}\scriptsize\sffamily{}\color{Gray}Data produced with COCO v2.3.1}
\providecommand{\numofalgs}{7}
\providecommand{\bbobECDFslegend}[1]{
Bootstrapped empirical cumulative distribution of the number of objective function evaluations divided by dimension (FEvals/DIM) for $51$ targets with target precision in $10^{[-8..2]}$ for all functions and subgroups in #1-D. As reference algorithm, the best algorithm from BBOB 2009 is shown as light thick line with diamond markers.
}
\providecommand{\bbobppfigslegend}[1]{
Average running time (\aRT\ in number of $f$-evaluations
                    as $\log_{10}$ value), divided by dimension for target function value $10^{-8}$
                    versus dimension. Slanted grid lines indicate quadratic scaling with the dimension. Different symbols correspond to different algorithms given in the legend of #1. Light symbols give the maximum number of function evaluations from the longest trial divided by dimension. Black stars indicate a statistically better result compared to all other algorithms with $p<0.01$ and Bonferroni correction number of dimensions (six).  
Legend: 
{\color{NavyBlue}$\circ$}: \algorithmA
, {\color{Magenta}$\diamondsuit$}: \algorithmB
, {\color{Orange}$\star$}: \algorithmC
, {\color{CornflowerBlue}$\triangledown$}: \algorithmD
, {\color{red}$\varhexagon$}: \algorithmE
, {\color{YellowGreen}$\triangle$}: \algorithmF
, {\color{cyan}$\pentagon$}: \algorithmG
}

\providecommand{\pptablesheader}{
\begin{tabularx}{1.0\textwidth}{@{}c@{}|*{7}{@{}r@{}X@{}}|@{}r@{}@{}l@{}}
$\Delta f_\mathrm{opt}$ & \multicolumn{2}{@{\,}l@{\,}}{1e1} & \multicolumn{2}{@{\,}l@{\,}}{1e0} & \multicolumn{2}{@{\,}l@{\,}}{1e-1} & \multicolumn{2}{@{\,}l@{\,}}{1e-2} & \multicolumn{2}{@{\,}l@{\,}}{1e-3} & \multicolumn{2}{@{\,}l@{\,}}{1e-5} & \multicolumn{2}{@{\,}l@{\,}}{1e-7} & \multicolumn{2}{|@{}l@{}}{\#succ}\\\hline
}
\providecommand{\algBtables}{\StrLeft{JaDE}{\ntables}}
\providecommand{\algAtables}{\StrLeft{R-SHADE}{\ntables}}
\providecommand{\ntables}{7}
\providecommand{\ntables}{7}
\providecommand{\ntables}{7}
\providecommand{\bbobpptablesmanylegend}[1]{%
        Average runtime (\aRT\ in number of function 
        evaluations) divided by the respective best \aRT\ measured during BBOB-2009 in
        #1.
        This \aRT\ ratio and, in braces as dispersion measure, the half difference between
        10 and 90\%-tile of bootstrapped run lengths appear for each algorithm and 
        target, the corresponding reference \aRT\
        in the first row. The different target \Df-values are shown in the top row.
        \#succ is the number of trials that reached the (final) target
        $\fopt + 10^{-8}$.
        The median number of conducted function evaluations is additionally given in 
        \textit{italics}, if the target in the last column was never reached.
        Entries, succeeded by a star, are statistically significantly better (according to
        the rank-sum test) when compared to all other algorithms of the table, with
        $p = 0.05$ or $p = 10^{-k}$ when the number $k$ following the star is larger
        than 1, with Bonferroni correction by the number of functions (24). A $\downarrow$ indicates the same tested against the best algorithm from BBOB 2009. Best results are printed in bold.
        \cocoversion}
\providecommand{\algsfolder}{U-AOS_R-SHA_RecPM_PM-Ad_F-AUC_Compa_JaDE_et_al/}
\providecommand{\algorithmA}{Compass}
\providecommand{\algorithmB}{F-AUC-MAB}
\providecommand{\algorithmC}{JaDE}
\providecommand{\algorithmD}{PM-AdapSS}
\providecommand{\algorithmE}{R-SHADE}
\providecommand{\algorithmF}{RecPM-AOS}
\providecommand{\algorithmG}{U-AOS-FW}
 
\begin{table*}[tp]\tiny
\caption{Average runtime \aRT}\label{tab:art-ufw1}
\mbox{
\begin{minipage}[t]{0.99\textwidth}\tiny
\centering
\providecommand{\algAtables}{\StrLeft{U-AOS-FW}{\ntables}}
\providecommand{\algBtables}{\StrLeft{R-SHADE}{\ntables}}
\providecommand{\algCtables}{\StrLeft{RecPM-AOS}{\ntables}}
\providecommand{\algDtables}{\StrLeft{PM-AdapSS}{\ntables}}
\providecommand{\algEtables}{\StrLeft{F-AUC-MAB}{\ntables}}
\providecommand{\algFtables}{\StrLeft{Compass}{\ntables}}
\providecommand{\algGtables}{\StrLeft{JaDE}{\ntables}}

\begin{tabularx}{1.0\textwidth}{@{}c@{}|*{7}{@{}r@{}X@{}}|@{}r@{}@{}l@{}}
$\Delta f_\mathrm{opt}$ & \multicolumn{2}{@{\,}l@{\,}}{1e1} & \multicolumn{2}{@{\,}l@{\,}}{1e0} & \multicolumn{2}{@{\,}l@{\,}}{1e-1} & \multicolumn{2}{@{\,}l@{\,}}{1e-2} & \multicolumn{2}{@{\,}l@{\,}}{1e-3} & \multicolumn{2}{@{\,}l@{\,}}{1e-5} & \multicolumn{2}{@{\,}l@{\,}}{1e-7} & \multicolumn{2}{|@{}l@{}}{\#succ}\\\hline

\textbf{f1} & \multicolumn{2}{@{}c@{}}{43 \quad} & \multicolumn{2}{@{}c@{}}{43 \quad} & \multicolumn{2}{@{}c@{}}{43 \quad} & \multicolumn{2}{@{}c@{}}{43 \quad} & \multicolumn{2}{@{}c@{}}{43 \quad} & \multicolumn{2}{@{}c@{}}{43 \quad} & \multicolumn{2}{@{}c@{}|}{43} & 15 & /15\\\hline
U-AOS-FW \hspace*{\fill} & 73 & \mbox{\tiny (11)} & 141 & \mbox{\tiny (8)} & 217 & \mbox{\tiny (11)} & 294 & \mbox{\tiny (17)} & 368 & \mbox{\tiny (18)} & 527 & \mbox{\tiny (21)} & 670 & \mbox{\tiny (35)} & 13 & /13\\
R-SHADE \hspace*{\fill} & \textbf{28} & \textbf{}\mbox{\tiny (7)}$^{\star3}$ & \textbf{61} & \textbf{}\mbox{\tiny (10)}$^{\star3}$ & \textbf{91} & \textbf{}\mbox{\tiny (11)}$^{\star3}$ & \textbf{122} & \textbf{}\mbox{\tiny (14)}$^{\star3}$ & \textbf{152} & \textbf{}\mbox{\tiny (13)}$^{\star3}$ & \textbf{211} & \textbf{}\mbox{\tiny (17)}$^{\star3}$ & \textbf{269} & \textbf{}\mbox{\tiny (11)}$^{\star3}$ & 13 & /13\\
RecPM-AOS \hspace*{\fill} & 86 & \mbox{\tiny (16)} & 169 & \mbox{\tiny (14)} & 255 & \mbox{\tiny (29)} & 345 & \mbox{\tiny (15)} & 434 & \mbox{\tiny (37)} & 611 & \mbox{\tiny (38)} & 790 & \mbox{\tiny (56)} & 13 & /13\\
PM-AdapSS \hspace*{\fill} & 92 & \mbox{\tiny (14)} & 192 & \mbox{\tiny (16)} & 297 & \mbox{\tiny (14)} & 400 & \mbox{\tiny (27)} & 507 & \mbox{\tiny (16)} & 710 & \mbox{\tiny (17)} & 926 & \mbox{\tiny (16)} & 13 & /13\\
F-AUC-MAB \hspace*{\fill} & 58 & \mbox{\tiny (7)} & 113 & \mbox{\tiny (6)} & 164 & \mbox{\tiny (9)} & 219 & \mbox{\tiny (16)} & 270 & \mbox{\tiny (32)} & 376 & \mbox{\tiny (24)} & 480 & \mbox{\tiny (45)} & 13 & /13\\
Compass \hspace*{\fill} & 67 & \mbox{\tiny (9)} & 141 & \mbox{\tiny (10)} & 217 & \mbox{\tiny (12)} & 291 & \mbox{\tiny (17)} & 368 & \mbox{\tiny (10)} & 517 & \mbox{\tiny (12)} & 667 & \mbox{\tiny (13)} & 13 & /13\\
JaDE \hspace*{\fill} & 48 & \mbox{\tiny (10)} & 94 & \mbox{\tiny (9)} & 144 & \mbox{\tiny (10)} & 193 & \mbox{\tiny (8)} & 241 & \mbox{\tiny (9)} & 341 & \mbox{\tiny (10)} & 438 & \mbox{\tiny (12)} & 13 & /13\\

\end{tabularx} 
\providecommand{\algAtables}{\StrLeft{U-AOS-FW}{\ntables}}
\providecommand{\algBtables}{\StrLeft{R-SHADE}{\ntables}}
\providecommand{\algCtables}{\StrLeft{RecPM-AOS}{\ntables}}
\providecommand{\algDtables}{\StrLeft{PM-AdapSS}{\ntables}}
\providecommand{\algEtables}{\StrLeft{F-AUC-MAB}{\ntables}}
\providecommand{\algFtables}{\StrLeft{Compass}{\ntables}}
\providecommand{\algGtables}{\StrLeft{JaDE}{\ntables}}

\begin{tabularx}{1.0\textwidth}{@{}c@{}|*{7}{@{}r@{}X@{}}|@{}r@{}@{}l@{}}
\hline

\textbf{f2} & \multicolumn{2}{@{}c@{}}{385 \quad} & \multicolumn{2}{@{}c@{}}{386 \quad} & \multicolumn{2}{@{}c@{}}{387 \quad} & \multicolumn{2}{@{}c@{}}{388 \quad} & \multicolumn{2}{@{}c@{}}{390 \quad} & \multicolumn{2}{@{}c@{}}{391 \quad} & \multicolumn{2}{@{}c@{}|}{393} & 15 & /15\\\hline
U-AOS-FW \hspace*{\fill} & 40 & \mbox{\tiny (3)} & 48 & \mbox{\tiny (2)} & 57 & \mbox{\tiny (4)} & 66 & \mbox{\tiny (4)} & 74 & \mbox{\tiny (5)} & 91 & \mbox{\tiny (5)} & 107 & \mbox{\tiny (5)} & 13 & /13\\
R-SHADE \hspace*{\fill} & \textbf{18} & \textbf{}\mbox{\tiny (2)} & \textbf{21} & \textbf{}\mbox{\tiny (2)} & \textbf{24} & \textbf{}\mbox{\tiny (2)} & \textbf{28} & \textbf{}\mbox{\tiny (2)} & \textbf{31} & \textbf{}\mbox{\tiny (2)} & \textbf{37} & \textbf{}\mbox{\tiny (3)} & \textbf{43} & \textbf{}\mbox{\tiny (3)} & 13 & /13\\
RecPM-AOS \hspace*{\fill} & 45 & \mbox{\tiny (5)} & 55 & \mbox{\tiny (4)} & 65 & \mbox{\tiny (3)} & 74 & \mbox{\tiny (5)} & 83 & \mbox{\tiny (4)} & 103 & \mbox{\tiny (6)} & 122 & \mbox{\tiny (6)} & 13 & /13\\
PM-AdapSS \hspace*{\fill} & 55 & \mbox{\tiny (3)} & 67 & \mbox{\tiny (0.9)} & 78 & \mbox{\tiny (3)} & 90 & \mbox{\tiny (3)} & 101 & \mbox{\tiny (3)} & 124 & \mbox{\tiny (2)} & 147 & \mbox{\tiny (2)} & 13 & /13\\
F-AUC-MAB \hspace*{\fill} & 24 & \mbox{\tiny (2)} & 30 & \mbox{\tiny (2)} & 35 & \mbox{\tiny (7)} & 40 & \mbox{\tiny (7)} & 46 & \mbox{\tiny (5)} & 56 & \mbox{\tiny (15)} & 67 & \mbox{\tiny (8)} & 13 & /13\\
Compass \hspace*{\fill} & 39 & \mbox{\tiny (1)} & 47 & \mbox{\tiny (2)} & 55 & \mbox{\tiny (2)} & 64 & \mbox{\tiny (2)} & 71 & \mbox{\tiny (2)} & 88 & \mbox{\tiny (2)} & 104 & \mbox{\tiny (2)} & 13 & /13\\
JaDE \hspace*{\fill} & 28 & \mbox{\tiny (1)} & 33 & \mbox{\tiny (1)} & 39 & \mbox{\tiny (2)} & 44 & \mbox{\tiny (2)} & 49 & \mbox{\tiny (2)} & 61 & \mbox{\tiny (3)} & 71 & \mbox{\tiny (3)} & 13 & /13\\

\end{tabularx}
\providecommand{\algAtables}{\StrLeft{U-AOS-FW}{\ntables}}
\providecommand{\algBtables}{\StrLeft{R-SHADE}{\ntables}}
\providecommand{\algCtables}{\StrLeft{RecPM-AOS}{\ntables}}
\providecommand{\algDtables}{\StrLeft{PM-AdapSS}{\ntables}}
\providecommand{\algEtables}{\StrLeft{F-AUC-MAB}{\ntables}}
\providecommand{\algFtables}{\StrLeft{Compass}{\ntables}}
\providecommand{\algGtables}{\StrLeft{JaDE}{\ntables}}

\begin{tabularx}{1.0\textwidth}{@{}c@{}|*{7}{@{}r@{}X@{}}|@{}r@{}@{}l@{}}
\hline

\textbf{f3} & \multicolumn{2}{@{}c@{}}{5066 \quad} & \multicolumn{2}{@{}c@{}}{7626 \quad} & \multicolumn{2}{@{}c@{}}{7635 \quad} & \multicolumn{2}{@{}c@{}}{7637 \quad} & \multicolumn{2}{@{}c@{}}{7643 \quad} & \multicolumn{2}{@{}c@{}}{7646 \quad} & \multicolumn{2}{@{}c@{}|}{7651} & 15 & /15\\\hline
U-AOS-FW \hspace*{\fill} & \multicolumn{2}{@{\,}l@{\,}}{$\infty$} & \multicolumn{2}{@{\,}l@{\,}}{$\infty$} & \multicolumn{2}{@{\,}l@{\,}}{$\infty$} & \multicolumn{2}{@{\,}l@{\,}}{$\infty$} & \multicolumn{2}{@{\,}l@{\,}}{$\infty$} & \multicolumn{2}{@{\,}l@{\,}}{$\infty$} & \multicolumn{2}{@{\,}l@{\,}|}{$\infty$\,\textit{2e6}} & 0 & /13\\
R-SHADE \hspace*{\fill} & 7 & .6\mbox{\tiny (0.5)} & 7 & .1\mbox{\tiny (0.3)} & 7 & .6\mbox{\tiny (0.3)} & 7 & .9\mbox{\tiny (0.2)} & \textbf{8} & \textbf{.1}\mbox{\tiny (0.2)} & \textbf{8} & \textbf{.4}\mbox{\tiny (0.2)}$^{\star3}$ & \textbf{8} & \textbf{.7}\mbox{\tiny (0.3)}$^{\star3}$ & 13 & /13\\
RecPM-AOS \hspace*{\fill} & \multicolumn{2}{@{\,}l@{\,}}{$\infty$} & \multicolumn{2}{@{\,}l@{\,}}{$\infty$} & \multicolumn{2}{@{\,}l@{\,}}{$\infty$} & \multicolumn{2}{@{\,}l@{\,}}{$\infty$} & \multicolumn{2}{@{\,}l@{\,}}{$\infty$} & \multicolumn{2}{@{\,}l@{\,}}{$\infty$} & \multicolumn{2}{@{\,}l@{\,}|}{$\infty$\,\textit{2e6}} & 0 & /13\\
PM-AdapSS \hspace*{\fill} & \multicolumn{2}{@{\,}l@{\,}}{$\infty$} & \multicolumn{2}{@{\,}l@{\,}}{$\infty$} & \multicolumn{2}{@{\,}l@{\,}}{$\infty$} & \multicolumn{2}{@{\,}l@{\,}}{$\infty$} & \multicolumn{2}{@{\,}l@{\,}}{$\infty$} & \multicolumn{2}{@{\,}l@{\,}}{$\infty$} & \multicolumn{2}{@{\,}l@{\,}|}{$\infty$\,\textit{2e6}} & 0 & /13\\
F-AUC-MAB \hspace*{\fill} & 11 & \mbox{\tiny (0.8)} & 8 & .4\mbox{\tiny (0.9)} & 8 & .7\mbox{\tiny (0.4)} & 9 & .0\mbox{\tiny (0.8)} & 9 & .3\mbox{\tiny (0.6)} & 10 & \mbox{\tiny (1)} & 11 & \mbox{\tiny (1)} & 13 & /13\\
Compass \hspace*{\fill} & 5069 & \mbox{\tiny (4441)} & \multicolumn{2}{@{\,}l@{\,}}{$\infty$} & \multicolumn{2}{@{\,}l@{\,}}{$\infty$} & \multicolumn{2}{@{\,}l@{\,}}{$\infty$} & \multicolumn{2}{@{\,}l@{\,}}{$\infty$} & \multicolumn{2}{@{\,}l@{\,}}{$\infty$} & \multicolumn{2}{@{\,}l@{\,}|}{$\infty$\,\textit{2e6}} & 0 & /13\\
JaDE \hspace*{\fill} & \textbf{6} & \textbf{.4}\mbox{\tiny (0.3)}$^{\star3}$ & \textbf{6} & \textbf{.0}\mbox{\tiny (0.2)}$^{\star3}$ & \textbf{6} & \textbf{.8}\mbox{\tiny (0.2)}$^{\star3}$ & \textbf{7} & \textbf{.6}\mbox{\tiny (0.2)}$^{\star}$ & 8 & .3\mbox{\tiny (0.1)} & 10 & \mbox{\tiny (0.1)} & 11 & \mbox{\tiny (0.1)} & 13 & /13\\

\end{tabularx}
\providecommand{\algAtables}{\StrLeft{U-AOS-FW}{\ntables}}
\providecommand{\algBtables}{\StrLeft{R-SHADE}{\ntables}}
\providecommand{\algCtables}{\StrLeft{RecPM-AOS}{\ntables}}
\providecommand{\algDtables}{\StrLeft{PM-AdapSS}{\ntables}}
\providecommand{\algEtables}{\StrLeft{F-AUC-MAB}{\ntables}}
\providecommand{\algFtables}{\StrLeft{Compass}{\ntables}}
\providecommand{\algGtables}{\StrLeft{JaDE}{\ntables}}

\begin{tabularx}{1.0\textwidth}{@{}c@{}|*{7}{@{}r@{}X@{}}|@{}r@{}@{}l@{}}
\hline

\textbf{f4} & \multicolumn{2}{@{}c@{}}{4722 \quad} & \multicolumn{2}{@{}c@{}}{7628 \quad} & \multicolumn{2}{@{}c@{}}{7666 \quad} & \multicolumn{2}{@{}c@{}}{7686 \quad} & \multicolumn{2}{@{}c@{}}{7700 \quad} & \multicolumn{2}{@{}c@{}}{7758 \quad} & \multicolumn{2}{@{}c@{}|}{1.4e5} & 9 & /15\\\hline
U-AOS-FW \hspace*{\fill} & 2563 & \mbox{\tiny (3709)} & \multicolumn{2}{@{\,}l@{\,}}{$\infty$} & \multicolumn{2}{@{\,}l@{\,}}{$\infty$} & \multicolumn{2}{@{\,}l@{\,}}{$\infty$} & \multicolumn{2}{@{\,}l@{\,}}{$\infty$} & \multicolumn{2}{@{\,}l@{\,}}{$\infty$} & \multicolumn{2}{@{\,}l@{\,}|}{$\infty$\,\textit{2e6}} & 0 & /13\\
R-SHADE \hspace*{\fill} & 9 & .5\mbox{\tiny (0.8)} & 8 & .3\mbox{\tiny (0.4)} & 11 & \mbox{\tiny (5)} & 12 & \mbox{\tiny (0.4)} & 12 & \mbox{\tiny (11)} & 12 & \mbox{\tiny (11)} & \textbf{0} & \textbf{.70}\mbox{\tiny (0.3)} & 13 & /13\\
RecPM-AOS \hspace*{\fill} & \multicolumn{2}{@{\,}l@{\,}}{$\infty$} & \multicolumn{2}{@{\,}l@{\,}}{$\infty$} & \multicolumn{2}{@{\,}l@{\,}}{$\infty$} & \multicolumn{2}{@{\,}l@{\,}}{$\infty$} & \multicolumn{2}{@{\,}l@{\,}}{$\infty$} & \multicolumn{2}{@{\,}l@{\,}}{$\infty$} & \multicolumn{2}{@{\,}l@{\,}|}{$\infty$\,\textit{2e6}} & 0 & /13\\
PM-AdapSS \hspace*{\fill} & 5416 & \mbox{\tiny (3812)} & \multicolumn{2}{@{\,}l@{\,}}{$\infty$} & \multicolumn{2}{@{\,}l@{\,}}{$\infty$} & \multicolumn{2}{@{\,}l@{\,}}{$\infty$} & \multicolumn{2}{@{\,}l@{\,}}{$\infty$} & \multicolumn{2}{@{\,}l@{\,}}{$\infty$} & \multicolumn{2}{@{\,}l@{\,}|}{$\infty$\,\textit{2e6}} & 0 & /13\\
F-AUC-MAB \hspace*{\fill} & 12 & \mbox{\tiny (2)} & 125 & \mbox{\tiny (0.7)} & 313 & \mbox{\tiny (587)} & 313 & \mbox{\tiny (520)} & 312 & \mbox{\tiny (260)} & 311 & \mbox{\tiny (451)} & 17 & \mbox{\tiny (25)} & 6 & /13\\
Compass \hspace*{\fill} & \multicolumn{2}{@{\,}l@{\,}}{$\infty$} & \multicolumn{2}{@{\,}l@{\,}}{$\infty$} & \multicolumn{2}{@{\,}l@{\,}}{$\infty$} & \multicolumn{2}{@{\,}l@{\,}}{$\infty$} & \multicolumn{2}{@{\,}l@{\,}}{$\infty$} & \multicolumn{2}{@{\,}l@{\,}}{$\infty$} & \multicolumn{2}{@{\,}l@{\,}|}{$\infty$\,\textit{2e6}} & 0 & /13\\
JaDE \hspace*{\fill} & \textbf{8} & \textbf{.1}\mbox{\tiny (0.3)}$^{\star2}$ & \textbf{7} & \textbf{.0}\mbox{\tiny (0.1)}$^{\star3}$ & \textbf{8} & \textbf{.0}\mbox{\tiny (0.2)}$^{\star3}$ & \textbf{8} & \textbf{.9}\mbox{\tiny (0.2)} & \textbf{10} & \textbf{}\mbox{\tiny (0.1)} & \textbf{11} & \textbf{}\mbox{\tiny (0.2)} & 0 & .71\mbox{\tiny (9e-3)} & 13 & /13\\

\end{tabularx}
\providecommand{\algAtables}{\StrLeft{U-AOS-FW}{\ntables}}
\providecommand{\algBtables}{\StrLeft{R-SHADE}{\ntables}}
\providecommand{\algCtables}{\StrLeft{RecPM-AOS}{\ntables}}
\providecommand{\algDtables}{\StrLeft{PM-AdapSS}{\ntables}}
\providecommand{\algEtables}{\StrLeft{F-AUC-MAB}{\ntables}}
\providecommand{\algFtables}{\StrLeft{Compass}{\ntables}}
\providecommand{\algGtables}{\StrLeft{JaDE}{\ntables}}

\begin{tabularx}{1.0\textwidth}{@{}c@{}|*{7}{@{}r@{}X@{}}|@{}r@{}@{}l@{}}
\hline

\textbf{f5} & \multicolumn{2}{@{}c@{}}{41 \quad} & \multicolumn{2}{@{}c@{}}{41 \quad} & \multicolumn{2}{@{}c@{}}{41 \quad} & \multicolumn{2}{@{}c@{}}{41 \quad} & \multicolumn{2}{@{}c@{}}{41 \quad} & \multicolumn{2}{@{}c@{}}{41 \quad} & \multicolumn{2}{@{}c@{}|}{41} & 15 & /15\\\hline
U-AOS-FW \hspace*{\fill} & 47 & \mbox{\tiny (8)} & 55 & \mbox{\tiny (7)} & 56 & \mbox{\tiny (12)} & 57 & \mbox{\tiny (7)} & 57 & \mbox{\tiny (9)} & 57 & \mbox{\tiny (9)} & 57 & \mbox{\tiny (11)} & 13 & /13\\
R-SHADE \hspace*{\fill} & 114 & \mbox{\tiny (11)} & 210 & \mbox{\tiny (19)} & 307 & \mbox{\tiny (25)} & 399 & \mbox{\tiny (30)} & 488 & \mbox{\tiny (19)} & 676 & \mbox{\tiny (34)} & 860 & \mbox{\tiny (45)} & 13 & /13\\
RecPM-AOS \hspace*{\fill} & \textbf{33} & \textbf{}\mbox{\tiny (4)} & \textbf{37} & \textbf{}\mbox{\tiny (8)} & \textbf{39} & \textbf{}\mbox{\tiny (7)}$^{\star2}$ & \textbf{39} & \textbf{}\mbox{\tiny (9)}$^{\star2}$ & \textbf{39} & \textbf{}\mbox{\tiny (7)}$^{\star2}$ & \textbf{39} & \textbf{}\mbox{\tiny (9)}$^{\star2}$ & \textbf{39} & \textbf{}\mbox{\tiny (8)}$^{\star2}$ & 13 & /13\\
PM-AdapSS \hspace*{\fill} & 69 & \mbox{\tiny (9)} & 83 & \mbox{\tiny (11)} & 85 & \mbox{\tiny (10)} & 85 & \mbox{\tiny (8)} & 85 & \mbox{\tiny (22)} & 85 & \mbox{\tiny (13)} & 85 & \mbox{\tiny (6)} & 13 & /13\\
F-AUC-MAB \hspace*{\fill} & 49 & \mbox{\tiny (3)} & 59 & \mbox{\tiny (9)} & 62 & \mbox{\tiny (11)} & 63 & \mbox{\tiny (10)} & 63 & \mbox{\tiny (8)} & 63 & \mbox{\tiny (8)} & 63 & \mbox{\tiny (7)} & 13 & /13\\
Compass \hspace*{\fill} & 42 & \mbox{\tiny (7)} & 50 & \mbox{\tiny (6)} & 51 & \mbox{\tiny (12)} & 51 & \mbox{\tiny (11)} & 51 & \mbox{\tiny (7)} & 51 & \mbox{\tiny (10)} & 51 & \mbox{\tiny (7)} & 13 & /13\\
JaDE \hspace*{\fill} & 42 & \mbox{\tiny (6)} & 52 & \mbox{\tiny (8)} & 53 & \mbox{\tiny (7)} & 53 & \mbox{\tiny (3)} & 53 & \mbox{\tiny (5)} & 53 & \mbox{\tiny (6)} & 53 & \mbox{\tiny (7)} & 13 & /13\\

\end{tabularx}
\providecommand{\algAtables}{\StrLeft{U-AOS-FW}{\ntables}}
\providecommand{\algBtables}{\StrLeft{R-SHADE}{\ntables}}
\providecommand{\algCtables}{\StrLeft{RecPM-AOS}{\ntables}}
\providecommand{\algDtables}{\StrLeft{PM-AdapSS}{\ntables}}
\providecommand{\algEtables}{\StrLeft{F-AUC-MAB}{\ntables}}
\providecommand{\algFtables}{\StrLeft{Compass}{\ntables}}
\providecommand{\algGtables}{\StrLeft{JaDE}{\ntables}}

\begin{tabularx}{1.0\textwidth}{@{}c@{}|*{7}{@{}r@{}X@{}}|@{}r@{}@{}l@{}}
\hline

\textbf{f6} & \multicolumn{2}{@{}c@{}}{1296 \quad} & \multicolumn{2}{@{}c@{}}{2343 \quad} & \multicolumn{2}{@{}c@{}}{3413 \quad} & \multicolumn{2}{@{}c@{}}{4255 \quad} & \multicolumn{2}{@{}c@{}}{5220 \quad} & \multicolumn{2}{@{}c@{}}{6728 \quad} & \multicolumn{2}{@{}c@{}|}{8409} & 15 & /15\\\hline
U-AOS-FW \hspace*{\fill} & 14 & \mbox{\tiny (2)} & 12 & \mbox{\tiny (1)} & 11 & \mbox{\tiny (0.6)} & 11 & \mbox{\tiny (1)} & 11 & \mbox{\tiny (1.0)} & 11 & \mbox{\tiny (0.5)} & 12 & \mbox{\tiny (0.3)} & 13 & /13\\
R-SHADE \hspace*{\fill} & \textbf{4} & \textbf{.1}\mbox{\tiny (0.5)}$^{\star3}$ & \textbf{3} & \textbf{.9}\mbox{\tiny (0.4)}$^{\star3}$ & \textbf{3} & \textbf{.8}\mbox{\tiny (0.5)}$^{\star3}$ & \textbf{3} & \textbf{.9}\mbox{\tiny (0.3)}$^{\star3}$ & \textbf{3} & \textbf{.9}\mbox{\tiny (0.5)}$^{\star3}$ & \textbf{4} & \textbf{.0}\mbox{\tiny (0.4)}$^{\star3}$ & \textbf{4} & \textbf{.1}\mbox{\tiny (0.4)}$^{\star3}$ & 13 & /13\\
RecPM-AOS \hspace*{\fill} & 19 & \mbox{\tiny (2)} & 15 & \mbox{\tiny (2)} & 14 & \mbox{\tiny (1)} & 14 & \mbox{\tiny (1)} & 14 & \mbox{\tiny (0.9)} & 14 & \mbox{\tiny (0.9)} & 14 & \mbox{\tiny (0.8)} & 13 & /13\\
PM-AdapSS \hspace*{\fill} & 18 & \mbox{\tiny (2)} & 15 & \mbox{\tiny (1)} & 14 & \mbox{\tiny (0.9)} & 14 & \mbox{\tiny (0.6)} & 14 & \mbox{\tiny (0.4)} & 15 & \mbox{\tiny (0.4)} & 15 & \mbox{\tiny (0.2)} & 13 & /13\\
F-AUC-MAB \hspace*{\fill} & 23 & \mbox{\tiny (4)} & 23 & \mbox{\tiny (5)} & 23 & \mbox{\tiny (9)} & 26 & \mbox{\tiny (8)} & 29 & \mbox{\tiny (15)} & 39 & \mbox{\tiny (24)} & 53 & \mbox{\tiny (33)} & 13 & /13\\
Compass \hspace*{\fill} & 14 & \mbox{\tiny (1)} & 11 & \mbox{\tiny (0.8)} & 10 & \mbox{\tiny (0.7)} & 11 & \mbox{\tiny (0.5)} & 10 & \mbox{\tiny (0.5)} & 11 & \mbox{\tiny (0.4)} & 11 & \mbox{\tiny (0.1)} & 13 & /13\\
JaDE \hspace*{\fill} & 9 & .4\mbox{\tiny (0.8)} & 7 & .8\mbox{\tiny (0.6)} & 7 & .3\mbox{\tiny (0.8)} & 7 & .3\mbox{\tiny (0.9)} & 7 & .2\mbox{\tiny (0.9)} & 7 & .4\mbox{\tiny (0.4)} & 7 & .4\mbox{\tiny (0.6)} & 13 & /13\\

\end{tabularx}
\providecommand{\algAtables}{\StrLeft{U-AOS-FW}{\ntables}}
\providecommand{\algBtables}{\StrLeft{R-SHADE}{\ntables}}
\providecommand{\algCtables}{\StrLeft{RecPM-AOS}{\ntables}}
\providecommand{\algDtables}{\StrLeft{PM-AdapSS}{\ntables}}
\providecommand{\algEtables}{\StrLeft{F-AUC-MAB}{\ntables}}
\providecommand{\algFtables}{\StrLeft{Compass}{\ntables}}
\providecommand{\algGtables}{\StrLeft{JaDE}{\ntables}}

\begin{tabularx}{1.0\textwidth}{@{}c@{}|*{7}{@{}r@{}X@{}}|@{}r@{}@{}l@{}}
\hline

\textbf{f7} & \multicolumn{2}{@{}c@{}}{1351 \quad} & \multicolumn{2}{@{}c@{}}{4274 \quad} & \multicolumn{2}{@{}c@{}}{9503 \quad} & \multicolumn{2}{@{}c@{}}{16523 \quad} & \multicolumn{2}{@{}c@{}}{16524 \quad} & \multicolumn{2}{@{}c@{}}{16524 \quad} & \multicolumn{2}{@{}c@{}|}{16969} & 15 & /15\\\hline
U-AOS-FW \hspace*{\fill} & 5 & .2\mbox{\tiny (0.6)} & 3 & .7\mbox{\tiny (0.6)} & 2 & .3\mbox{\tiny (0.2)} & 1 & .9\mbox{\tiny (0.1)} & 1 & .9\mbox{\tiny (0.1)} & 1 & .9\mbox{\tiny (0.2)} & 1 & .9\mbox{\tiny (0.2)} & 13 & /13\\
R-SHADE \hspace*{\fill} & \textbf{2} & \textbf{.1}\mbox{\tiny (0.3)}$^{\star3}$ & 33 & \mbox{\tiny (19)} & 76 & \mbox{\tiny (87)} & \multicolumn{2}{@{\,}l@{\,}}{$\infty$} & \multicolumn{2}{@{\,}l@{\,}}{$\infty$} & \multicolumn{2}{@{\,}l@{\,}}{$\infty$} & \multicolumn{2}{@{\,}l@{\,}|}{$\infty$\,\textit{2e6}} & 0 & /13\\
RecPM-AOS \hspace*{\fill} & 6 & .5\mbox{\tiny (1.0)} & 4 & .6\mbox{\tiny (0.6)} & 3 & .0\mbox{\tiny (0.3)} & 2 & .5\mbox{\tiny (0.3)} & 2 & .5\mbox{\tiny (0.3)} & 2 & .5\mbox{\tiny (0.4)} & 2 & .6\mbox{\tiny (0.4)} & 13 & /13\\
PM-AdapSS \hspace*{\fill} & 6 & .7\mbox{\tiny (0.5)} & 43 & \mbox{\tiny (117)} & 20 & \mbox{\tiny (0.5)} & 12 & \mbox{\tiny (0.2)} & 12 & \mbox{\tiny (31)} & 12 & \mbox{\tiny (0.2)} & 12 & \mbox{\tiny (88)} & 12 & /13\\
F-AUC-MAB \hspace*{\fill} & 47 & \mbox{\tiny (29)} & 1025 & \mbox{\tiny (1075)} & \multicolumn{2}{@{\,}l@{\,}}{$\infty$} & \multicolumn{2}{@{\,}l@{\,}}{$\infty$} & \multicolumn{2}{@{\,}l@{\,}}{$\infty$} & \multicolumn{2}{@{\,}l@{\,}}{$\infty$} & \multicolumn{2}{@{\,}l@{\,}|}{$\infty$\,\textit{2e6}} & 0 & /13\\
Compass \hspace*{\fill} & 5 & .1\mbox{\tiny (0.6)} & \textbf{3} & \textbf{.5}\mbox{\tiny (0.7)} & \textbf{2} & \textbf{.2}\mbox{\tiny (0.4)} & \textbf{1} & \textbf{.8}\mbox{\tiny (0.2)} & \textbf{1} & \textbf{.8}\mbox{\tiny (0.2)} & \textbf{1} & \textbf{.8}\mbox{\tiny (0.2)} & \textbf{1} & \textbf{.9}\mbox{\tiny (0.2)} & 13 & /13\\
JaDE \hspace*{\fill} & 4 & .9\mbox{\tiny (1.0)} & 379 & \mbox{\tiny (761)} & 1265 & \mbox{\tiny (2157)} & 739 & \mbox{\tiny (726)} & 739 & \mbox{\tiny (1044)} & 739 & \mbox{\tiny (1044)} & 719 & \mbox{\tiny (368)} & 1 & /13\\

\end{tabularx}
\providecommand{\algAtables}{\StrLeft{U-AOS-FW}{\ntables}}
\providecommand{\algBtables}{\StrLeft{R-SHADE}{\ntables}}
\providecommand{\algCtables}{\StrLeft{RecPM-AOS}{\ntables}}
\providecommand{\algDtables}{\StrLeft{PM-AdapSS}{\ntables}}
\providecommand{\algEtables}{\StrLeft{F-AUC-MAB}{\ntables}}
\providecommand{\algFtables}{\StrLeft{Compass}{\ntables}}
\providecommand{\algGtables}{\StrLeft{JaDE}{\ntables}}

\begin{tabularx}{1.0\textwidth}{@{}c@{}|*{7}{@{}r@{}X@{}}|@{}r@{}@{}l@{}}
\hline

\textbf{f8} & \multicolumn{2}{@{}c@{}}{2039 \quad} & \multicolumn{2}{@{}c@{}}{3871 \quad} & \multicolumn{2}{@{}c@{}}{4040 \quad} & \multicolumn{2}{@{}c@{}}{4148 \quad} & \multicolumn{2}{@{}c@{}}{4219 \quad} & \multicolumn{2}{@{}c@{}}{4371 \quad} & \multicolumn{2}{@{}c@{}|}{4484} & 15 & /15\\\hline
U-AOS-FW \hspace*{\fill} & 22 & \mbox{\tiny (2)} & 22 & \mbox{\tiny (2)} & 23 & \mbox{\tiny (2)} & 24 & \mbox{\tiny (2)} & 24 & \mbox{\tiny (2)} & 25 & \mbox{\tiny (2)} & 27 & \mbox{\tiny (1)} & 13 & /13\\
R-SHADE \hspace*{\fill} & \textbf{10} & \textbf{}\mbox{\tiny (0.9)}$^{\star3}$ & \textbf{14} & \textbf{}\mbox{\tiny (8)} & \textbf{15} & \textbf{}\mbox{\tiny (7)} & \textbf{15} & \textbf{}\mbox{\tiny (7)} & \textbf{15} & \textbf{}\mbox{\tiny (7)} & \textbf{16} & \textbf{}\mbox{\tiny (3)} & \textbf{16} & \textbf{}\mbox{\tiny (7)} & 13 & /13\\
RecPM-AOS \hspace*{\fill} & 19 & \mbox{\tiny (4)} & 59 & \mbox{\tiny (259)} & 59 & \mbox{\tiny (1)} & 58 & \mbox{\tiny (121)} & 58 & \mbox{\tiny (1)} & 59 & \mbox{\tiny (1)} & 60 & \mbox{\tiny (1)} & 12 & /13\\
PM-AdapSS \hspace*{\fill} & 31 & \mbox{\tiny (5)} & 33 & \mbox{\tiny (2)} & 36 & \mbox{\tiny (2)} & 36 & \mbox{\tiny (1)} & 37 & \mbox{\tiny (2)} & 38 & \mbox{\tiny (2)} & 39 & \mbox{\tiny (2)} & 13 & /13\\
F-AUC-MAB \hspace*{\fill} & 461 & \mbox{\tiny (414)} & 3238 & \mbox{\tiny (9429)} & \multicolumn{2}{@{\,}l@{\,}}{$\infty$} & \multicolumn{2}{@{\,}l@{\,}}{$\infty$} & \multicolumn{2}{@{\,}l@{\,}}{$\infty$} & \multicolumn{2}{@{\,}l@{\,}}{$\infty$} & \multicolumn{2}{@{\,}l@{\,}|}{$\infty$\,\textit{2e6}} & 0 & /13\\
Compass \hspace*{\fill} & 20 & \mbox{\tiny (2)} & 19 & \mbox{\tiny (1)} & 20 & \mbox{\tiny (2)} & 21 & \mbox{\tiny (1)} & 22 & \mbox{\tiny (1)} & 23 & \mbox{\tiny (1)} & 24 & \mbox{\tiny (0.8)} & 13 & /13\\
JaDE \hspace*{\fill} & 18 & \mbox{\tiny (0.7)} & 15 & \mbox{\tiny (0.5)} & 16 & \mbox{\tiny (0.4)} & 17 & \mbox{\tiny (0.6)} & 17 & \mbox{\tiny (0.4)} & 17 & \mbox{\tiny (0.5)} & 18 & \mbox{\tiny (0.7)} & 13 & /13\\

\end{tabularx}
\providecommand{\algAtables}{\StrLeft{U-AOS-FW}{\ntables}}
\providecommand{\algBtables}{\StrLeft{R-SHADE}{\ntables}}
\providecommand{\algCtables}{\StrLeft{RecPM-AOS}{\ntables}}
\providecommand{\algDtables}{\StrLeft{PM-AdapSS}{\ntables}}
\providecommand{\algEtables}{\StrLeft{F-AUC-MAB}{\ntables}}
\providecommand{\algFtables}{\StrLeft{Compass}{\ntables}}
\providecommand{\algGtables}{\StrLeft{JaDE}{\ntables}}

\begin{tabularx}{1.0\textwidth}{@{}c@{}|*{7}{@{}r@{}X@{}}|@{}r@{}@{}l@{}}
\hline

\textbf{f9} & \multicolumn{2}{@{}c@{}}{1716 \quad} & \multicolumn{2}{@{}c@{}}{3102 \quad} & \multicolumn{2}{@{}c@{}}{3277 \quad} & \multicolumn{2}{@{}c@{}}{3379 \quad} & \multicolumn{2}{@{}c@{}}{3455 \quad} & \multicolumn{2}{@{}c@{}}{3594 \quad} & \multicolumn{2}{@{}c@{}|}{3727} & 15 & /15\\\hline
U-AOS-FW \hspace*{\fill} & 31 & \mbox{\tiny (1)} & 34 & \mbox{\tiny (3)} & 37 & \mbox{\tiny (3)} & 38 & \mbox{\tiny (3)} & 39 & \mbox{\tiny (3)} & 42 & \mbox{\tiny (3)} & 44 & \mbox{\tiny (3)} & 13 & /13\\
R-SHADE \hspace*{\fill} & \textbf{16} & \textbf{}\mbox{\tiny (4)}$^{\star3}$ & \textbf{20} & \textbf{}\mbox{\tiny (2)}$^{\star2}$ & \textbf{22} & \textbf{}\mbox{\tiny (13)}$^{\star2}$ & \textbf{23} & \textbf{}\mbox{\tiny (3)}$^{\star2}$ & \textbf{24} & \textbf{}\mbox{\tiny (2)}$^{\star2}$ & \textbf{24} & \textbf{}\mbox{\tiny (2)}$^{\star2}$ & \textbf{24} & \textbf{}\mbox{\tiny (2)}$^{\star2}$ & 13 & /13\\
RecPM-AOS \hspace*{\fill} & 25 & \mbox{\tiny (1)} & 25 & \mbox{\tiny (1)} & 28 & \mbox{\tiny (1)} & 30 & \mbox{\tiny (1)} & 31 & \mbox{\tiny (1)} & 35 & \mbox{\tiny (2)} & 39 & \mbox{\tiny (2)} & 13 & /13\\
PM-AdapSS \hspace*{\fill} & 44 & \mbox{\tiny (3)} & 51 & \mbox{\tiny (4)} & 55 & \mbox{\tiny (3)} & 56 & \mbox{\tiny (5)} & 57 & \mbox{\tiny (3)} & 58 & \mbox{\tiny (5)} & 60 & \mbox{\tiny (5)} & 13 & /13\\
F-AUC-MAB \hspace*{\fill} & 689 & \mbox{\tiny (775)} & 8342 & \mbox{\tiny (7253)} & \multicolumn{2}{@{\,}l@{\,}}{$\infty$} & \multicolumn{2}{@{\,}l@{\,}}{$\infty$} & \multicolumn{2}{@{\,}l@{\,}}{$\infty$} & \multicolumn{2}{@{\,}l@{\,}}{$\infty$} & \multicolumn{2}{@{\,}l@{\,}|}{$\infty$\,\textit{2e6}} & 0 & /13\\
Compass \hspace*{\fill} & 27 & \mbox{\tiny (2)} & 145 & \mbox{\tiny (323)} & 141 & \mbox{\tiny (154)} & 139 & \mbox{\tiny (3)} & 138 & \mbox{\tiny (290)} & 136 & \mbox{\tiny (280)} & 134 & \mbox{\tiny (2)} & 11 & /13\\
JaDE \hspace*{\fill} & 35 & \mbox{\tiny (3)} & 30 & \mbox{\tiny (3)} & 32 & \mbox{\tiny (2)} & 32 & \mbox{\tiny (2)} & 32 & \mbox{\tiny (1)} & 33 & \mbox{\tiny (2)} & 33 & \mbox{\tiny (3)} & 13 & /13\\

\end{tabularx}
\providecommand{\algAtables}{\StrLeft{U-AOS-FW}{\ntables}}
\providecommand{\algBtables}{\StrLeft{R-SHADE}{\ntables}}
\providecommand{\algCtables}{\StrLeft{RecPM-AOS}{\ntables}}
\providecommand{\algDtables}{\StrLeft{PM-AdapSS}{\ntables}}
\providecommand{\algEtables}{\StrLeft{F-AUC-MAB}{\ntables}}
\providecommand{\algFtables}{\StrLeft{Compass}{\ntables}}
\providecommand{\algGtables}{\StrLeft{JaDE}{\ntables}}

\begin{tabularx}{1.0\textwidth}{@{}c@{}|*{7}{@{}r@{}X@{}}|@{}r@{}@{}l@{}}
\hline

\textbf{f10} & \multicolumn{2}{@{}c@{}}{7413 \quad} & \multicolumn{2}{@{}c@{}}{8661 \quad} & \multicolumn{2}{@{}c@{}}{10735 \quad} & \multicolumn{2}{@{}c@{}}{13641 \quad} & \multicolumn{2}{@{}c@{}}{14920 \quad} & \multicolumn{2}{@{}c@{}}{17073 \quad} & \multicolumn{2}{@{}c@{}|}{17476} & 15 & /15\\\hline
U-AOS-FW \hspace*{\fill} & 6 & .4\mbox{\tiny (0.7)} & 7 & .1\mbox{\tiny (1)} & 6 & .8\mbox{\tiny (1)} & 6 & .4\mbox{\tiny (2)} & 6 & .5\mbox{\tiny (1)} & 7 & .0\mbox{\tiny (0.3)} & 8 & .0\mbox{\tiny (2)} & 13 & /13\\
R-SHADE \hspace*{\fill} & 16 & \mbox{\tiny (6)} & 22 & \mbox{\tiny (6)} & 25 & \mbox{\tiny (6)} & 26 & \mbox{\tiny (6)} & 28 & \mbox{\tiny (7)} & 32 & \mbox{\tiny (8)} & 40 & \mbox{\tiny (14)} & 9 & /13\\
RecPM-AOS \hspace*{\fill} & 8 & .0\mbox{\tiny (0.7)} & 8 & .5\mbox{\tiny (0.5)} & 8 & .3\mbox{\tiny (0.7)} & 7 & .5\mbox{\tiny (0.4)} & 7 & .9\mbox{\tiny (0.5)} & 8 & .6\mbox{\tiny (0.5)} & 10 & \mbox{\tiny (0.4)} & 13 & /13\\
PM-AdapSS \hspace*{\fill} & 5 & .6\mbox{\tiny (0.5)} & 6 & .3\mbox{\tiny (1)} & \textbf{6} & \textbf{.0}\mbox{\tiny (1)} & \textbf{5} & \textbf{.5}\mbox{\tiny (2)} & \textbf{5} & \textbf{.7}\mbox{\tiny (0.8)} & \textbf{6} & \textbf{.1}\mbox{\tiny (1)} & \textbf{7} & \textbf{.0}\mbox{\tiny (2)} & 13 & /13\\
F-AUC-MAB \hspace*{\fill} & \multicolumn{2}{@{\,}l@{\,}}{$\infty$} & \multicolumn{2}{@{\,}l@{\,}}{$\infty$} & \multicolumn{2}{@{\,}l@{\,}}{$\infty$} & \multicolumn{2}{@{\,}l@{\,}}{$\infty$} & \multicolumn{2}{@{\,}l@{\,}}{$\infty$} & \multicolumn{2}{@{\,}l@{\,}}{$\infty$} & \multicolumn{2}{@{\,}l@{\,}|}{$\infty$\,\textit{2e6}} & 0 & /13\\
Compass \hspace*{\fill} & \textbf{5} & \textbf{.5}\mbox{\tiny (0.9)} & \textbf{6} & \textbf{.3}\mbox{\tiny (2)} & 6 & .2\mbox{\tiny (1)} & 6 & .0\mbox{\tiny (1)} & 6 & .3\mbox{\tiny (1)} & 7 & .1\mbox{\tiny (3)} & 8 & .3\mbox{\tiny (3)} & 13 & /13\\
JaDE \hspace*{\fill} & 12 & \mbox{\tiny (4)} & 14 & \mbox{\tiny (7)} & 15 & \mbox{\tiny (4)} & 14 & \mbox{\tiny (4)} & 14 & \mbox{\tiny (3)} & 15 & \mbox{\tiny (3)} & 18 & \mbox{\tiny (5)} & 13 & /13\\

\end{tabularx}
\providecommand{\algAtables}{\StrLeft{U-AOS-FW}{\ntables}}
\providecommand{\algBtables}{\StrLeft{R-SHADE}{\ntables}}
\providecommand{\algCtables}{\StrLeft{RecPM-AOS}{\ntables}}
\providecommand{\algDtables}{\StrLeft{PM-AdapSS}{\ntables}}
\providecommand{\algEtables}{\StrLeft{F-AUC-MAB}{\ntables}}
\providecommand{\algFtables}{\StrLeft{Compass}{\ntables}}
\providecommand{\algGtables}{\StrLeft{JaDE}{\ntables}}

\begin{tabularx}{1.0\textwidth}{@{}c@{}|*{7}{@{}r@{}X@{}}|@{}r@{}@{}l@{}}
\hline

\textbf{f11} & \multicolumn{2}{@{}c@{}}{1002 \quad} & \multicolumn{2}{@{}c@{}}{2228 \quad} & \multicolumn{2}{@{}c@{}}{6278 \quad} & \multicolumn{2}{@{}c@{}}{8586 \quad} & \multicolumn{2}{@{}c@{}}{9762 \quad} & \multicolumn{2}{@{}c@{}}{12285 \quad} & \multicolumn{2}{@{}c@{}|}{14831} & 15 & /15\\\hline
U-AOS-FW \hspace*{\fill} & 15 & \mbox{\tiny (2)} & 11 & \mbox{\tiny (0.8)} & 5 & .7\mbox{\tiny (0.3)} & 5 & .4\mbox{\tiny (0.3)} & 5 & .7\mbox{\tiny (0.1)} & 6 & .2\mbox{\tiny (0.2)} & 6 & .5\mbox{\tiny (0.3)} & 13 & /13\\
R-SHADE \hspace*{\fill} & \textbf{7} & \textbf{.9}\mbox{\tiny (5)}$^{\star}$ & 13 & \mbox{\tiny (5)} & 8 & .0\mbox{\tiny (3)} & 8 & .2\mbox{\tiny (2)} & 10 & \mbox{\tiny (2)} & 11 & \mbox{\tiny (3)} & 12 & \mbox{\tiny (4)} & 13 & /13\\
RecPM-AOS \hspace*{\fill} & 23 & \mbox{\tiny (3)} & 16 & \mbox{\tiny (2)} & 8 & .0\mbox{\tiny (0.4)} & 7 & .4\mbox{\tiny (0.2)} & 7 & .9\mbox{\tiny (0.6)} & 8 & .6\mbox{\tiny (0.4)} & 9 & .0\mbox{\tiny (0.6)} & 13 & /13\\
PM-AdapSS \hspace*{\fill} & 15 & \mbox{\tiny (1)} & 10 & \mbox{\tiny (0.4)} & 5 & .2\mbox{\tiny (0.2)} & 4 & .9\mbox{\tiny (0.2)} & 5 & .2\mbox{\tiny (0.2)} & 5 & .6\mbox{\tiny (0.2)} & 5 & .8\mbox{\tiny (0.2)} & 13 & /13\\
F-AUC-MAB \hspace*{\fill} & 917 & \mbox{\tiny (159)} & 942 & \mbox{\tiny (459)} & 4071 & \mbox{\tiny (8920)} & \multicolumn{2}{@{\,}l@{\,}}{$\infty$} & \multicolumn{2}{@{\,}l@{\,}}{$\infty$} & \multicolumn{2}{@{\,}l@{\,}}{$\infty$} & \multicolumn{2}{@{\,}l@{\,}|}{$\infty$\,\textit{2e6}} & 0 & /13\\
Compass \hspace*{\fill} & 13 & \mbox{\tiny (2)} & \textbf{10} & \textbf{}\mbox{\tiny (0.9)} & \textbf{4} & \textbf{.8}\mbox{\tiny (0.3)}$^{\star}$ & \textbf{4} & \textbf{.4}\mbox{\tiny (0.3)}$^{\star2}$ & \textbf{4} & \textbf{.8}\mbox{\tiny (0.4)}$^{\star}$ & \textbf{5} & \textbf{.1}\mbox{\tiny (0.6)}$^{\star}$ & \textbf{5} & \textbf{.4}\mbox{\tiny (0.5)}$^{\star2}$ & 13 & /13\\
JaDE \hspace*{\fill} & 105 & \mbox{\tiny (501)} & 52 & \mbox{\tiny (117)} & 21 & \mbox{\tiny (2)} & 17 & \mbox{\tiny (31)} & 17 & \mbox{\tiny (28)} & 16 & \mbox{\tiny (41)} & 16 & \mbox{\tiny (3)} & 12 & /13\\

\end{tabularx}
\providecommand{\algAtables}{\StrLeft{U-AOS-FW}{\ntables}}
\providecommand{\algBtables}{\StrLeft{R-SHADE}{\ntables}}
\providecommand{\algCtables}{\StrLeft{RecPM-AOS}{\ntables}}
\providecommand{\algDtables}{\StrLeft{PM-AdapSS}{\ntables}}
\providecommand{\algEtables}{\StrLeft{F-AUC-MAB}{\ntables}}
\providecommand{\algFtables}{\StrLeft{Compass}{\ntables}}
\providecommand{\algGtables}{\StrLeft{JaDE}{\ntables}}

\begin{tabularx}{1.0\textwidth}{@{}c@{}|*{7}{@{}r@{}X@{}}|@{}r@{}@{}l@{}}
\hline

\textbf{f12} & \multicolumn{2}{@{}c@{}}{1042 \quad} & \multicolumn{2}{@{}c@{}}{1938 \quad} & \multicolumn{2}{@{}c@{}}{2740 \quad} & \multicolumn{2}{@{}c@{}}{3156 \quad} & \multicolumn{2}{@{}c@{}}{4140 \quad} & \multicolumn{2}{@{}c@{}}{12407 \quad} & \multicolumn{2}{@{}c@{}|}{13827} & 15 & /15\\\hline
U-AOS-FW \hspace*{\fill} & 29 & \mbox{\tiny (2)} & 21 & \mbox{\tiny (4)} & 25 & \mbox{\tiny (12)} & 32 & \mbox{\tiny (13)} & 33 & \mbox{\tiny (8)} & 17 & \mbox{\tiny (4)} & 19 & \mbox{\tiny (3)} & 13 & /13\\
R-SHADE \hspace*{\fill} & \textbf{8} & \textbf{.5}\mbox{\tiny (1)}$^{\star3}$ & \textbf{17} & \textbf{}\mbox{\tiny (11)} & \textbf{22} & \textbf{}\mbox{\tiny (13)} & \textbf{25} & \textbf{}\mbox{\tiny (12)} & \textbf{24} & \textbf{}\mbox{\tiny (9)} & \textbf{12} & \textbf{}\mbox{\tiny (4)} & \textbf{13} & \textbf{}\mbox{\tiny (4)} & 13 & /13\\
RecPM-AOS \hspace*{\fill} & 37 & \mbox{\tiny (5)} & 31 & \mbox{\tiny (19)} & 31 & \mbox{\tiny (20)} & 37 & \mbox{\tiny (15)} & 37 & \mbox{\tiny (8)} & 18 & \mbox{\tiny (3)} & 20 & \mbox{\tiny (3)} & 13 & /13\\
PM-AdapSS \hspace*{\fill} & 44 & \mbox{\tiny (18)} & 40 & \mbox{\tiny (10)} & 42 & \mbox{\tiny (31)} & 47 & \mbox{\tiny (29)} & 44 & \mbox{\tiny (13)} & 20 & \mbox{\tiny (2)} & 22 & \mbox{\tiny (3)} & 13 & /13\\
F-AUC-MAB \hspace*{\fill} & 1091 & \mbox{\tiny (2590)} & 1500 & \mbox{\tiny (653)} & 4257 & \mbox{\tiny (6447)} & \multicolumn{2}{@{\,}l@{\,}}{$\infty$} & \multicolumn{2}{@{\,}l@{\,}}{$\infty$} & \multicolumn{2}{@{\,}l@{\,}}{$\infty$} & \multicolumn{2}{@{\,}l@{\,}|}{$\infty$\,\textit{2e6}} & 0 & /13\\
Compass \hspace*{\fill} & 29 & \mbox{\tiny (2)} & 31 & \mbox{\tiny (21)} & 38 & \mbox{\tiny (19)} & 41 & \mbox{\tiny (19)} & 38 & \mbox{\tiny (12)} & 18 & \mbox{\tiny (8)} & 19 & \mbox{\tiny (6)} & 13 & /13\\
JaDE \hspace*{\fill} & 19 & \mbox{\tiny (2)} & 20 & \mbox{\tiny (14)} & 28 & \mbox{\tiny (13)} & 31 & \mbox{\tiny (10)} & 29 & \mbox{\tiny (8)} & 13 & \mbox{\tiny (3)} & 14 & \mbox{\tiny (3)} & 13 & /13\\

\end{tabularx}
\end{minipage}}
\end{table*}
\begin{table*}[tp]\tiny
\caption{Average runtime \aRT}\label{tab:art-ufw2}
\mbox{
\begin{minipage}[t]{0.99\textwidth}\tiny
\centering
\providecommand{\algAtables}{\StrLeft{U-AOS-FW}{\ntables}}
\providecommand{\algBtables}{\StrLeft{R-SHADE}{\ntables}}
\providecommand{\algCtables}{\StrLeft{RecPM-AOS}{\ntables}}
\providecommand{\algDtables}{\StrLeft{PM-AdapSS}{\ntables}}
\providecommand{\algEtables}{\StrLeft{F-AUC-MAB}{\ntables}}
\providecommand{\algFtables}{\StrLeft{Compass}{\ntables}}
\providecommand{\algGtables}{\StrLeft{JaDE}{\ntables}}

\begin{tabularx}{1.0\textwidth}{@{}c@{}|*{7}{@{}r@{}X@{}}|@{}r@{}@{}l@{}}
$\Delta f_\mathrm{opt}$ & \multicolumn{2}{@{\,}l@{\,}}{1e1} & \multicolumn{2}{@{\,}l@{\,}}{1e0} & \multicolumn{2}{@{\,}l@{\,}}{1e-1} & \multicolumn{2}{@{\,}l@{\,}}{1e-2} & \multicolumn{2}{@{\,}l@{\,}}{1e-3} & \multicolumn{2}{@{\,}l@{\,}}{1e-5} & \multicolumn{2}{@{\,}l@{\,}}{1e-7} & \multicolumn{2}{|@{}l@{}}{\#succ}\\\hline

\textbf{f13} & \multicolumn{2}{@{}c@{}}{652 \quad} & \multicolumn{2}{@{}c@{}}{2021 \quad} & \multicolumn{2}{@{}c@{}}{2751 \quad} & \multicolumn{2}{@{}c@{}}{3507 \quad} & \multicolumn{2}{@{}c@{}}{18749 \quad} & \multicolumn{2}{@{}c@{}}{24455 \quad} & \multicolumn{2}{@{}c@{}|}{30201} & 15 & /15\\\hline
U-AOS-FW \hspace*{\fill} & 25 & \mbox{\tiny (3)} & 13 & \mbox{\tiny (2)} & 15 & \mbox{\tiny (1)} & 17 & \mbox{\tiny (2)} & 4 & .2\mbox{\tiny (0.5)} & 5 & .0\mbox{\tiny (0.3)} & 5 & .6\mbox{\tiny (0.5)} & 13 & /13\\
R-SHADE \hspace*{\fill} & \textbf{12} & \textbf{}\mbox{\tiny (7)} & \textbf{8} & \textbf{.4}\mbox{\tiny (3)} & \textbf{11} & \textbf{}\mbox{\tiny (2)} & \textbf{12} & \textbf{}\mbox{\tiny (3)} & \textbf{3} & \textbf{.5}\mbox{\tiny (1)} & 5 & .2\mbox{\tiny (1)} & 27 & \mbox{\tiny (22)} & 0 & /13\\
RecPM-AOS \hspace*{\fill} & 31 & \mbox{\tiny (4)} & 17 & \mbox{\tiny (1)} & 19 & \mbox{\tiny (2)} & 22 & \mbox{\tiny (2)} & 5 & .5\mbox{\tiny (0.2)} & 6 & .5\mbox{\tiny (0.3)} & 7 & .2\mbox{\tiny (0.5)} & 13 & /13\\
PM-AdapSS \hspace*{\fill} & 32 & \mbox{\tiny (1)} & 16 & \mbox{\tiny (1)} & 18 & \mbox{\tiny (1)} & 19 & \mbox{\tiny (0.6)} & 4 & .4\mbox{\tiny (0.2)} & 4 & .9\mbox{\tiny (0.1)} & \textbf{5} & \textbf{.2}\mbox{\tiny (0.1)} & 13 & /13\\
F-AUC-MAB \hspace*{\fill} & 49 & \mbox{\tiny (8)} & 504 & \mbox{\tiny (1134)} & 2877 & \mbox{\tiny (2212)} & \multicolumn{2}{@{\,}l@{\,}}{$\infty$} & \multicolumn{2}{@{\,}l@{\,}}{$\infty$} & \multicolumn{2}{@{\,}l@{\,}}{$\infty$} & \multicolumn{2}{@{\,}l@{\,}|}{$\infty$\,\textit{2e6}} & 0 & /13\\
Compass \hspace*{\fill} & 24 & \mbox{\tiny (2)} & 13 & \mbox{\tiny (0.6)} & 14 & \mbox{\tiny (0.6)} & 15 & \mbox{\tiny (0.6)} & 3 & .8\mbox{\tiny (0.4)} & \textbf{4} & \textbf{.5}\mbox{\tiny (0.5)} & 5 & .7\mbox{\tiny (1)} & 13 & /13\\
JaDE \hspace*{\fill} & 17 & \mbox{\tiny (1)} & 14 & \mbox{\tiny (6)} & 15 & \mbox{\tiny (5)} & 14 & \mbox{\tiny (4)} & 3 & .7\mbox{\tiny (0.3)} & 4 & .8\mbox{\tiny (0.9)} & 8 & .7\mbox{\tiny (2)} & 13 & /13\\

\end{tabularx} 
\providecommand{\algAtables}{\StrLeft{U-AOS-FW}{\ntables}}
\providecommand{\algBtables}{\StrLeft{R-SHADE}{\ntables}}
\providecommand{\algCtables}{\StrLeft{RecPM-AOS}{\ntables}}
\providecommand{\algDtables}{\StrLeft{PM-AdapSS}{\ntables}}
\providecommand{\algEtables}{\StrLeft{F-AUC-MAB}{\ntables}}
\providecommand{\algFtables}{\StrLeft{Compass}{\ntables}}
\providecommand{\algGtables}{\StrLeft{JaDE}{\ntables}}

\begin{tabularx}{1.0\textwidth}{@{}c@{}|*{7}{@{}r@{}X@{}}|@{}r@{}@{}l@{}}
\hline

\textbf{f14} & \multicolumn{2}{@{}c@{}}{75 \quad} & \multicolumn{2}{@{}c@{}}{239 \quad} & \multicolumn{2}{@{}c@{}}{304 \quad} & \multicolumn{2}{@{}c@{}}{451 \quad} & \multicolumn{2}{@{}c@{}}{932 \quad} & \multicolumn{2}{@{}c@{}}{1648 \quad} & \multicolumn{2}{@{}c@{}|}{15661} & 15 & /15\\\hline
U-AOS-FW \hspace*{\fill} & 29 & \mbox{\tiny (10)} & 26 & \mbox{\tiny (1)} & 35 & \mbox{\tiny (0.9)} & 35 & \mbox{\tiny (1)} & 26 & \mbox{\tiny (0.7)} & 29 & \mbox{\tiny (2)} & 5 & .3\mbox{\tiny (0.9)} & 12 & /13\\
R-SHADE \hspace*{\fill} & \textbf{8} & \textbf{.1}\mbox{\tiny (2)}$^{\star2}$ & \textbf{10} & \textbf{}\mbox{\tiny (2)}$^{\star3}$ & \textbf{13} & \textbf{}\mbox{\tiny (2)}$^{\star3}$ & \textbf{13} & \textbf{}\mbox{\tiny (2)}$^{\star3}$ & \textbf{11} & \textbf{}\mbox{\tiny (2)}$^{\star3}$ & 62 & \mbox{\tiny (18)} & 1606 & \mbox{\tiny (2171)} & 0 & /13\\
RecPM-AOS \hspace*{\fill} & 32 & \mbox{\tiny (4)} & 31 & \mbox{\tiny (4)} & 40 & \mbox{\tiny (4)} & 41 & \mbox{\tiny (3)} & 31 & \mbox{\tiny (1)} & 39 & \mbox{\tiny (2)} & 6 & .5\mbox{\tiny (0.5)} & 13 & /13\\
PM-AdapSS \hspace*{\fill} & 33 & \mbox{\tiny (9)} & 33 & \mbox{\tiny (3)} & 45 & \mbox{\tiny (1.0)} & 44 & \mbox{\tiny (2)} & 30 & \mbox{\tiny (2)} & 31 & \mbox{\tiny (1)} & \textbf{4} & \textbf{.9}\mbox{\tiny (0.2)} & 13 & /13\\
F-AUC-MAB \hspace*{\fill} & 37 & \mbox{\tiny (9)} & 33 & \mbox{\tiny (3)} & 40 & \mbox{\tiny (5)} & 91 & \mbox{\tiny (26)} & \multicolumn{2}{@{\,}l@{\,}}{2.6e4\mbox{\tiny (5e4)}} & \multicolumn{2}{@{\,}l@{\,}}{$\infty$} & \multicolumn{2}{@{\,}l@{\,}|}{$\infty$\,\textit{2e6}} & 0 & /13\\
Compass \hspace*{\fill} & 23 & \mbox{\tiny (5)} & 24 & \mbox{\tiny (2)} & 32 & \mbox{\tiny (2)} & 32 & \mbox{\tiny (1)} & 23 & \mbox{\tiny (1)} & \textbf{26} & \textbf{}\mbox{\tiny (1.0)} & 7 & .4\mbox{\tiny (0.3)} & 11 & /13\\
JaDE \hspace*{\fill} & 18 & \mbox{\tiny (8)} & 18 & \mbox{\tiny (2)} & 23 & \mbox{\tiny (2)} & 25 & \mbox{\tiny (2)} & 20 & \mbox{\tiny (0.9)} & 36 & \mbox{\tiny (26)} & 68 & \mbox{\tiny (88)} & 4 & /13\\

\end{tabularx}
\providecommand{\algAtables}{\StrLeft{U-AOS-FW}{\ntables}}
\providecommand{\algBtables}{\StrLeft{R-SHADE}{\ntables}}
\providecommand{\algCtables}{\StrLeft{RecPM-AOS}{\ntables}}
\providecommand{\algDtables}{\StrLeft{PM-AdapSS}{\ntables}}
\providecommand{\algEtables}{\StrLeft{F-AUC-MAB}{\ntables}}
\providecommand{\algFtables}{\StrLeft{Compass}{\ntables}}
\providecommand{\algGtables}{\StrLeft{JaDE}{\ntables}}

\begin{tabularx}{1.0\textwidth}{@{}c@{}|*{7}{@{}r@{}X@{}}|@{}r@{}@{}l@{}}
\hline

\textbf{f15} & 30378 & \mbox{} & 1.5e5 & \mbox{} & 3.1e5 & \mbox{} & 3.2e5 & \mbox{} & 3.2e5 & \mbox{} & 4.5e5 & \mbox{} & 4.6e5 & \mbox{} & 15 & /15\\\hline
U-AOS-FW \hspace*{\fill} & $\infty$ & \mbox{} & \textbf{$\infty$} & \mbox{} & \textbf{$\infty$} & \mbox{} & \textbf{$\infty$} & \mbox{} & \textbf{$\infty$} & \mbox{} & \textbf{$\infty$} & \mbox{} & \multicolumn{2}{@{\,}l@{\,}|}{$\infty$\,\textit{2e6}} & 0 & /13\\
R-SHADE \hspace*{\fill} & 56 & \mbox{\tiny (35)} & \textbf{$\infty$} & \mbox{} & \textbf{$\infty$} & \mbox{} & \textbf{$\infty$} & \mbox{} & \textbf{$\infty$} & \mbox{} & \textbf{$\infty$} & \mbox{} & \multicolumn{2}{@{\,}l@{\,}|}{$\infty$\,\textit{2e6}} & 0 & /13\\
RecPM-AOS \hspace*{\fill} & $\infty$ & \mbox{} & \textbf{$\infty$} & \mbox{} & \textbf{$\infty$} & \mbox{} & \textbf{$\infty$} & \mbox{} & \textbf{$\infty$} & \mbox{} & \textbf{$\infty$} & \mbox{} & \multicolumn{2}{@{\,}l@{\,}|}{$\infty$\,\textit{2e6}} & 0 & /13\\
PM-AdapSS \hspace*{\fill} & $\infty$ & \mbox{} & \textbf{$\infty$} & \mbox{} & \textbf{$\infty$} & \mbox{} & \textbf{$\infty$} & \mbox{} & \textbf{$\infty$} & \mbox{} & \textbf{$\infty$} & \mbox{} & \multicolumn{2}{@{\,}l@{\,}|}{$\infty$\,\textit{2e6}} & 0 & /13\\
F-AUC-MAB \hspace*{\fill} & $\infty$ & \mbox{} & \textbf{$\infty$} & \mbox{} & \textbf{$\infty$} & \mbox{} & \textbf{$\infty$} & \mbox{} & \textbf{$\infty$} & \mbox{} & \textbf{$\infty$} & \mbox{} & \multicolumn{2}{@{\,}l@{\,}|}{$\infty$\,\textit{2e6}} & 0 & /13\\
Compass \hspace*{\fill} & $\infty$ & \mbox{} & \textbf{$\infty$} & \mbox{} & \textbf{$\infty$} & \mbox{} & \textbf{$\infty$} & \mbox{} & \textbf{$\infty$} & \mbox{} & \textbf{$\infty$} & \mbox{} & \multicolumn{2}{@{\,}l@{\,}|}{$\infty$\,\textit{2e6}} & 0 & /13\\
JaDE \hspace*{\fill} & \textbf{36} & \textbf{}\mbox{\tiny (22)} & \textbf{$\infty$} & \mbox{} & \textbf{$\infty$} & \mbox{} & \textbf{$\infty$} & \mbox{} & \textbf{$\infty$} & \mbox{} & \textbf{$\infty$} & \mbox{} & \multicolumn{2}{@{\,}l@{\,}|}{$\infty$\,\textit{1e6}} & 0 & /13\\

\end{tabularx}
\providecommand{\algAtables}{\StrLeft{U-AOS-FW}{\ntables}}
\providecommand{\algBtables}{\StrLeft{R-SHADE}{\ntables}}
\providecommand{\algCtables}{\StrLeft{RecPM-AOS}{\ntables}}
\providecommand{\algDtables}{\StrLeft{PM-AdapSS}{\ntables}}
\providecommand{\algEtables}{\StrLeft{F-AUC-MAB}{\ntables}}
\providecommand{\algFtables}{\StrLeft{Compass}{\ntables}}
\providecommand{\algGtables}{\StrLeft{JaDE}{\ntables}}

\begin{tabularx}{1.0\textwidth}{@{}c@{}|*{7}{@{}r@{}X@{}}|@{}r@{}@{}l@{}}
\hline

\textbf{f16} & 1384 & \mbox{} & 27265 & \mbox{} & 77015 & \mbox{} & 1.4e5 & \mbox{} & 1.9e5 & \mbox{} & 2.0e5 & \mbox{} & 2.2e5 & \mbox{} & 15 & /15\\\hline
U-AOS-FW \hspace*{\fill} & $\infty$ & \mbox{} & $\infty$ & \mbox{} & \textbf{$\infty$} & \mbox{} & \textbf{$\infty$} & \mbox{} & \textbf{$\infty$} & \mbox{} & \textbf{$\infty$} & \mbox{} & \multicolumn{2}{@{\,}l@{\,}|}{$\infty$\,\textit{2e6}} & 0 & /13\\
R-SHADE \hspace*{\fill} & 27 & \mbox{\tiny (22)} & \textbf{280} & \textbf{}\mbox{\tiny (290)} & \textbf{$\infty$} & \mbox{} & \textbf{$\infty$} & \mbox{} & \textbf{$\infty$} & \mbox{} & \textbf{$\infty$} & \mbox{} & \multicolumn{2}{@{\,}l@{\,}|}{$\infty$\,\textit{2e6}} & 0 & /13\\
RecPM-AOS \hspace*{\fill} & $\infty$ & \mbox{} & $\infty$ & \mbox{} & \textbf{$\infty$} & \mbox{} & \textbf{$\infty$} & \mbox{} & \textbf{$\infty$} & \mbox{} & \textbf{$\infty$} & \mbox{} & \multicolumn{2}{@{\,}l@{\,}|}{$\infty$\,\textit{2e6}} & 0 & /13\\
PM-AdapSS \hspace*{\fill} & $\infty$ & \mbox{} & $\infty$ & \mbox{} & \textbf{$\infty$} & \mbox{} & \textbf{$\infty$} & \mbox{} & \textbf{$\infty$} & \mbox{} & \textbf{$\infty$} & \mbox{} & \multicolumn{2}{@{\,}l@{\,}|}{$\infty$\,\textit{2e6}} & 0 & /13\\
F-AUC-MAB \hspace*{\fill} & 944 & \mbox{\tiny (431)} & $\infty$ & \mbox{} & \textbf{$\infty$} & \mbox{} & \textbf{$\infty$} & \mbox{} & \textbf{$\infty$} & \mbox{} & \textbf{$\infty$} & \mbox{} & \multicolumn{2}{@{\,}l@{\,}|}{$\infty$\,\textit{2e6}} & 0 & /13\\
Compass \hspace*{\fill} & $\infty$ & \mbox{} & $\infty$ & \mbox{} & \textbf{$\infty$} & \mbox{} & \textbf{$\infty$} & \mbox{} & \textbf{$\infty$} & \mbox{} & \textbf{$\infty$} & \mbox{} & \multicolumn{2}{@{\,}l@{\,}|}{$\infty$\,\textit{2e6}} & 0 & /13\\
JaDE \hspace*{\fill} & \textbf{24} & \textbf{}\mbox{\tiny (6)} & $\infty$ & \mbox{} & \textbf{$\infty$} & \mbox{} & \textbf{$\infty$} & \mbox{} & \textbf{$\infty$} & \mbox{} & \textbf{$\infty$} & \mbox{} & \multicolumn{2}{@{\,}l@{\,}|}{$\infty$\,\textit{1e6}} & 0 & /13\\

\end{tabularx}
\providecommand{\algAtables}{\StrLeft{U-AOS-FW}{\ntables}}
\providecommand{\algBtables}{\StrLeft{R-SHADE}{\ntables}}
\providecommand{\algCtables}{\StrLeft{RecPM-AOS}{\ntables}}
\providecommand{\algDtables}{\StrLeft{PM-AdapSS}{\ntables}}
\providecommand{\algEtables}{\StrLeft{F-AUC-MAB}{\ntables}}
\providecommand{\algFtables}{\StrLeft{Compass}{\ntables}}
\providecommand{\algGtables}{\StrLeft{JaDE}{\ntables}}

\begin{tabularx}{1.0\textwidth}{@{}c@{}|*{7}{@{}r@{}X@{}}|@{}r@{}@{}l@{}}
\hline

\textbf{f17} & \multicolumn{2}{@{}c@{}}{63 \quad} & \multicolumn{2}{@{}c@{}}{1030 \quad} & \multicolumn{2}{@{}c@{}}{4005 \quad} & \multicolumn{2}{@{}c@{}}{12242 \quad} & \multicolumn{2}{@{}c@{}}{30677 \quad} & \multicolumn{2}{@{}c@{}}{56288 \quad} & \multicolumn{2}{@{}c@{}|}{80472} & 15 & /15\\\hline
U-AOS-FW \hspace*{\fill} & 14 & \mbox{\tiny (4)} & 9 & .4\mbox{\tiny (0.7)} & 5 & .2\mbox{\tiny (0.4)} & 2 & .8\mbox{\tiny (0.2)} & 1 & .7\mbox{\tiny (0.3)} & 1 & .7\mbox{\tiny (0.1)} & 3 & .6\mbox{\tiny (6)} & 12 & /13\\
R-SHADE \hspace*{\fill} & \textbf{4} & \textbf{.0}\mbox{\tiny (2)} & \textbf{3} & \textbf{.4}\mbox{\tiny (0.5)}$^{\star3}$ & \textbf{4} & \textbf{.1}\mbox{\tiny (12)}$^{\star}$ & 14 & \mbox{\tiny (27)} & 51 & \mbox{\tiny (57)} & \multicolumn{2}{@{\,}l@{\,}}{$\infty$} & \multicolumn{2}{@{\,}l@{\,}|}{$\infty$\,\textit{2e6}} & 0 & /13\\
RecPM-AOS \hspace*{\fill} & 16 & \mbox{\tiny (11)} & 11 & \mbox{\tiny (1)} & 6 & .9\mbox{\tiny (0.7)} & 3 & .7\mbox{\tiny (0.3)} & 2 & .1\mbox{\tiny (0.2)} & 2 & .2\mbox{\tiny (0.1)} & 2 & .0\mbox{\tiny (0.3)} & 12 & /13\\
PM-AdapSS \hspace*{\fill} & 16 & \mbox{\tiny (9)} & 11 & \mbox{\tiny (2)} & 6 & .5\mbox{\tiny (0.4)} & 3 & .5\mbox{\tiny (0.3)} & 2 & .0\mbox{\tiny (0.1)} & 1 & .9\mbox{\tiny (0.1)} & 6 & .2\mbox{\tiny (12)} & 11 & /13\\
F-AUC-MAB \hspace*{\fill} & 13 & \mbox{\tiny (5)} & 26 & \mbox{\tiny (14)} & 66 & \mbox{\tiny (124)} & 182 & \mbox{\tiny (544)} & 788 & \mbox{\tiny (962)} & \multicolumn{2}{@{\,}l@{\,}}{$\infty$} & \multicolumn{2}{@{\,}l@{\,}|}{$\infty$\,\textit{2e6}} & 0 & /13\\
Compass \hspace*{\fill} & 15 & \mbox{\tiny (12)} & 9 & .1\mbox{\tiny (1)} & 5 & .1\mbox{\tiny (0.6)} & 2 & .8\mbox{\tiny (0.3)} & \textbf{1} & \textbf{.6}\mbox{\tiny (0.1)} & \textbf{1} & \textbf{.6}\mbox{\tiny (0.1)} & \textbf{1} & \textbf{.4}\mbox{\tiny (0.1)} & 12 & /13\\
JaDE \hspace*{\fill} & 8 & .0\mbox{\tiny (6)} & 7 & .3\mbox{\tiny (1)} & 4 & .4\mbox{\tiny (0.7)} & \textbf{2} & \textbf{.5}\mbox{\tiny (0.4)} & 1 & .7\mbox{\tiny (2)} & 8 & .1\mbox{\tiny (8)} & 19 & \mbox{\tiny (20)} & 2 & /13\\

\end{tabularx}
\providecommand{\algAtables}{\StrLeft{U-AOS-FW}{\ntables}}
\providecommand{\algBtables}{\StrLeft{R-SHADE}{\ntables}}
\providecommand{\algCtables}{\StrLeft{RecPM-AOS}{\ntables}}
\providecommand{\algDtables}{\StrLeft{PM-AdapSS}{\ntables}}
\providecommand{\algEtables}{\StrLeft{F-AUC-MAB}{\ntables}}
\providecommand{\algFtables}{\StrLeft{Compass}{\ntables}}
\providecommand{\algGtables}{\StrLeft{JaDE}{\ntables}}

\begin{tabularx}{1.0\textwidth}{@{}c@{}|*{7}{@{}r@{}X@{}}|@{}r@{}@{}l@{}}
\hline

\textbf{f18} & \multicolumn{2}{@{}c@{}}{621 \quad} & \multicolumn{2}{@{}c@{}}{3972 \quad} & \multicolumn{2}{@{}c@{}}{19561 \quad} & \multicolumn{2}{@{}c@{}}{28555 \quad} & \multicolumn{2}{@{}c@{}}{67569 \quad} & \multicolumn{2}{@{}c@{}}{1.3e5 \quad} & \multicolumn{2}{@{}c@{}|}{1.5e5} & 15 & /15\\\hline
U-AOS-FW \hspace*{\fill} & 8 & .7\mbox{\tiny (2)} & 4 & .6\mbox{\tiny (0.6)} & 1 & .9\mbox{\tiny (0.4)} & 2 & .1\mbox{\tiny (0.3)} & \textbf{1} & \textbf{.3}\mbox{\tiny (0.1)} & 3 & .9\mbox{\tiny (8)} & 3 & .8\mbox{\tiny (3)} & 11 & /13\\
R-SHADE \hspace*{\fill} & \textbf{3} & \textbf{.1}\mbox{\tiny (0.7)}$^{\star3}$ & \textbf{2} & \textbf{.1}\mbox{\tiny (1.0)}$^{\star3}$ & 34 & \mbox{\tiny (32)} & 910 & \mbox{\tiny (788)} & \multicolumn{2}{@{\,}l@{\,}}{$\infty$} & \multicolumn{2}{@{\,}l@{\,}}{$\infty$} & \multicolumn{2}{@{\,}l@{\,}|}{$\infty$\,\textit{2e6}} & 0 & /13\\
RecPM-AOS \hspace*{\fill} & 11 & \mbox{\tiny (2)} & 5 & .9\mbox{\tiny (0.6)} & 2 & .6\mbox{\tiny (0.2)} & 3 & .0\mbox{\tiny (0.5)} & 1 & .8\mbox{\tiny (0.2)} & \textbf{1} & \textbf{.7}\mbox{\tiny (0.2)} & \textbf{1} & \textbf{.9}\mbox{\tiny (0.2)} & 12 & /13\\
PM-AdapSS \hspace*{\fill} & 10 & \mbox{\tiny (0.8)} & 5 & .1\mbox{\tiny (0.6)} & 2 & .2\mbox{\tiny (0.3)} & 2 & .5\mbox{\tiny (1)} & 6 & .7\mbox{\tiny (15)} & 4 & .0\mbox{\tiny (0.1)} & 10 & \mbox{\tiny (14)} & 8 & /13\\
F-AUC-MAB \hspace*{\fill} & 20 & \mbox{\tiny (6)} & 188 & \mbox{\tiny (54)} & \multicolumn{2}{@{\,}l@{\,}}{$\infty$} & \multicolumn{2}{@{\,}l@{\,}}{$\infty$} & \multicolumn{2}{@{\,}l@{\,}}{$\infty$} & \multicolumn{2}{@{\,}l@{\,}}{$\infty$} & \multicolumn{2}{@{\,}l@{\,}|}{$\infty$\,\textit{2e6}} & 0 & /13\\
Compass \hspace*{\fill} & 8 & .5\mbox{\tiny (1)} & 4 & .5\mbox{\tiny (0.5)} & 1 & .8\mbox{\tiny (0.2)} & \textbf{2} & \textbf{.1}\mbox{\tiny (0.2)} & 3 & .7\mbox{\tiny (7)} & 3 & .9\mbox{\tiny (4)} & 5 & .3\mbox{\tiny (7)} & 10 & /13\\
JaDE \hspace*{\fill} & 7 & .3\mbox{\tiny (1)} & 4 & .5\mbox{\tiny (1)} & \textbf{1} & \textbf{.6}\mbox{\tiny (0.4)} & 9 & .4\mbox{\tiny (13)} & 20 & \mbox{\tiny (21)} & \multicolumn{2}{@{\,}l@{\,}}{$\infty$} & \multicolumn{2}{@{\,}l@{\,}|}{$\infty$\,\textit{1e6}} & 0 & /13\\

\end{tabularx}
\providecommand{\algAtables}{\StrLeft{U-AOS-FW}{\ntables}}
\providecommand{\algBtables}{\StrLeft{R-SHADE}{\ntables}}
\providecommand{\algCtables}{\StrLeft{RecPM-AOS}{\ntables}}
\providecommand{\algDtables}{\StrLeft{PM-AdapSS}{\ntables}}
\providecommand{\algEtables}{\StrLeft{F-AUC-MAB}{\ntables}}
\providecommand{\algFtables}{\StrLeft{Compass}{\ntables}}
\providecommand{\algGtables}{\StrLeft{JaDE}{\ntables}}

\begin{tabularx}{1.0\textwidth}{@{}c@{}|*{7}{@{}r@{}X@{}}|@{}r@{}@{}l@{}}
\hline

\textbf{f19} & 1 & \mbox{} & 1 & \mbox{} & 3.4e5 & \mbox{} & 4.7e6 & \mbox{} & 6.2e6 & \mbox{} & 6.7e6 & \mbox{} & 6.7e6 & \mbox{} & 15 & /15\\\hline
U-AOS-FW \hspace*{\fill} & \textbf{1} & \textbf{}\mbox{\tiny (0)} & \textbf{1} & \textbf{}\mbox{\tiny (0)} & \textbf{$\infty$} & \mbox{} & \textbf{$\infty$} & \mbox{} & \textbf{$\infty$} & \mbox{} & \textbf{$\infty$} & \mbox{} & \multicolumn{2}{@{\,}l@{\,}|}{$\infty$\,\textit{2e6}} & 0 & /13\\
R-SHADE \hspace*{\fill} & 344 & \mbox{\tiny (84)} & \multicolumn{2}{@{\,}l@{\,}}{1.1e6\mbox{\tiny (2e6)}} & \textbf{$\infty$} & \mbox{} & \textbf{$\infty$} & \mbox{} & \textbf{$\infty$} & \mbox{} & \textbf{$\infty$} & \mbox{} & \multicolumn{2}{@{\,}l@{\,}|}{$\infty$\,\textit{2e6}} & 0 & /13\\
RecPM-AOS \hspace*{\fill} & \textbf{1} & \textbf{}\mbox{\tiny (0)} & \textbf{1} & \textbf{}\mbox{\tiny (0)} & \textbf{$\infty$} & \mbox{} & \textbf{$\infty$} & \mbox{} & \textbf{$\infty$} & \mbox{} & \textbf{$\infty$} & \mbox{} & \multicolumn{2}{@{\,}l@{\,}|}{$\infty$\,\textit{2e6}} & 0 & /13\\
PM-AdapSS \hspace*{\fill} & \textbf{1} & \textbf{}\mbox{\tiny (0)} & \textbf{1} & \textbf{}\mbox{\tiny (0)} & \textbf{$\infty$} & \mbox{} & \textbf{$\infty$} & \mbox{} & \textbf{$\infty$} & \mbox{} & \textbf{$\infty$} & \mbox{} & \multicolumn{2}{@{\,}l@{\,}|}{$\infty$\,\textit{2e6}} & 0 & /13\\
F-AUC-MAB \hspace*{\fill} & \textbf{1} & \textbf{}\mbox{\tiny (0)} & \textbf{1} & \textbf{}\mbox{\tiny (0)} & \textbf{$\infty$} & \mbox{} & \textbf{$\infty$} & \mbox{} & \textbf{$\infty$} & \mbox{} & \textbf{$\infty$} & \mbox{} & \multicolumn{2}{@{\,}l@{\,}|}{$\infty$\,\textit{2e6}} & 0 & /13\\
Compass \hspace*{\fill} & \textbf{1} & \textbf{}\mbox{\tiny (0)} & \textbf{1} & \textbf{}\mbox{\tiny (0)} & \textbf{$\infty$} & \mbox{} & \textbf{$\infty$} & \mbox{} & \textbf{$\infty$} & \mbox{} & \textbf{$\infty$} & \mbox{} & \multicolumn{2}{@{\,}l@{\,}|}{$\infty$\,\textit{2e6}} & 0 & /13\\
JaDE \hspace*{\fill} & 845 & \mbox{\tiny (175)} & \multicolumn{2}{@{\,}l@{\,}}{7.7e5\mbox{\tiny (4e5)}} & \textbf{$\infty$} & \mbox{} & \textbf{$\infty$} & \mbox{} & \textbf{$\infty$} & \mbox{} & \textbf{$\infty$} & \mbox{} & \multicolumn{2}{@{\,}l@{\,}|}{$\infty$\,\textit{1e6}} & 0 & /13\\

\end{tabularx}
\providecommand{\algAtables}{\StrLeft{U-AOS-FW}{\ntables}}
\providecommand{\algBtables}{\StrLeft{R-SHADE}{\ntables}}
\providecommand{\algCtables}{\StrLeft{RecPM-AOS}{\ntables}}
\providecommand{\algDtables}{\StrLeft{PM-AdapSS}{\ntables}}
\providecommand{\algEtables}{\StrLeft{F-AUC-MAB}{\ntables}}
\providecommand{\algFtables}{\StrLeft{Compass}{\ntables}}
\providecommand{\algGtables}{\StrLeft{JaDE}{\ntables}}

\begin{tabularx}{1.0\textwidth}{@{}c@{}|*{7}{@{}r@{}X@{}}|@{}r@{}@{}l@{}}
\hline

\textbf{f20} & \multicolumn{2}{@{}c@{}}{82 \quad} & \multicolumn{2}{@{}c@{}}{46150 \quad} & \multicolumn{2}{@{}c@{}}{3.1e6 \quad} & \multicolumn{2}{@{}c@{}}{5.5e6 \quad} & \multicolumn{2}{@{}c@{}}{5.5e6 \quad} & \multicolumn{2}{@{}c@{}}{5.6e6 \quad} & \multicolumn{2}{@{}c@{}|}{5.6e6} & 14 & /15\\\hline
U-AOS-FW \hspace*{\fill} & 36 & \mbox{\tiny (4)} & \multicolumn{2}{@{\,}l@{\,}}{$\infty$} & \multicolumn{2}{@{\,}l@{\,}}{$\infty$} & \multicolumn{2}{@{\,}l@{\,}}{$\infty$} & \multicolumn{2}{@{\,}l@{\,}}{$\infty$} & \multicolumn{2}{@{\,}l@{\,}}{$\infty$} & \multicolumn{2}{@{\,}l@{\,}|}{$\infty$\,\textit{2e6}} & 0 & /13\\
R-SHADE \hspace*{\fill} & \textbf{11} & \textbf{}\mbox{\tiny (3)}$^{\star3}$ & 1 & .3\mbox{\tiny (0.2)} & \multicolumn{2}{@{\,}l@{\,}}{$\infty$} & \multicolumn{2}{@{\,}l@{\,}}{$\infty$} & \multicolumn{2}{@{\,}l@{\,}}{$\infty$} & \multicolumn{2}{@{\,}l@{\,}}{$\infty$} & \multicolumn{2}{@{\,}l@{\,}|}{$\infty$\,\textit{2e6}} & 0 & /13\\
RecPM-AOS \hspace*{\fill} & 44 & \mbox{\tiny (7)} & \multicolumn{2}{@{\,}l@{\,}}{$\infty$} & \multicolumn{2}{@{\,}l@{\,}}{$\infty$} & \multicolumn{2}{@{\,}l@{\,}}{$\infty$} & \multicolumn{2}{@{\,}l@{\,}}{$\infty$} & \multicolumn{2}{@{\,}l@{\,}}{$\infty$} & \multicolumn{2}{@{\,}l@{\,}|}{$\infty$\,\textit{2e6}} & 0 & /13\\
PM-AdapSS \hspace*{\fill} & 46 & \mbox{\tiny (4)} & \multicolumn{2}{@{\,}l@{\,}}{$\infty$} & \multicolumn{2}{@{\,}l@{\,}}{$\infty$} & \multicolumn{2}{@{\,}l@{\,}}{$\infty$} & \multicolumn{2}{@{\,}l@{\,}}{$\infty$} & \multicolumn{2}{@{\,}l@{\,}}{$\infty$} & \multicolumn{2}{@{\,}l@{\,}|}{$\infty$\,\textit{2e6}} & 0 & /13\\
F-AUC-MAB \hspace*{\fill} & 36 & \mbox{\tiny (6)} & 2 & .3\mbox{\tiny (1)} & \textbf{0} & \textbf{.16}\mbox{\tiny (0.2)}$^{\star}$ & \textbf{0} & \textbf{.10}\mbox{\tiny (0.0)}$^{\star2}$ & \textbf{0} & \textbf{.10}\mbox{\tiny (0.1)}$^{\star2}$ & \textbf{0} & \textbf{.10}\mbox{\tiny (8e-3)}$^{\star2}$ & \textbf{0} & \textbf{.10}\mbox{\tiny (0.2)}$^{\star2}$ & 12 & /13\\
Compass \hspace*{\fill} & 35 & \mbox{\tiny (3)} & \multicolumn{2}{@{\,}l@{\,}}{$\infty$} & \multicolumn{2}{@{\,}l@{\,}}{$\infty$} & \multicolumn{2}{@{\,}l@{\,}}{$\infty$} & \multicolumn{2}{@{\,}l@{\,}}{$\infty$} & \multicolumn{2}{@{\,}l@{\,}}{$\infty$} & \multicolumn{2}{@{\,}l@{\,}|}{$\infty$\,\textit{2e6}} & 0 & /13\\
JaDE \hspace*{\fill} & 24 & \mbox{\tiny (3)} & \textbf{1} & \textbf{.2}\mbox{\tiny (0.2)} & 0 & .46\mbox{\tiny (0.3)} & 0 & .74\mbox{\tiny (0.7)} & 1 & .1\mbox{\tiny (1)} & \multicolumn{2}{@{\,}l@{\,}}{$\infty$} & \multicolumn{2}{@{\,}l@{\,}|}{$\infty$\,\textit{1e6}} & 0 & /13\\

\end{tabularx}
\providecommand{\algAtables}{\StrLeft{U-AOS-FW}{\ntables}}
\providecommand{\algBtables}{\StrLeft{R-SHADE}{\ntables}}
\providecommand{\algCtables}{\StrLeft{RecPM-AOS}{\ntables}}
\providecommand{\algDtables}{\StrLeft{PM-AdapSS}{\ntables}}
\providecommand{\algEtables}{\StrLeft{F-AUC-MAB}{\ntables}}
\providecommand{\algFtables}{\StrLeft{Compass}{\ntables}}
\providecommand{\algGtables}{\StrLeft{JaDE}{\ntables}}

\begin{tabularx}{1.0\textwidth}{@{}c@{}|*{7}{@{}r@{}X@{}}|@{}r@{}@{}l@{}}
\hline

\textbf{f21} & \multicolumn{2}{@{}c@{}}{561 \quad} & \multicolumn{2}{@{}c@{}}{6541 \quad} & \multicolumn{2}{@{}c@{}}{14103 \quad} & \multicolumn{2}{@{}c@{}}{14318 \quad} & \multicolumn{2}{@{}c@{}}{14643 \quad} & \multicolumn{2}{@{}c@{}}{15567 \quad} & \multicolumn{2}{@{}c@{}|}{17589} & 15 & /15\\\hline
U-AOS-FW \hspace*{\fill} & 8 & .1\mbox{\tiny (2)} & 263 & \mbox{\tiny (229)} & 122 & \mbox{\tiny (106)} & 120 & \mbox{\tiny (210)} & 118 & \mbox{\tiny (68)} & 111 & \mbox{\tiny (161)} & 99 & \mbox{\tiny (171)} & 7 & /13\\
R-SHADE \hspace*{\fill} & \textbf{3} & \textbf{.0}\mbox{\tiny (1)}$^{\star3}$ & \textbf{6} & \textbf{.5}\mbox{\tiny (9)} & \textbf{7} & \textbf{.0}\mbox{\tiny (6)} & \textbf{7} & \textbf{.0}\mbox{\tiny (4)} & \textbf{6} & \textbf{.9}\mbox{\tiny (4)} & \textbf{6} & \textbf{.6}\mbox{\tiny (16)} & \textbf{5} & \textbf{.9}\mbox{\tiny (14)} & 13 & /13\\
RecPM-AOS \hspace*{\fill} & 659 & \mbox{\tiny (893)} & 689 & \mbox{\tiny (917)} & 320 & \mbox{\tiny (355)} & 315 & \mbox{\tiny (559)} & 308 & \mbox{\tiny (307)} & 290 & \mbox{\tiny (385)} & 257 & \mbox{\tiny (341)} & 4 & /13\\
PM-AdapSS \hspace*{\fill} & 309 & \mbox{\tiny (6)} & 1683 & \mbox{\tiny (2523)} & 781 & \mbox{\tiny (390)} & 769 & \mbox{\tiny (1257)} & 752 & \mbox{\tiny (1639)} & 708 & \mbox{\tiny (2280)} & 627 & \mbox{\tiny (398)} & 2 & /13\\
F-AUC-MAB \hspace*{\fill} & 32 & \mbox{\tiny (36)} & 39 & \mbox{\tiny (53)} & 80 & \mbox{\tiny (109)} & 121 & \mbox{\tiny (92)} & 223 & \mbox{\tiny (149)} & 466 & \mbox{\tiny (390)} & 659 & \mbox{\tiny (279)} & 2 & /13\\
Compass \hspace*{\fill} & 305 & \mbox{\tiny (1783)} & 490 & \mbox{\tiny (764)} & 320 & \mbox{\tiny (284)} & 315 & \mbox{\tiny (244)} & 308 & \mbox{\tiny (273)} & 290 & \mbox{\tiny (578)} & 257 & \mbox{\tiny (227)} & 4 & /13\\
JaDE \hspace*{\fill} & 7 & .4\mbox{\tiny (2)} & 26 & \mbox{\tiny (82)} & 16 & \mbox{\tiny (15)} & 16 & \mbox{\tiny (41)} & 15 & \mbox{\tiny (4)} & 15 & \mbox{\tiny (36)} & 14 & \mbox{\tiny (23)} & 12 & /13\\

\end{tabularx}
\providecommand{\algAtables}{\StrLeft{U-AOS-FW}{\ntables}}
\providecommand{\algBtables}{\StrLeft{R-SHADE}{\ntables}}
\providecommand{\algCtables}{\StrLeft{RecPM-AOS}{\ntables}}
\providecommand{\algDtables}{\StrLeft{PM-AdapSS}{\ntables}}
\providecommand{\algEtables}{\StrLeft{F-AUC-MAB}{\ntables}}
\providecommand{\algFtables}{\StrLeft{Compass}{\ntables}}
\providecommand{\algGtables}{\StrLeft{JaDE}{\ntables}}

\begin{tabularx}{1.0\textwidth}{@{}c@{}|*{7}{@{}r@{}X@{}}|@{}r@{}@{}l@{}}
\hline

\textbf{f22} & \textbf{467} & \mbox{} & \textbf{5580} & \mbox{} & \textbf{23491} & \mbox{} & \textbf{24163} & \mbox{} & \textbf{24948} & \mbox{} & \textbf{26847} & \mbox{} & \textbf{1.3e5} & \mbox{} & 12 & /15\\\hline
U-AOS-FW \hspace*{\fill} & 1913 & \mbox{\tiny (1072)} & 1196 & \mbox{\tiny (806)} & \textbf{$\infty$} & \mbox{} & \textbf{$\infty$} & \mbox{} & \textbf{$\infty$} & \mbox{} & \textbf{$\infty$} & \mbox{} & \multicolumn{2}{@{\,}l@{\,}|}{$\infty$\,\textit{2e6}} & 0 & /13\\
R-SHADE \hspace*{\fill} & 16 & \mbox{\tiny (2)} & \textbf{29} & \textbf{}\mbox{\tiny (33)} & \textbf{$\infty$} & \mbox{} & \textbf{$\infty$} & \mbox{} & \textbf{$\infty$} & \mbox{} & \textbf{$\infty$} & \mbox{} & \multicolumn{2}{@{\,}l@{\,}|}{$\infty$\,\textit{2e6}} & 0 & /13\\
RecPM-AOS \hspace*{\fill} & 791 & \mbox{\tiny (2145)} & 1197 & \mbox{\tiny (1792)} & \textbf{$\infty$} & \mbox{} & \textbf{$\infty$} & \mbox{} & \textbf{$\infty$} & \mbox{} & \textbf{$\infty$} & \mbox{} & \multicolumn{2}{@{\,}l@{\,}|}{$\infty$\,\textit{2e6}} & 0 & /13\\
PM-AdapSS \hspace*{\fill} & 369 & \mbox{\tiny (3)} & 808 & \mbox{\tiny (1613)} & \textbf{$\infty$} & \mbox{} & \textbf{$\infty$} & \mbox{} & \textbf{$\infty$} & \mbox{} & \textbf{$\infty$} & \mbox{} & \multicolumn{2}{@{\,}l@{\,}|}{$\infty$\,\textit{2e6}} & 0 & /13\\
F-AUC-MAB \hspace*{\fill} & 388 & \mbox{\tiny (47)} & 216 & \mbox{\tiny (327)} & \textbf{$\infty$} & \mbox{} & \textbf{$\infty$} & \mbox{} & \textbf{$\infty$} & \mbox{} & \textbf{$\infty$} & \mbox{} & \multicolumn{2}{@{\,}l@{\,}|}{$\infty$\,\textit{2e6}} & 0 & /13\\
Compass \hspace*{\fill} & \textbf{10} & \textbf{}\mbox{\tiny (4)} & 808 & \mbox{\tiny (806)} & \textbf{$\infty$} & \mbox{} & \textbf{$\infty$} & \mbox{} & \textbf{$\infty$} & \mbox{} & \textbf{$\infty$} & \mbox{} & \multicolumn{2}{@{\,}l@{\,}|}{$\infty$\,\textit{2e6}} & 0 & /13\\
JaDE \hspace*{\fill} & 28 & \mbox{\tiny (65)} & 246 & \mbox{\tiny (403)} & \textbf{$\infty$} & \mbox{} & \textbf{$\infty$} & \mbox{} & \textbf{$\infty$} & \mbox{} & \textbf{$\infty$} & \mbox{} & \multicolumn{2}{@{\,}l@{\,}|}{$\infty$\,\textit{1e6}} & 0 & /13\\

\end{tabularx}
\providecommand{\algAtables}{\StrLeft{U-AOS-FW}{\ntables}}
\providecommand{\algBtables}{\StrLeft{R-SHADE}{\ntables}}
\providecommand{\algCtables}{\StrLeft{RecPM-AOS}{\ntables}}
\providecommand{\algDtables}{\StrLeft{PM-AdapSS}{\ntables}}
\providecommand{\algEtables}{\StrLeft{F-AUC-MAB}{\ntables}}
\providecommand{\algFtables}{\StrLeft{Compass}{\ntables}}
\providecommand{\algGtables}{\StrLeft{JaDE}{\ntables}}

\begin{tabularx}{1.0\textwidth}{@{}c@{}|*{7}{@{}r@{}X@{}}|@{}r@{}@{}l@{}}
\hline

\textbf{f23} & 3.0 & \mbox{} & 1614 & \mbox{} & 67457 & \mbox{} & 3.7e5 & \mbox{} & 4.9e5 & \mbox{} & 8.1e5 & \mbox{} & 8.4e5 & \mbox{} & 15 & /15\\\hline
U-AOS-FW \hspace*{\fill} & 2 & .4\mbox{\tiny (2)} & $\infty$ & \mbox{} & $\infty$ & \mbox{} & \textbf{$\infty$} & \mbox{} & \textbf{$\infty$} & \mbox{} & \textbf{$\infty$} & \mbox{} & \multicolumn{2}{@{\,}l@{\,}|}{$\infty$\,\textit{2e6}} & 0 & /13\\
R-SHADE \hspace*{\fill} & 2 & .5\mbox{\tiny (2)} & \textbf{88} & \textbf{}\mbox{\tiny (61)} & \textbf{11} & \textbf{}\mbox{\tiny (12)}$^{\star3}$ & \textbf{$\infty$} & \mbox{} & \textbf{$\infty$} & \mbox{} & \textbf{$\infty$} & \mbox{} & \multicolumn{2}{@{\,}l@{\,}|}{$\infty$\,\textit{2e6}} & 0 & /13\\
RecPM-AOS \hspace*{\fill} & 2 & .4\mbox{\tiny (2)} & $\infty$ & \mbox{} & $\infty$ & \mbox{} & \multicolumn{2}{@{\,}l@{\,}}{\textbf{$\infty$}} & \textbf{$\infty$} & \mbox{} & \textbf{$\infty$} & \mbox{} & \multicolumn{2}{@{\,}l@{\,}|}{$\infty$\,\textit{2e6}} & 0 & /13\\
PM-AdapSS \hspace*{\fill} & \textbf{1} & \textbf{.8}\mbox{\tiny (3)} & $\infty$ & \mbox{} & $\infty$ & \mbox{} & \textbf{$\infty$} & \mbox{} & \textbf{$\infty$} & \mbox{} & \textbf{$\infty$} & \mbox{} & \multicolumn{2}{@{\,}l@{\,}|}{$\infty$\,\textit{2e6}} & 0 & /13\\
F-AUC-MAB \hspace*{\fill} & 2 & .0\mbox{\tiny (2)} & $\infty$ & \mbox{} & $\infty$ & \mbox{} & \textbf{$\infty$} & \mbox{} & \textbf{$\infty$} & \mbox{} & \textbf{$\infty$} & \mbox{} & \multicolumn{2}{@{\,}l@{\,}|}{$\infty$\,\textit{2e6}} & 0 & /13\\
Compass \hspace*{\fill} & 2 & .8\mbox{\tiny (3)} & $\infty$ & \mbox{} & $\infty$ & \mbox{} & \textbf{$\infty$} & \mbox{} & \textbf{$\infty$} & \mbox{} & \textbf{$\infty$} & \mbox{} & \multicolumn{2}{@{\,}l@{\,}|}{$\infty$\,\textit{2e6}} & 0 & /13\\
JaDE \hspace*{\fill} & 2 & .1\mbox{\tiny (2)} & 123 & \mbox{\tiny (31)} & $\infty$ & \mbox{} & \textbf{$\infty$} & \mbox{} & \textbf{$\infty$} & \mbox{} & \textbf{$\infty$} & \mbox{} & \multicolumn{2}{@{\,}l@{\,}|}{$\infty$\,\textit{1e6}} & 0 & /13\\

\end{tabularx}
\providecommand{\algAtables}{\StrLeft{U-AOS-FW}{\ntables}}
\providecommand{\algBtables}{\StrLeft{R-SHADE}{\ntables}}
\providecommand{\algCtables}{\StrLeft{RecPM-AOS}{\ntables}}
\providecommand{\algDtables}{\StrLeft{PM-AdapSS}{\ntables}}
\providecommand{\algEtables}{\StrLeft{F-AUC-MAB}{\ntables}}
\providecommand{\algFtables}{\StrLeft{Compass}{\ntables}}
\providecommand{\algGtables}{\StrLeft{JaDE}{\ntables}}

\begin{tabularx}{1.0\textwidth}{@{}c@{}|*{7}{@{}r@{}X@{}}|@{}r@{}@{}l@{}}
\hline

\textbf{f24} & 1.3e6 & \mbox{} & 7.5e6 & \mbox{} & 5.2e7 & \mbox{} & 5.2e7 & \mbox{} & 5.2e7 & \mbox{} & 5.2e7 & \mbox{} & 5.2e7 & \mbox{} & 3 & /15\\\hline
U-AOS-FW \hspace*{\fill} & \textbf{$\infty$} & \mbox{} & \textbf{$\infty$} & \mbox{} & \textbf{$\infty$} & \mbox{} & \textbf{$\infty$} & \mbox{} & \textbf{$\infty$} & \mbox{} & \textbf{$\infty$} & \mbox{} & \multicolumn{2}{@{\,}l@{\,}|}{$\infty$\,\textit{2e6}} & 0 & /13\\
R-SHADE \hspace*{\fill} & \textbf{$\infty$} & \mbox{} & \textbf{$\infty$} & \mbox{} & \textbf{$\infty$} & \mbox{} & \textbf{$\infty$} & \mbox{} & \textbf{$\infty$} & \mbox{} & \textbf{$\infty$} & \mbox{} & \multicolumn{2}{@{\,}l@{\,}|}{$\infty$\,\textit{2e6}} & 0 & /13\\
RecPM-AOS \hspace*{\fill} & \textbf{$\infty$} & \mbox{} & \textbf{$\infty$} & \mbox{} & \textbf{$\infty$} & \mbox{} & \textbf{$\infty$} & \mbox{} & \textbf{$\infty$} & \mbox{} & \textbf{$\infty$} & \mbox{} & \multicolumn{2}{@{\,}l@{\,}|}{$\infty$\,\textit{2e6}} & 0 & /13\\
PM-AdapSS \hspace*{\fill} & \textbf{$\infty$} & \mbox{} & \textbf{$\infty$} & \mbox{} & \textbf{$\infty$} & \mbox{} & \textbf{$\infty$} & \mbox{} & \textbf{$\infty$} & \mbox{} & \textbf{$\infty$} & \mbox{} & \multicolumn{2}{@{\,}l@{\,}|}{$\infty$\,\textit{2e6}} & 0 & /13\\
F-AUC-MAB \hspace*{\fill} & \textbf{$\infty$} & \mbox{} & \textbf{$\infty$} & \mbox{} & \textbf{$\infty$} & \mbox{} & \textbf{$\infty$} & \mbox{} & \textbf{$\infty$} & \mbox{} & \textbf{$\infty$} & \mbox{} & \multicolumn{2}{@{\,}l@{\,}|}{$\infty$\,\textit{2e6}} & 0 & /13\\
Compass \hspace*{\fill} & \textbf{$\infty$} & \mbox{} & \textbf{$\infty$} & \mbox{} & \textbf{$\infty$} & \mbox{} & \textbf{$\infty$} & \mbox{} & \textbf{$\infty$} & \mbox{} & \textbf{$\infty$} & \mbox{} & \multicolumn{2}{@{\,}l@{\,}|}{$\infty$\,\textit{2e6}} & 0 & /13\\
JaDE \hspace*{\fill} & \textbf{$\infty$} & \mbox{} & \textbf{$\infty$} & \mbox{} & \textbf{$\infty$} & \mbox{} & \textbf{$\infty$} & \mbox{} & \textbf{$\infty$} & \mbox{} & \textbf{$\infty$} & \mbox{} & \multicolumn{2}{@{\,}l@{\,}|}{$\infty$\,\textit{1e6}} & 0 & /13\\

\end{tabularx}
\end{minipage}}
\end{table*}

As seen in $f05 i01$, \UAOSFW reaches all targets within $10$ generations whereas \RecPMi takes $18$ generations. While both current-to-best/1 and best/2 operators are good operator choices to reach the targets for linear function, the faster speed is achieved with the utilisation of best/2 operator by most of the solutions though the generations. Although best/2 is one of the four operators adapted by \RecPMi, its best utilised in the presence of other new operators included which take care of premature convergence and stagnation.
For function $f07 i01$, a uni modal non-separable function, \UAOSFW decided to evolve solutions in the initial generations with the employment of rand-to-best/2 operator and rand/2 during the rest of the generations. It is clear from the graphs that \UAOSFW reaches optimum (92.94000000000176) with much higher speed ($103$ generations) than \RecPMi which manages to find solution with fitness $94.79693005303372$ in $11903$ generations.  
Thus, it becomes clear that adding new operators has improved the convergence speed by significant amount.

\section{Conclusion and Future work}
We conduct an exhaustive survey of adaptive selection of operators (AOS) in Evolutionary Algorithms (EAs). We looked at the commonality among AOS methods from the literature to combine them and we have added more components to the framework to built upon the existing categorisation of AOS methods. The formulas for the alternative choices of each component are presented. The goal is to select an AOS method from a range of AOS designs given by the presented framework such that the AOS method can learn to adaptively select the operators of differential evolution algorithm. Due to the large number of AOS choices, we employed \irace, an offline tuner, to select an AOS method for the \bbob problem set. The set of operators consists of nine mutation strategies that have shown good performance in the literature. 

\irace returned a variant of \RecPMaos named \UAOSFW. It outperformed the four well-known \irace tuned AOS methods, namely \Compass, \PMAdapSS, \FAUCMAB and \RecPMaos. Among the non-AOS methods, \UAOSFW outperformed \JaDE 
and \RSHADE in three function classes, namely low/moderate conditioning, high conditioning and multi modal classes. Overall, \UAOSFW solves 5\% less problems than \JaDE.
In the experiments, we concluded that adaptation of carefully selected nine operators has improved the speed of convergence relative to a subset of four operators.

\section*{Acknowledgment}
This research was conducted while the first author was a Ph.D. candidate with the Department of Computer Science at the University of York, UK.

\ifCLASSOPTIONcaptionsoff
  \newpage
\fi

\bibliographystyle{IEEEtranN}

\bibliography{literature,abbrev,authors,journals,abbrevshort,biblio,crossref}

\end{document}